\documentclass[lettersize,journal]{IEEEtran}
\usepackage{amsmath,amsfonts}
\usepackage{array}
\usepackage{textcomp}
\usepackage{stfloats}
\usepackage{url}
\usepackage{verbatim}
\usepackage{graphicx}
\usepackage{cite}
\usepackage{times}
\usepackage{amsmath}
\usepackage{amssymb}
\usepackage{amsthm}
\usepackage{blkarray}
\usepackage{threeparttable}
\usepackage{multirow}
\usepackage{makecell}
\usepackage{colortbl}
\usepackage{diagbox}
\usepackage{booktabs}
\usepackage{tikz}
\usepackage{fbox}
\usepackage{bm}
\usepackage{bbm}
\usepackage[linesnumbered,ruled,boxed,norelsize,vlined,commentsnumbered]{algorithm2e}
\usepackage[colorlinks,linkcolor=red,anchorcolor=blue,citecolor=black]{hyperref}
\usepackage{comment}
\usepackage{marginnote}
\usepackage{caption,subcaption}
\usepackage{xcolor}
\usepackage{cuted}

\newcommand{\revise}[1]{{\color{black}{#1}}}

\newcommand{\diag}{{\mathop{\mathrm{diag}}}}
\newcommand{\ie}{{\textit{i.e.}}}
\newcommand{\etal}{{\textit{et al.}}}
\newcommand{\eg}{{\textit{e.g.}}}
\newcommand{\best}[1]{{\color{blue}{#1}}}
\newcommand{\second}[1]{{\color{orange}{#1}}}
\newcommand{\AngErr}{{\mathop{\mathrm{AngErr}}}}
\newcommand{\Tr}{{\mathop{\mathrm{Tr}}}}

\newtheorem{theorem}{Theorem}

\newtheorem{definition}{Definition}

\definecolor{best}{rgb}{0.77,1.0,0.77}
\definecolor{second}{rgb}{1,0.77,0.77}

\graphicspath{{./fig1/}, {./photo/}}

\begin{document}

\title{GraphReg: Dynamical Point Cloud Registration with
	Geometry-aware Graph Signal Processing}

\author{Mingyang~Zhao,
	Lei~Ma,
	Xiaohong~Jia,
	Dong-Ming Yan,
	~and Tiejun Huang
\thanks{This work was supported in part by the National Key R\&D Program of China (2021YFB1715900), the National Natural Science of Foundation for Outstanding Young Scholars under Grant 12022117, CAS Project for Young Scientists in Basic Research under Grant No.YSBR-034, the National Natural Science Foundation of China under Grant 61872354 and Grant 62172415, the Open Research Fund Program of State key Laboratory of Hydroscience and Engineering, Tsinghua University under Grant sklhse-2022-D-04. \emph{(Corresponding author: Dong-Ming Yan and Lei Ma.)}}
\thanks{M. Zhao is with the Beijing Academy of Artificial Intelligence (BAAI) and the National Laboratory of Pattern Recognition (NLPR), Institute of Automation, CAS, Beijing, China. E-mail: myzhao@baai.ac.cn.}
\thanks{L. Ma and T. Huang are with the BAAI and Institute for Artificial Intelligence, Peking University, Beijing, China. Ma is also with National Biomedical Imaging Center, Peking University. Huang is also with National Engineering Research Center of Visual Technology, School of Computer Science, Peking University. E-mail: \{lei.ma, tjhuang\}@pku.edu.cn.\protect}
\thanks{X. Jia is with the Academy of Mathematics and Systems Science, CAS, and UCAS, Beijing, China. E-mail: xhjia@amss.ac.cn.\protect}
\thanks{D. Yan is with the NLPR, Institute of Automation, CAS, and School of AI, UCAS, Beijing, China. E-mail: yandongming@gmail.com.\protect}
}
\markboth{IEEE TRANSACTIONS ON IMAGE PROCESSING}%
{Shell \MakeLowercase{\textit{et al.}}: A Sample Article Using IEEEtran.cls for IEEE Journals}


\maketitle

\begin{abstract}
This \revise{study} presents a high-accuracy, efficient, and physically induced method for 3D point cloud registration, which is the core of many important 3D vision problems. \revise{In contrast to} existing physics-based methods that merely consider spatial point information and ignore \emph{surface geometry}, we explore geometry aware rigid-body dynamics to regulate the particle (point) motion, which \revise{results in} more precise and robust registration. Our proposed method consists of four major modules. First, we leverage the \emph{graph signal processing} (GSP) framework to define a new signature, \ie, ~{\emph{point response intensity}} for each point, by which we \revise{succeed in} describing the local surface variation, resampling keypoints, and distinguishing different particles. Then, to address the shortcomings of \revise{current} physics-based approaches that are \revise{sensitive} to outliers, we accommodate the defined {{point response intensity}} to \emph{median absolute deviation} (MAD) in robust statistics and adopt the \emph{X84 principle} for adaptive outlier depression, ensuring a robust and stable registration. Subsequently, we propose a novel \emph{geometric invariant} under rigid transformations to incorporate higher-order features of point clouds, which is further embedded for force modeling to guide the correspondence between pairwise scans \revise{credibly}. Finally, we introduce an \emph{adaptive simulated annealing} (ASA) method to search for the global optimum and substantially accelerate the registration process. We perform comprehensive experiments to evaluate the proposed method on various datasets captured from range scanners to LiDAR. Results demonstrate that our proposed method outperforms representative state-of-the-art approaches in terms of {accuracy} and is more suitable for registering large-scale point clouds. Furthermore, it is \revise{considerably} faster and more robust than most competitors. \color{black}{Our implementation is publicly available at \url{https://github.com/zikai1/GraphReg}}.
\end{abstract}


\section{Introduction} 
\begin{figure}[t]
	\includegraphics[width=0.45\textwidth]{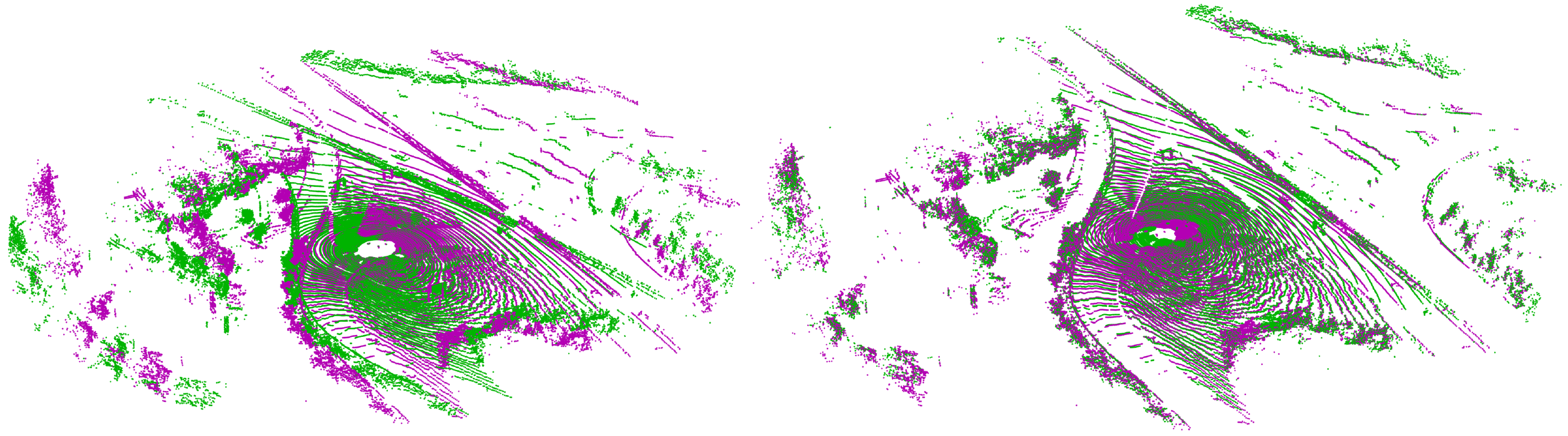}
	
	\includegraphics[width=0.45\textwidth]{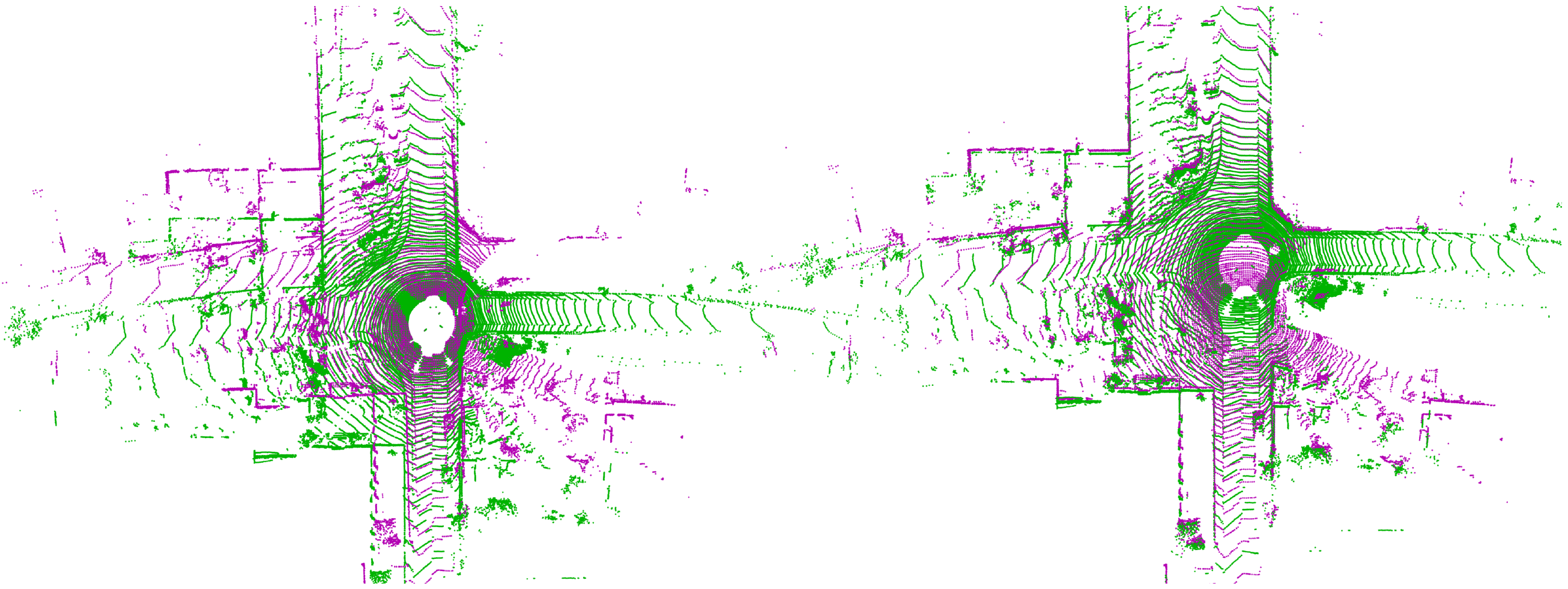}
	\vskip -0.2cm
	\caption{Registration on the large-scale KITTI LiDAR point clouds~\cite{geiger2012we}. Left: Misalignment of the source (green) and the target (purple) point clouds; right: Registration by the proposed method. It achieves high-quality results with deviations less than $1^\circ$ ($0.1267^\circ$ and $0.3385^\circ$ of the two scenes, respectively).}
	\vskip -0.5cm
	\label{fig:intro}
\end{figure}

Point cloud registration is an important problem in computer vision and robotics, with many applications in various domains, such as 3D reconstruction and motion estimation, simultaneous localization and mapping (SLAM), shape recognition or retrieval, and cultural heritage.
The goal is to estimate the \revise{optimal} spatial transformation, including the rotation matrix $\bf{R}$ and the translation vector $\bf{t}$ (6DoF), to align two point clouds into a common coordinate system, \ie, coincide the \textit{source} point cloud $\mathcal{X}$ to the \textit{target}  $\mathcal{Y}$ in an optimal \revise{manner} (see Fig.~\ref{fig:intro} as an example). 

The most widely used registration method, \textit{iterative closest point} (ICP)~\cite{besl1992method}, iteratively assumes point proximity and computes the transformation $(\mathbf{R}, \mathbf{t})\in SO(3)$\footnote{$SO(3):=\{\mathbf{R}\in \mathbb{R}^{3 \times 3}|\mathbf{R}\mathbf{R}^{T}=\mathbf{R}^{T}\mathbf{R}=\mathbf{I}_3, \det{\mathbf{R}}=1\}$.}$\times \mathbb{R}^3 $ by the \emph{least-squares estimator} and the \emph{singular value decomposition} (SVD). ICP usually performs efficiently and has high accuracy for small-scale scenarios~\cite{jauer2018efficient}; however, it is susceptible to noise and outliers and heavily depends on the initial alignment~\cite{gao2019filterreg}. \revise{Given} its simplicity, many variants of ICP have been proposed to mitigate these problems~\cite{song2015local}. However,  a universal variant that jointly handles most registration challenges \revise{is still lacking}.

Alternatively, probabilistic registration methods generalize the binary  correspondence of ICP to \emph{soft assignments} with probability; \revise{thus,} they are more robust against outliers in principle~\cite{jian2010robust,myronenko2010point,horaud2010rigid}. The probabilistic framework commonly formulates points as the centroids of the Gaussian density functions; then, the whole point cloud is interpreted as a \textit{Gaussian mixture model} (GMM). Probabilistic methods utilize \textit{kernel correlation} (KC) or \revise{\textit{expectation-maximization}} (EM) to update the point correspondences and transformations by turns, resulting in improved robustness~\cite{biber2003normal,tsin2004correlation}. However, these methods are 
typically computationally expensive and much slower than ICP-based approaches~\cite{golyanik2016gravitational}, especially for large-scale or high-resolution scans.

Recently, physics-based methods have shown great potential for point cloud registration~\cite{golyanik2016gravitational,jauer2018efficient,golyanik2019accelerated}, in which each point is interpreted as a particle (the terms \emph{point} and \emph{particle} are used synonymously), and the rigid transformation (dynamical movement) is realized through virtual forces, such as gravity or springs. These simulation methods are appealing \revise{because} the force computation can be parallelized on modern GPUs, 
ensuring high computational efficiency. {However, most of them trivially utilize the point coordinates and ignore the local surface property, such as \emph{geometry} or \emph{topology}, as well as the \emph{surface variation}, which empirically enables high registration accuracy. In addition, they equally treat each particle and fail to handle outlier contamination, which is inevitable in practical applications. However, point clouds are typically unorganized, thus making it difficult to  describe their surface properties explicitly. The present study is inspired by the impressive progress of \emph{graph signal processing} (GSP)~\cite{ortega2018graph,deng2019saliency,zeng20193d,sakiyama2019eigendecomposition}, which expands classical signal processing methods to irregular graph structure~\cite{sandryhaila2013discrete,sandryhaila2014big} and can effectively represent local connections among points,  thus providing powerful tools for high-dimensional data analysis. In this work, we choose to leverage the GSP framework to model the surface geometry systematically and jointly consider point attributes, from which we can define useful geometric invariants, depress the outlier disturbance, boost the point-pair correspondence, and integrate them to  advance the performance of dynamical point cloud registration in terms of precision, efficiency, and robustness.
 }

Concretely, we use a Haar-like graph filter in GSP to encode the local neighbor connections of point clouds to define a new signature--{point response intensity}, which is closely related to the local surface geometry. We prove that the \revise{point response intensity} is invariant under rigid transformations and then take it as \emph{particle mass} to weigh or distinguish different particles of the many-particle system (point cloud). \revise{For efficiency}, we resample points with {high response intensity} (keypoints) and design a novel \textit{geometric invariant} based on the \textit{graph Laplacian operator} {from GSP}~\cite{sandryhaila2013discrete}, which is also invariant about the rotation and translation of point clouds. Critically, to make point correspondence discriminative, we combine  \emph{Coulomb's law} of electrical field and the proposed geometric invariant to induce electrical forces. The registration of point clouds is then recast as solving differential equations derived from rigid-body dynamics. We also propose a novel outlier depression scheme by accommodating the {point response intensity defined via GSP} to robust statistics on the basis of median absolute deviation, from which most noisy outliers are successfully removed. Moreover, we introduce an {adaptive simulated annealing} scheme to find the global solution and \revise{considerably} boost the convergence of the designed algorithm. Finally, we conduct comprehensive experiments on a wide variety of datasets to assess the performance of the proposed method \revise{with respect to} precision, robustness, and efficiency. \revise{Results confirm the prominent advantages of our method over representative state-of-the-art approaches}. The main contributions of this work are summarized as follows: 
\begin{itemize}
	\item We propose a novel method integrating \revise{the} GSP framework with rigid-body dynamics for point cloud registration, which achieves high accuracy and efficiency.
	\item We use GSP to model the local surface geometry and derive new geometric invariants to estimate the correspondence confidence between point pairs with probability.
	\item We accommodate the proposed \revise{point response intensity} to robust statistics for outlier depression, developing a highly robust and stable registration framework.
	\item We introduce an adaptive simulated annealing scheme to solve a globally optimal transformation, which significantly accelerates the registration process. 
\end{itemize} 
The rest of the paper is organized as follows: Section~\ref{related} reviews the point cloud registration approaches related to our study. The technical  contributions of our proposed method are presented in Section~\ref{method}, followed by the performance evaluation on \revise{various} numerical experiments in Section~\ref{sec:experiment}. We conclude our study, discuss the limitations, and point out possible extensions in Section~\ref{conclusion}. 

\section{Related Work}\label{related}
\revise{Many} pioneering works have been developed for point cloud registration. We review the most relevant work in this section. Readers \revise{can refer} to~\cite{pomerleau2015review} for a  comprehensive overview.

\subsection{ICP-based Methods}
Initially, most registration methods are based on the classical ICP~\cite{besl1992method}. Over the past several decades, many variants have been proposed. For instance, to improve  registration accuracy and efficiency, the original point-to-point correspondence error was replaced by other metrics such as the point-to-plane~\cite{chen1992object} distance.
Zhou \etal~\cite{park2017colored} integrated color information to deduce a joint color point space, but these approaches are limited to the color property. Aiger~\etal~\cite{aiger20084} extracted all coplanar 4-point congruent sets (4PCS) and aligned them between two point clouds, which is later \revise{sped} up by~\cite{mellado2014super}
to linear time complexity using smart indexing. To circumvent the weakness of ICP that converges to the local minimum, Yang~\etal~\cite{yang2013go} combined \emph{branch-and-bound} (BnB) with ICP to  seek the globally optimal solution exhaustively; \revise{however, the combination} suffered from high complexity~\cite{dong2020registration}. Another line of studies focuses on promoting robustness against noise and outliers, such as the pioneer trimmed ICP (TrICP)~\cite{chetverikov2002trimmed} and the use of sparsity-inducing $l_p~(p\in [0,1])$ norm~\cite{bouaziz2013sparse}. Zhou~\etal~\cite{zhou2016fast} designed a fast global registration (FGR) method with \emph{fast point feature histograms}~\cite{rusu2009fast}, and used the \emph{Geman--McClure estimator} to reduce  outlier influence. Recently, Yang~\etal~\cite{yang2020teaser} combined the \emph{Truncated least-squares (TLS) Estimation And SEmidefinite Relaxation} (TEASER) to register point clouds \revise{certifiably}, in which outliers or spurious correspondence were restrained by TLS. Zhang~\etal~\cite{zhang2021fast} leveraged an effective error metric \emph{Welsch's function} to achieve robust registration and proposed the fast and robust ICP (FRICP).

\subsection{Probabilistic Methods}
Probabilistic \revise{frameworks} usually use GMM to model the point cloud distribution and formulate the registration task as a density estimation problem, which can generally be categorized into KC- and EM-based methods. KC-based approaches model both point clouds. Tsin~\etal~\cite{tsin2004correlation} described both point clouds as GMMs, took KC as an affinity measure and found the transformation by minimizing {Renyi’s quadratic entropy} between these two distributions. Similarly, Jian~\etal~\cite{jian2010robust} aligned two GMMs, but the objective function was based on the $L_2$ distance. Instead, EM-based algorithms represent one point cloud (source) as GMM centroids and treat the other point cloud (target) as data points. Myronenko and Song~\cite{myronenko2010point} proposed the \emph{coherent point drift} (CPD), which fitted the isotropic GMM to the target point cloud under the maximum likelihood sense, \revise{and} then the correspondence and transformation were iteratively updated by EM.  Horaud~\etal~\cite{horaud2010rigid} used general covariance matrices to model the Gaussian component and estimated parameters \revise{via} EM-like \emph{expectation conditional maximization} (ECM). As incorporate an additional \emph{uniform distribution} to GMM, CPD and ECM are robust against noise and outliers. However, they require \emph{a priori} setting of outlier ratio to trade off GMM and uniform distribution, which is commonly unknown in practice. \revise{In addition}, probabilistic approaches are often associated with expensive computational cost, especially for large-scale point clouds.

\subsection{Physics-based Methods}
Recently, point cloud registration analogies to physical processes are gaining much popularity. Golyanik~\etal~\cite{golyanik2016gravitational} took each point as a single body and moved them under a gravitational field. \revise{Similar to} ICP, \revise{rotation was} estimated by SVD, whereas the translation was based on the rigid-body dynamics. The method showed satisfactory results and was comparable to ICP and CPD. However, as pointed out in~\cite{golyanik2019accelerated}, its quadratic time complexity impeded  widespread practical applications. Jauer~\etal~\cite{jauer2018efficient} designed an efficient framework based on the principle of the classical mechanics or thermodynamics and improved the registration accuracy by \revise{involving} the RGB feature, although this involvement is not common for scans without color. These physics-based methods cast a new light for point cloud registration and exhibit promising results; however, \revise{most of them merely focus on spatial points, and little has been explored to integrate the more \emph{universal} and \emph{effective} geometric features instead of color. Moreover, they give each point, including outliers, the same weight, and the outlier-contaminated scenarios are not fully emphasized, which inevitably \revise{degrade} their performance.} {To address these problems, in this paper, we invoke the GSP framework to capture the geometry of point clouds and develop useful geometric features to boost the accuracy, robustness, and convergence of physics-based point cloud registration.}

\subsection{Learning-based Methods}
Recent advances in \emph{deep neural networks} for learning in 2D image \revise{domains} have led to the development of 3D deep learning for point cloud registration. Aoki~\etal~\cite{aoki2019pointnetlk} leveraged PointNet~\cite{qi2017pointnet} for feature representation and \revise{adopted} it into the classical image alignment algorithm, \emph{Lucas \& Kanade} (LK), to form a PointNetLK structure. Inspired by ICP, Wang~\etal~\cite{wang2019deep} designed a transformer network along with SVD for point-to-point registration. Choy~\etal~\cite{choy2020deep} used \emph{fully convolutional geometric features} (FCGF) as keypoint descriptors, \revise{and} then the pairwise correspondence between two point clouds was predicted by a 6-dimensional convolutional network, followed by the solution analogous to a weighted ICP. Deep learning-based methods embody the advantages to represent features effectively; however, at present, the main problems are from the longstanding training phase. To the best of our knowledge, the application of trained networks to multiple types of point clouds is not immediate, which is still an open challenge.

{\subsection{GSP for Point Clouds}
Signal processing on graphs generalizes the concepts and methodology from conventional signal processing theory to data indexed by the graph structure~\cite{shuman2013emerging,sandryhaila2014big}. GSP models each data element as a node and their relationship, such as similarity by an edge. On the basis of this construction, the concept including graph filter, spectrum ordering, and Fourier transform can be conveniently defined and applied. Benefiting from the modeling capabilities of GSP, some approaches have exploited GSP for point cloud analysis. For instance, Zhang~\etal~\cite{zhang2020hypergraph} conducted 3D denoising by estimating the hypergraph spectrum components and frequency coefficients of point clouds, whereas Dinesh~\etal~\cite{dinesh2020point}  designed a signal induced by a feature graph Laplacian {regularizer to remove both Gaussian and Laplacian-like point cloud noise}. In addition, the resampling of point clouds through the detection of high-frequency signal components has facilitated the development of point cloud storage and processing. Chen~\etal~\cite{chen2017fast} developed a high-pass graph filter for fast point cloud resampling, by which the visualization and shape modeling of large-scale point clouds were achieved. Deng~\etal~\cite{deng2022efficient} went a step further and used hypergraph spectral filters to capture signal interactions of point clouds for resampling. In contrast to the above approaches, we take data points as graph nodes and particles jointly and exploit GSP for the embedding of inherent surface geometry to remove outliers and constrain the dynamical motion of particles for point cloud registration.}

\section{Methodology}\label{method}
This section presents the main contributions of our study. First, GSP is customized to describe the local surface topology, followed by an effective outlier pruning method based on our proposed \revise{point response intensity}. Then, graph-induced geometric invariants are used to guide the dynamical registration 
process. Finally, a fast global optimization scheme for maximizing energy is presented.

\subsection{Graph Construction}
Given two point clouds with finite cardinality $\mathcal{X}=\{\bm{x}_i\in \mathbb{R}^3\}_{i=1}^M$ and $\mathcal{Y}=\{\bm{y}_j\in \mathbb{R}^3\}_{j=1}^N$, we call $\mathcal{X}$ as the \emph{source} and $\mathcal{Y}$ as the \emph{target}. Our purpose is to align these two point clouds into a common coordinate system. We use \textit{graph nodes} to represent points and encode the local geometry of each point cloud by an \emph{adjacency matrix} $\mathbf{W} \in \mathbb{R}^{M \times M}$ (take $\mathcal{X}$ as the example) with the edge weight
\begin{eqnarray}
	\mathbf{W}_{ij}=\left\{
	\begin{aligned}
		&e^{-\frac{\|\bm{x}_i-\bm{x}_j\|_{2}^{2}}{\sigma^2}},& \|\bm{x}_i-\bm{x}_j\|_{2}^{2}<\tau; \\
		&0, &otherwise.\\
	\end{aligned}
	\right.
	\label{eq:weight}
\end{eqnarray}
Here $\|\bm{x}_i-\bm{x}_j\|_{2}$ is the Euclidean distance between $\bm{x}_i$ and $\bm{x}_j$, $\sigma$ is the variance of graph nodes, and $\tau$ is a distance threshold controlling the graph sparsity. Hereinafter, we use graphs $\mathcal{G}_{\mathcal{X}}=<\mathcal{X}, \mathbf{W}_\mathcal{X}>$ and $\mathcal{G}_{\mathcal{Y}}=<\mathcal{Y}, \mathbf{W}_\mathcal{Y}>$ to express point clouds and their local neighbor weights.

\subsection{Graph-induced Point Response \revise{Intensity}} 
We adopt the above graph structure $\mathcal{G}_{\mathcal{X}}$, $\mathcal{G}_{\mathcal{Y}}$ and introduce \textit{graph filtering}~\cite{chen2017fast}  
to model the local surface variation of point clouds, which can be quantitatively evaluated by the following defined \revise{\emph{point response intensity} $\mathcal{I}(\bm{x})$}.

Specifically, let matrix $\mathbf{A} \in \mathbb{R}^{M \times M}$
be a \emph{graph-shift operator}, which can be realized by the, \eg, \emph{adjacency matrix} $\mathbf{W}$,  \emph{transition matrix} $\mathbf{D}^{-1}\mathbf{W}$, {\emph{graph Laplacian matrix}} $\mathbf{D}-\mathbf{W}$, or other structure-related matrices, where $\mathbf{D}=\diag(d_1, d_2, \cdots, d_M)$ is the diagonal \textit{degree matrix}, and $d_i=\sum_{\bm{x}_j\in\mathcal{X}}\mathbf{W}_{ij}$. Each linear and shift-invariant \emph{graph filter} $h(\mathbf{A})$ is a polynomial function of $\mathbf{A}$~\cite{sandryhaila2013discrete}:
\begin{eqnarray}
	h(\mathbf{A})=\sum_{l=0}^{L-1}h_{l}\mathbf{A}^{l}=h_{0}\mathbf{I}+h_{1}\mathbf{A}+...+h_{L-1}\mathbf{A}^{L-1},
\end{eqnarray}
where $\{h_{l}\}_{l=0}^{L-1}$ and $L$ are the coefficients and length of $h(\mathbf{A})$.

Given a \revise{general} \textit{graph signal} $\revise{\bm{s}(\bm{x}_{i})}\in \mathbb{R}^n$ defined at the graph vertex $\bm{x}_{i}$ as an input, the graph filter $h(\mathbf{A})$ outputs another new graph signal, \ie, $h(\mathbf{A})\cdot \revise{\bm{s}(\bm{x}_{i})}$. If $\mathbf{A}=\mathbf{D}^{-1}\mathbf{W}$, then a high-pass filter referred to as \emph{Haar-like graph filter}~\cite{chen2017fast} is attained $h_{Har}(\mathbf{A})=\mathbf{I}-\mathbf{A}$. {Let $\mathbf{X}=[\bm{x}_{1}, \bm{x}_{2}, \cdots, \bm{x}_{M}]^T$ represent the point cloud $\mathcal{X}$, $\bm{s}(\bm{x}_{i})=\bm{x}_{i}\in \mathbb{R}^3$, 
then the new output graph signal of each $\revise{\bm{s}(\bm{x}_{i})}$ after graph filtering is~\cite{chen2017fast} 
\begin{eqnarray}
	(h_{Har}(\mathbf{A}) \cdot \mathbf{X})_{i}:=\bm{x}_{i}-\sum_{\bm{x}_{j}\in \mathcal{N}_{i}}\mathbf{A}_{ij}\bm{x}_{j}, 
\end{eqnarray}}
where $\mathcal{N}_{i}=\{\bm{x}_{j}\in \mathcal{X}|\mathbf{W}_{ij}\neq0\}$.
\begin{definition}
	The \revise{point response intensity} of each point $\bm{x}_{i}$ is defined as $\revise{\mathcal{I}(\bm{x}_{i})=\|(h_{Har}(\mathbf{A}) \cdot \mathbf{X})_{i}\|_2^{2}}$, which reflects the local surface variation of $\bm{x}_{i}$ with respect to its neighborhoods. 
\end{definition}

\begin{theorem}
	\label{the1}
	The total \revise{point response intensity} (local surface variation) $\revise{\mathcal{I}(\mathbf{X})=\diag(h_{Har}(\mathbf{A})\mathbf{X}\mathbf{X}^Th_{Har}(\mathbf{A})^T) \in \mathbb{R}^M}$ is invariant under the rotation and translation of point clouds.
\end{theorem} 
We provide all proofs in Supplemental Material.

\begin{figure}[t]
	\centering
	\includegraphics[width=0.4\textwidth]{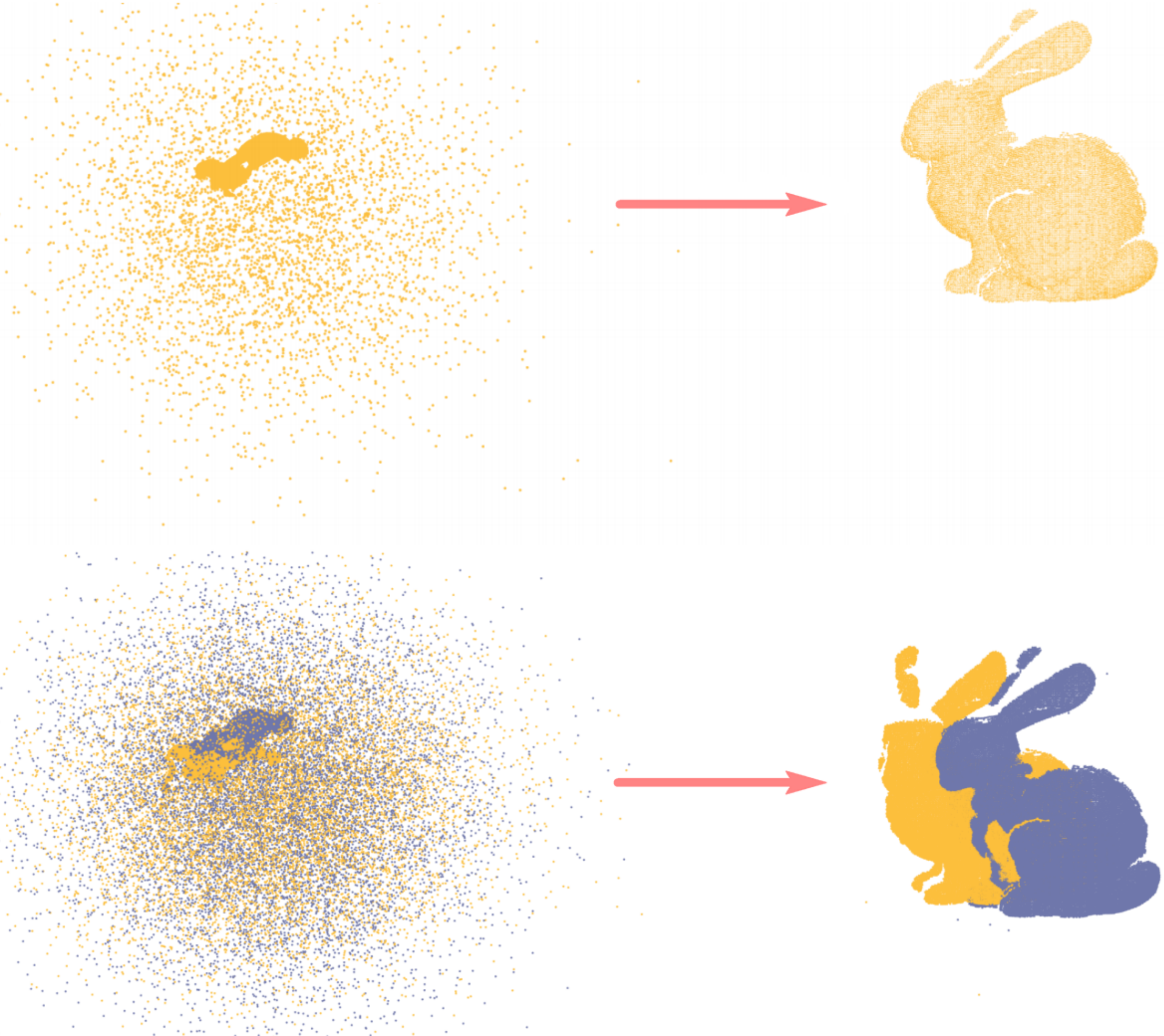}
	\vskip -0.2cm
	\caption{Outlier depression by the proposed \revise{point response intensity $\mathcal{I}(\bm{x})$} and MAD. Left: Outlier-contaminated point clouds, where the outlier percentage of the two rows are 20\% and 50\%, \ie, there are separately 4,153 and 10,785+10,384=21,169 outliers in the two scenes. Right: The outlier pruning results.   }
	\label{fig:outlier_prune}
	\vskip -0.3cm
\end{figure}

\subsection{Outlier Depression by the \revise{Point Response Intensity}}
\emph{Outliers are that lie outside the overall pattern of the underlying distribution or data model}~\cite{cohen2015sampler}. To depress their adverse \revise{effect} for point cloud registration, we propose to filter them out by the designed \revise{point response intensity $\mathcal{I}(\bm{x})$}. The motivation is that \revise{$\mathcal{I}(\bm{x})$} can effectively describe the local surface variation, \revise{whereas} outliers usually have more extreme values than correctly measured points due to their \emph{scatter properties}, as illustrated in the left panel of Fig.~\ref{fig:outlier_prune}. \revise{The outliers are sampled from zero mean Gaussian distribution with the} {standard variance determined by the size of the point cloud bounding box, with an aim to simulate the outside disturbance}.

We \revise{use} the \emph{X84 criterion} from robust statistics~\cite{hampel2011robust} \revise{(pp. 64--69)} to set the rejection threshold, {which is based upon the median and median deviation instead of the mean and standard deviation}. \revise{Benefiting from} the robust location estimator \emph{median}, the \emph{median absolute deviation} (MAD) is defined as 
\begin{eqnarray}
	\text{MAD}(\revise{\mathcal{I}(\mathbf{X}))}=\mathop{med}_{i=1,\cdots, M}\{|\revise{\mathcal{I}(\bm{x}_i)}-\mathop{med}_{j=1,\cdots, M}(\revise{\mathcal{I}(\bm{x}_j))}|\},
\end{eqnarray}
\revise{which is more robust than the interquartile range and typically has 50\% breakdown point for the {distribution scale} estimation~\cite{hampel2011robust}}. As reported in \cite{hampel2011robust}, the X84 criterion regulates to remove the point $\bm{x}_i$ with \revise{$\mathcal{I}(\bm{x}_i)>\alpha \cdot \text{MAD}(\mathcal{I}(\mathbf{X}))$}, where \revise{$\alpha=5.2$ is recommended without additional assumptions}. {If we assume $\mathcal{I}(\bm{x}_i)$ satisfying Gaussian distribution, then $5.2$MAD approximately corresponds to the $3\sigma$ of Gaussian distribution, where $\sigma$ is the standard deviation~\cite{hampel2011robust} (p. 69).} The right panel of Fig.~\ref{fig:outlier_prune} shows sample results after outlier depression. It is observed that our proposed method based on the point response \revise{intensity $\mathcal{I}(\bm{x})$} successfully removes most of the noisy outliers, \revise{hence ensuring a highly robust dynamical registration of outlier-contaminated point clouds}.

\subsection{Graph-based Feature Description}
\begin{figure}[t]
	\centering
	\subcaptionbox{$M=921,753$}{
		\includegraphics[width=0.14\textwidth]{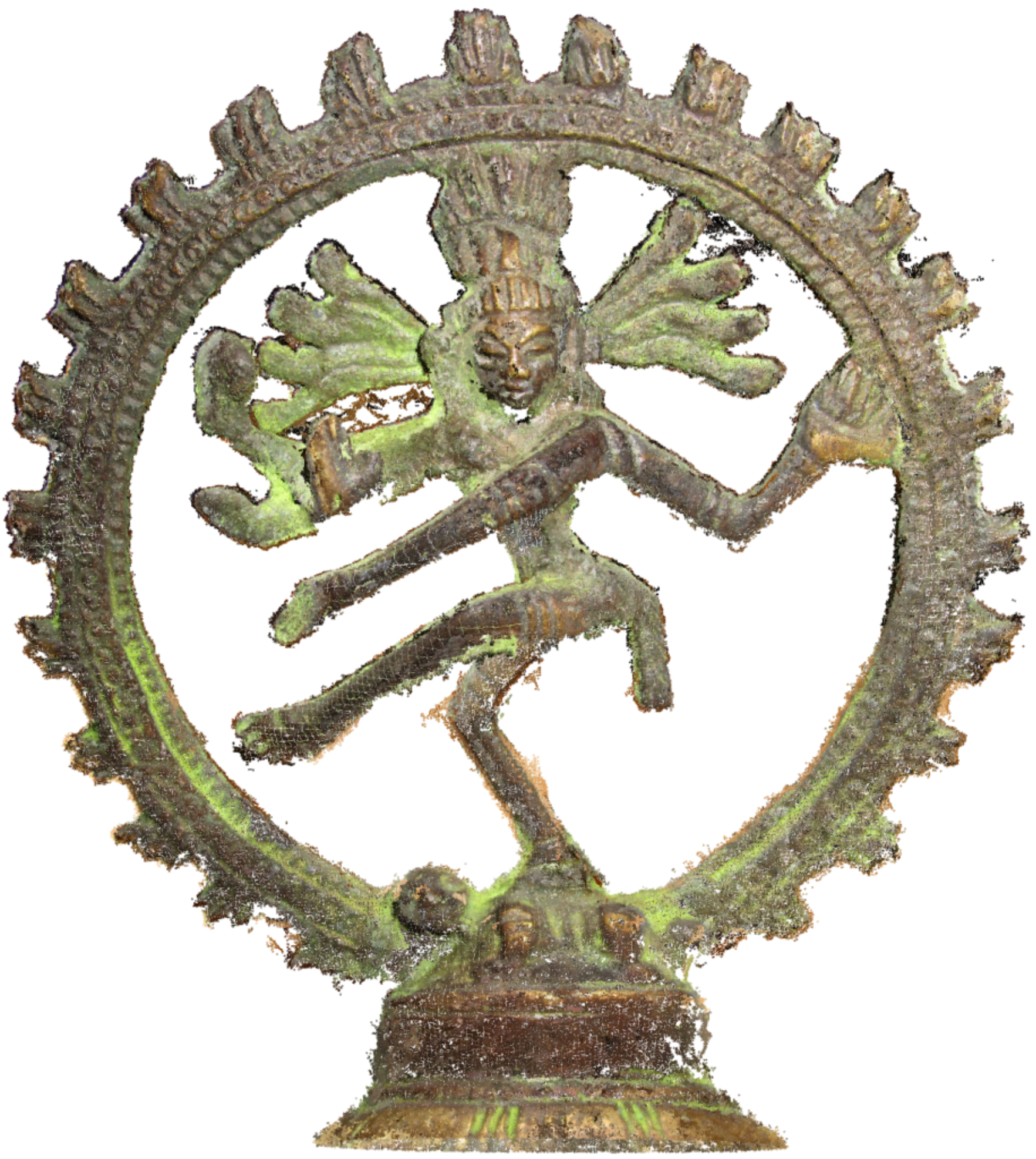}
	}
	\subcaptionbox{$M/100$}{
		\includegraphics[width=0.14\textwidth]{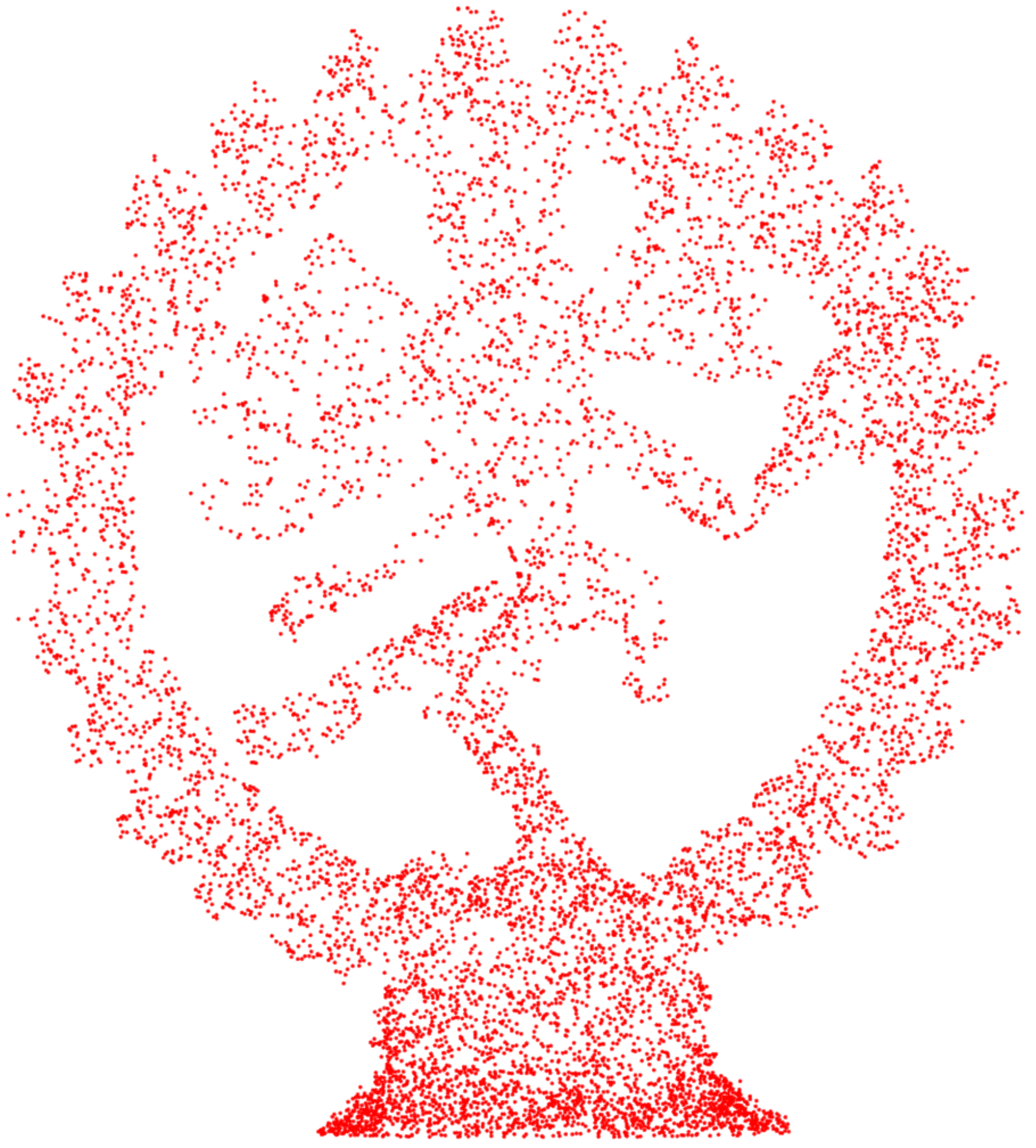}
	}
	\subcaptionbox{$M/1000$}{
		\includegraphics[width=0.14\textwidth]{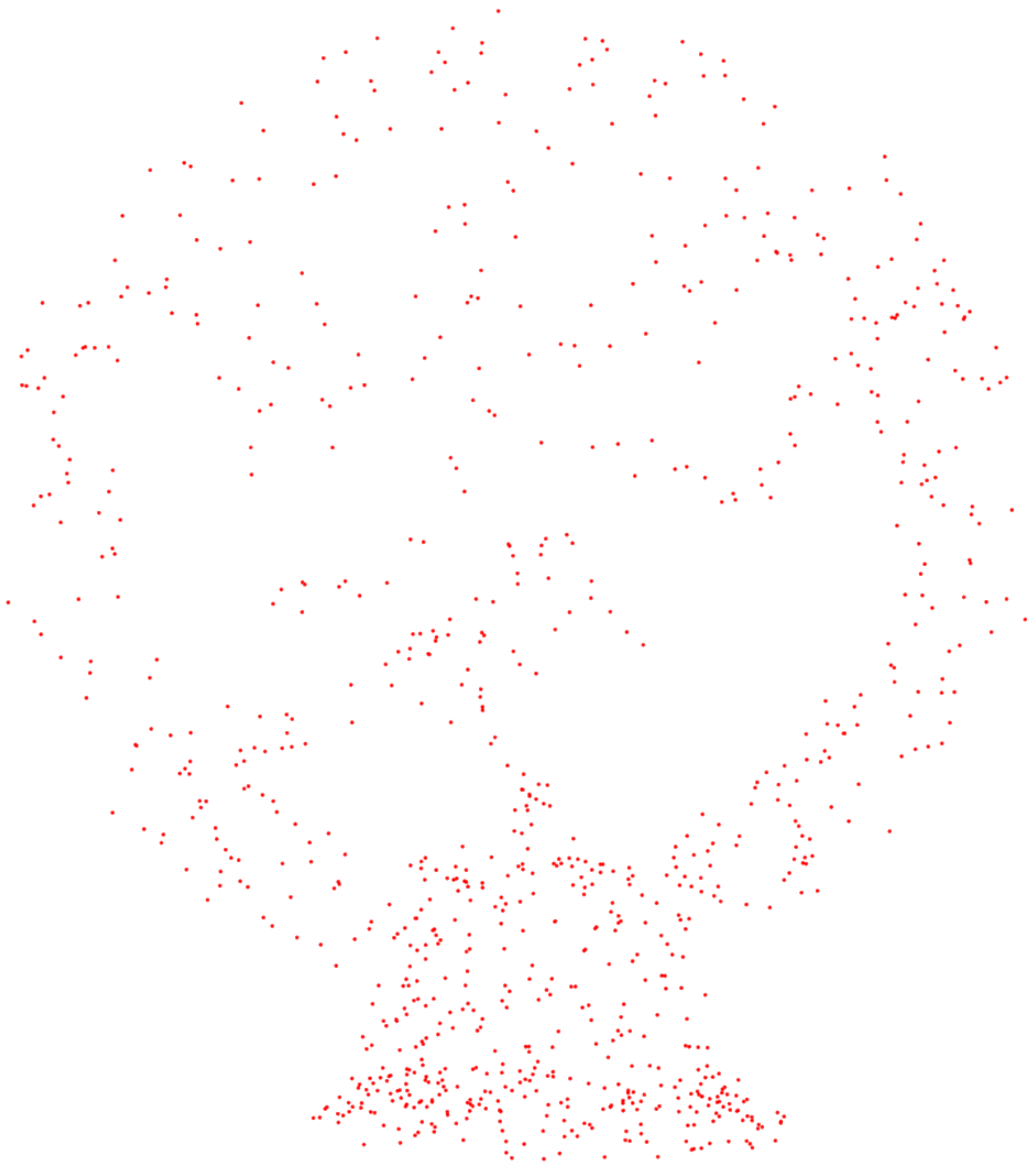}
	}
	\vskip -0.2cm
	\caption{Resampling of points with high response intensity.}
	
	\label{fig:resample}
	\vskip -0.3cm
\end{figure}

{After attaining the \revise{response intensity} of each point and the elimination of outliers, we resample salient points that have high response intensity and are more discriminative of the structure feature than individual points}. Using \revise{$\mathcal{I}(\bm{x})$} to resample salient points not only enables the \emph{structure similarity} of two point clouds but also reduces the computational complexity, which is of practical \revise{interest}. Fig.~\ref{fig:resample} presents two resampled point clouds with the model from~\cite{DataSJ}.

\revise{Subsequently}, we further explore the involvement of the \emph{geometric features} of these high response intensity points by GSP. Suppose the graph signal $\revise{\bm{s}}$ at $\bm{x}_i\in \mathcal{G}_{\mathcal{X}}$ is a vectorized normal and curvature $\revise{\bm{s}(\bm{x}_i)}=(n_{ix}, n_{iy}, n_{iz}, c_i)\in \mathbb{R}^4$, where $(n_{ix}, n_{iy}, n_{iz})$ and $c_i$ are the normal and curvature of $\bm{x}_i$, respectively. \revise{Then}, the \textit{graph gradient} of \revise{$\bm{s}$} at $\bm{x}_i$ is defined as

\begin{small}
	\begin{eqnarray}
		\centering
		\nabla_{\bm{x}_i}\revise{\bm{s}}=\sum_{e_{ij} \in \mathcal{E}}\frac{\partial{\revise{\bm{s}}}}{\partial{e_{ij}}}(\bm{x}_i)=\sum_{\bm{x}_j \in \mathcal{N}_{i}}\sqrt{\mathbf{W}_{ij}}(\revise{\bm{s}}(\bm{x}_i)-\revise{\bm{s}}(\bm{x}_j)),
		\label{eq:gradient}
	\end{eqnarray}
\end{small}where $\mathcal{E}$ is the set of edges connecting with $\bm{x}_i$. 
Eq.~\ref{eq:gradient} reflects the variation of the local surface geometry in the neighborhood $\mathcal{N}_{i}=\{\bm{x}_{j}\in \mathcal{X}|\mathbf{W}_{ij}\neq0\}$,  
inducing the graph Laplacian matrix $\mathbf{L}=\mathbf{D}-\mathbf{W}$, which is available to create another geometric invariant $\mathcal{V}_g(\bm{x}_i)$. 
\begin{theorem}
	\label{theo:normal}
	The variation of local surface geometry $\mathcal{V}_g(\bm{x}_i)$ with respect to the normals and curvatures is defined as  
	\begin{small}
		\begin{eqnarray}
			\mathcal{V}_g(\bm{x}_i)=\sum_{e_{ij} \in \mathcal{E}}(\frac{\partial{\revise{\bm{s}}}}{\partial{e_{ij}}}(\bm{x}_i))^2=\sum_{\bm{x}_j \in \mathcal{N}_{i}}\mathbf{W}_{ij}\|\revise{\bm{s}}(\bm{x}_i)-\revise{\bm{s}}(\bm{x}_j)\|_2^2,
		\end{eqnarray}
	\end{small}which is invariant under the rigid translation of point clouds.
\end{theorem}

\begin{figure}[t]
	\centering
	\includegraphics[width=1\linewidth]{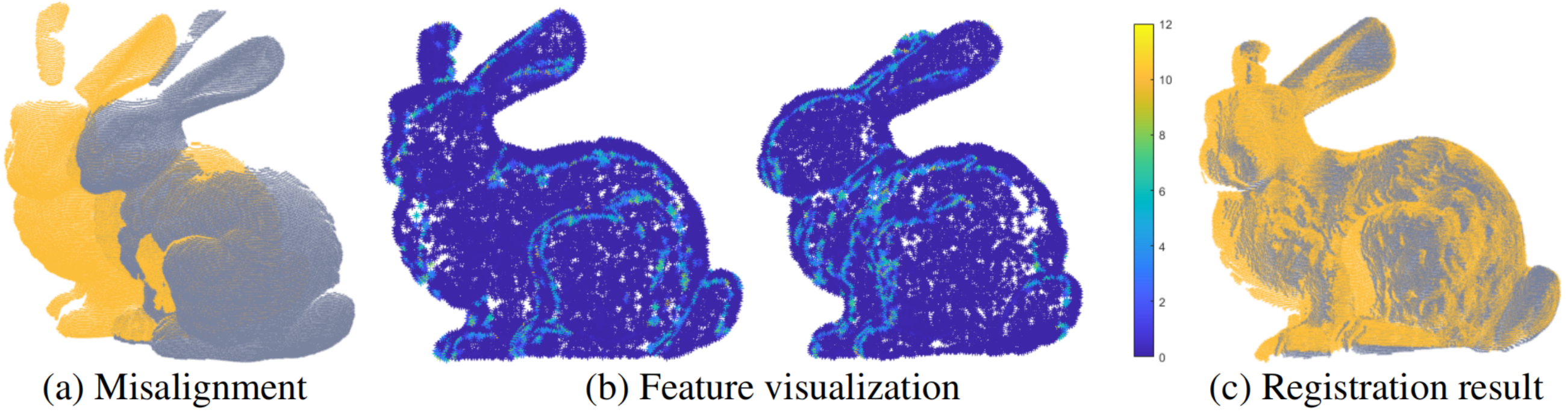}
	\vskip -0.2cm
	\caption{(a) Misalignment of two partial overlapping Bunny data~\cite{DataStanford}; (b) The geometric feature $\mathcal{V}_g(\bm{x})$ encoded by color is invariant under rigid transformations; (c) The registration result by the proposed method.}
	\vskip -0.3cm
	\label{fig:featVis}
\end{figure}

After graph-based resampling, we endow each point $\bm{x}_i$ with a {response intensity $\mathcal{I}(\bm{x}_{i})$} and a geometric invariant $\mathcal{V}_g(\bm{x}_i)$; both are invariant under rigid transformations. If we take \revise{$\mathcal{I}(\bm{x}_{i})$} as a lower-order statistic of the feature variation in the point level, then $\mathcal{V}_g(\bm{x}_i)$ describes the higher-order variation of the local surface geometry. Fig.~\ref{fig:featVis} 
visualizes the geometric invariant $\mathcal{V}_g(\bm{x})$ of two perspective point clouds via different colors, from which we observe that the similar geometry topology, such as the {ear region}, is with the similar $\mathcal{V}_g(\bm{x})$; hence, $\mathcal{V}_g(\bm{x})$ is an effective signature to improve the correspondence precision among pairwise points of two scans.

\subsection{Dynamical Registration} We use rigid-body dynamics and integrate the proposed  {point response intensity} with geometric invariant for point cloud registration. We take points of the source $\mathcal{X}=\{\bm{x}_i\in \mathbb{R}^3\}_{i=1}^M$ and the target $\mathcal{Y}=\{\bm{y}_j\in \mathbb{R}^3\}_{j=1}^N$ as particles, and then a rotation matrix ${\bf{R}}\in SO(3)$ and a translation vector $\bf{t}\in \mathbb{R}^3$ exist to align these two \textit{particle systems} optimally. Each particle is characterized by its \emph{mass} $m_i$ and \emph{location} $\mathbf{r}_i$ ($\bm{r}_i=\bm{x}_i~\text{or}~\bm{y}_i$). The force $\mathbf{f}_i$ acting on it is composed of an internal force $\bm{f}_{i}^{in}$ and an external force $\mathbf{f}_{i}^{out}$: $\bm{f}_i=\bm{f}_{i}^{in}+\bm{f}_{i}^{out}=\sum_{j=1}^{M}\bm{f}_{ij}^{in}+\bm{f}_{i}^{out}$, where $\bm{f}_{ij}^{in}$ is the interaction force between the $j^{th}$ and the $i^{th}$ particles in the same particle system, and $\bm{f}_{i}^{out}$ is from the outside. According to  \emph{Newton's third law}~\cite{dreizler2010theoretical} (or from the perspective of \emph{the law of momentum conservation}), the summation of all inner forces becomes 
\begin{eqnarray}
	\sum_{i=1}^{M}\sum_{j=1}^{M}\bm{f}_{ij}^{in}=\frac{1}{2}\sum_{i=1}^{M}\sum_{j=1}^{M}(\bm{f}_{ij}^{in}+\bm{f}_{ji}^{in})=0,
\end{eqnarray}
as $\bm{f}_{ij}^{in}=-\bm{f}_{ji}^{in}$. \revise{Therefore}, the total force $\bm{f}$ exerting on the whole particle system is $\bm{f}=\sum_{i=1}^{M}\bm{f}_{i}^{out}$. \revise{In contrast to} previous approaches, such as~\cite{jauer2018efficient} that gives equal treatment of each particle and merely set their mass $m_i=1$, we use the {point response intensity $\mathcal{I}(\bm{x})$}, a natural reflection of the distribution importance, to weigh diverse particles, \ie, $m_i=\revise{\mathcal{I}(\bm{x}_{i})}$. \revise{Hence}, the center of mass is defined as $\bm{c}=\frac{1}{G}\sum_{i=1}^{M}\revise{\mathcal{I}(\bm{x}_{i})} \cdot \bm{r}_i$, where $G=\sum_{i=1}^{M}\revise{\mathcal{I}(\bm{x}_{i})}$.

\subsection{Particle Motion} From  \emph{Newton's second law}~\cite{dreizler2010theoretical}, the state change of the particle system, such as rotation and translation, is due to the external force $\bm{f}=\bm{a}\cdot G$, where $\bm{a}$ is the linear acceleration. In light of the kinematic equation from rigid-body dynamics, the displacement $\revise{\bm{\phi}(t)}$ with respect to time $t$ is 
\begin{eqnarray}
	\bm{\phi}(t)=\bm{\phi}(0)+\dot{\bm{\phi}}(0)\cdot t+\frac{1}{2}\bm{a}\cdot t^2,
	\label{eq:displacement}
\end{eqnarray}
where $\dot{\bm{\phi}}(0)=\frac{d\bm{\phi}}{dt}|_{t=0}$ is the initial velocity. Similarly, the rotational displacement $\bm{\sigma}(t)$ is defined as
\begin{eqnarray}
	\bm{\sigma}(t)=\bm{\sigma}(0)+\dot{\bm{\sigma}}(0)\cdot t+\frac{1}{2}\bm{\alpha} \cdot t^2,
	\label{eq:rotation}
\end{eqnarray}
where $\dot{\bm{\sigma}}(0)=\frac{d\bm{\sigma}}{dt}|_{t=0}$ is the initial angular velocity, and $\bm{\alpha}=\ddot{\bm{\sigma}}(0)=\frac{d^2\bm{\sigma}}{dt^2}|_{t=0}$ is the angular acceleration. \revise{Given that} 
the source point cloud $\mathcal{X}$ is static at the beginning, we initialize $\{\bm{\phi}(0), \dot{\bm{\phi}}(0), \bm{\sigma}(0), \dot{\bm{\sigma}}(0)\}=0$. According to the \emph{theorem of angular momentum} and the \emph{law of rotation}~\cite{dreizler2010theoretical}, we have 
\begin{eqnarray}
	\frac{d\bm{L}}{dt}=\bm{M}=J\cdot \bm{\alpha},
	\label{inertial}
\end{eqnarray} 
where $\bm{L}=\sum_{i=1}^{M}\tilde{\bm{x}}_i\times \bm{p}_i=\sum_{i=1}^{M}\tilde{\bm{x}}_i\times \revise{\mathcal{I}(\bm{x}_{i})\dot{\bm{\phi}}_i}$, $\bm{M}=\sum_{i=1}^{M}\tilde{\bm{x}}_i\times \bm{f}_i^{out}$ are the summation of the angular momentum and the torque of the particle system, respectively. $\tilde{\bm{x}}_i=\bm{x}_i-\bm{c}$ is the lever from the mass center $\bm{c}$ to the point $\bm{x}_i$, $\bm{p}_i$ is the linear momentum of a particle, and $J=\sum_{i=1}^{M}\revise{\mathcal{I}}(\bm{x}_{i})\|\tilde{\bm{
		x}}_i\|^2$ is the inertial moment of the system. \revise{Considering} $\bm{L}=J\cdot \int \bm{\alpha}dt$, the rotational axis of the particle system is consistent with $\bm{L}$, thus we take $\frac{\bm{L}}{\|\bm{L}\|_2}$ as the unit rotation axis.

\subsection{Conversion of Vectors to Rotation Matrices}
In rigid-body mechanics, the rotation of particle systems is typically represented by a vector with three parameters (\textit{Euler's theorem}~\cite{tomasi2013vector}), \ie, a unit rotation axis $\bm{l}$ and an angle $\theta$ around it. {Therefore, we utilize the \textit{Rodrigues formula}~\cite{kanatani2015understanding} to convert the rotation vector to a rotation matrix for convenient representation of the point cloud registration process.}

To ease the computation of the above particle motion, similar to~\cite{jauer2018efficient}, we assume stepwise motion; hence, the time interval is set as $\Delta t=1$. Therefore, Eq.~\ref{eq:displacement} and \ref{eq:rotation} are simplified as $\bm{\phi}=\frac{1}{2}\bm{a}, ~\bm{\sigma}=\frac{1}{2}\bm{\alpha}, ~\text{and} ~\bm{\alpha}=\bm{L}/J.$ For each iteration $k$, the source point cloud $\mathcal{X}$ moves \revise{toward} the target $\mathcal{Y}$ under the external force induced by $\mathcal{X}$ and $\mathcal{Y}$, and the transformation is
\begin{eqnarray}
	\mathcal{T}_k=
	\begin{bmatrix}
		\bf{R}_{\theta} & \bf{t}\\
		\bf{0}^T &1
	\end{bmatrix}\in SE(3),
\label{eq:transformation}
\end{eqnarray}
where $\bf{R}_{\theta}$ is the rotation matrix with  $\theta=|\bm{\sigma}|$, $\bf{t}=\bm{\phi}$ is the translation vector, {and $SE(3)$ is the \emph{special Euclidean group} defined as 
\begin{eqnarray*}
SE(3)=\{\mathcal{T}=
\begin{bmatrix}
	\bf{R} & \bf{t}\\
	\bf{0}^T &1
\end{bmatrix}\in \mathbb{R}^{4\times 4}|{\bf{R}}\in SO(3), \bf{t}\in \mathbb{R}^{3}
\}.	
\end{eqnarray*}}
\subsection{Geometry-aware Force Modeling} 
To align $\mathcal{X}$ to $\mathcal{Y}$, we adopt \emph{Coulomb's law}~\cite{kovacs2001coulomb} to generate the external force acting on $\mathcal{X}$. In \revise{an} electrical field, the electrostatic force $\bm{F}_{ij}$ between charged particles $Q_i$ and $Q_j$ is modeled by 
\begin{eqnarray}
	\bm{F}_{ij}=k_e\frac{Q_i\cdot Q_j}{\|Q_i-Q_j\|_2^2} \cdot \frac{Q_i-Q_j}{\|Q_i-Q_j\|_2},
\end{eqnarray}
where the constant $k_e$ is the electric force constant, and $Q_i-Q_j$ is a vector pointing from $Q_j$ to $Q_i$. Analogies to Coulomb's law, \ie, attraction between the opposite sign charges and repulsion between the same sign charges, we leverage the geometric invariant $\mathcal{V}_g(\bm{x})$ to guide the generation of the outer force.

Given the invariants $\mathcal{V}_g(\mathcal{X})=\{\mathcal{V}_g(\bm{x}_1), \cdots, \mathcal{V}_g(\bm{x}_M)\}$ and $\mathcal{V}_g(\mathcal{Y})=\{\mathcal{V}_g(\bm{y}_1), \cdots, \mathcal{V}_g(\bm{y}_N)\}$ of two 3D scans of the same object,  we adopt the nearest neighbor searching (accelerated by \emph{kDtree}~\cite{bentley1975multidimensional}) in the \emph{feature space} to generate a set of deterministic correspondences \revise{because they are invariant under rigid transformations}: 

{\begin{footnotesize}
	\begin{eqnarray}
		\mathcal{C}=\{(\bm{x}_i,\bm{y}_j)\in\mathcal{X}\times\mathcal{Y}, i\in[1, M]|\min_{j=1,\cdots, N} \|\mathcal{V}_g(\bm{x}_i)-\mathcal{V}_g(\bm{y}_j)\|_2 \}.
	\end{eqnarray} 
\end{footnotesize}}Then the total outer force is defined as
\begin{eqnarray}
	\bm{f}=\sum_{(\bm{x}_i,\bm{y}_j)\in\mathcal{C}}k_{ij}\frac{\bm{y}_j-\bm{x}_i}{\|\bm{y}_i-\bm{x}_i\|_2^3}.
	\label{eq:force}
\end{eqnarray}
Here we use the \textit{normalized Gaussian kernel}~\cite{friedman2017elements} $\frac{1}{\sqrt{2\pi }}exp(-\frac{\|x-y\|_2^2}{2})$ to weigh different geometric \revise{similarities}, \ie, the correspondence of pairwise particles is soft with probability. To generate repulsive forces, we let 
\begin{eqnarray}
	k_{ij}=\frac{\sqrt{2}}{\pi}exp(-\frac{\|\mathcal{V}_g(\bm{x}_i)-\mathcal{V}_g(\bm{y}_j)\|_2^2}{2})-1\in [-1,1].
\end{eqnarray}
Therefore, $k_{ij}$ not only encourages pairwise points with similar geometric features but also penalizes  different pairs, thus enabling more valid motions of $\mathcal{X}$.

\subsection{Fast Transformation Solving} \revise{Given} the stepwise motion, we introduce an \emph{adaptive simulated annealing} (ASA) framework~\cite{cicirello2007design} with cool-down rate to regulate the registration process and solve for the globally optimal transformation. ASA can be \revise{considered} as a variant of the gradient descent search paradigm \revise{because} it incorporates \emph{uncertainties} during the state transition process. \revise{In contrast to} ordinary simulated annealing (SA), ASA approximates the original \textit{thermal equilibrium} of SA by the \emph{D-equilibrium} to accelerate the algorithm convergence, thereby \revise{considerably} reducing the time consumption for point cloud registration. Moreover, ASA is almost free to tune parameters because a feedback mechanism based on the {\emph{AcceptRate} and \emph{LamRate}} is adopted to adjust the annealing temperature adaptively, rather than setting a fixed cooling schedule as implemented in SA. As in~\cite{jauer2018efficient}, we take the \emph{kinetic energy} $E_k$ as the objective function to evaluate the registration quality of each iteration
\begin{eqnarray}
	E=\frac{1}{2}G\dot{\bm{x}}^2+\frac{1}{2}J\dot{\bm{\sigma}}^2.
	\label{eq:energy}
\end{eqnarray}
The \textit{acceptance probability} $P$ in ASA is defined as 
\begin{align}
	P(\Delta E)&=\frac{exp^{(-E_{k+1}/T_k)}}{exp^{(-E_{k+1}/T_k)}+exp^{(-E_{k}/T_k)}}\\
	&=\frac{1}{exp(\Delta E/T_k)+1}\approx exp(-\frac{\Delta E}{T_k}),
\end{align}	
which essentially satisfies the \emph{Boltzmann--Gibbs distribution}~\cite{jauer2018efficient}. {$T_k$ is the current system temperature, which is typically initialized as 1, and then updated via the composite operations with the cool down rate $c$.} $\Delta E=E_{k+1}-E_k$ is the energy difference between consecutive states. \revise{We summarize the ASA process and our complete algorithm in Alg.~\ref{alg:ASA} and Alg.~\ref{alg:pipeline} separately.} Interested readers can refer to ~\cite{ingberadaptive} for \revise{further} analysis of the simulated annealing framework.

\begin{algorithm}
	\revise{\caption{Adaptive Simulated Annealing Process}\label{alg:ASA}
	\KwIn{Current temperature $T_k$, $AcceptRate$, energy $E_{k}$, $E_{k-1}$, and cool-down rate $\beta\in [0.7, 0.9]$.\\ }
	\KwOut{$T_{k+1}$, $AcceptRate$, $state$.\\}
	$state=0, \Delta E=E_{k}-E_{k-1}$.\\
	\eIf{$\Delta E<0$}{$AcceptRate = 1/500\cdot(499\cdot AcceptRate + 1)$.}{
	$P(\Delta E) = \min\{1, \exp^{(-\Delta E / T_k)}\}$.\\
	\eIf{$rand(0,1)<P(\Delta E)$}{$AcceptRate = 1/500\cdot(499\cdot AcceptRate + 1)$.}{$state=1$\;
		$AcceptRate = 1/500\cdot(499\cdot AcceptRate)$.}
}
    \eIf{$k/maxNumIter<0.15$}{$LamRate = 0.44 + 0.56\cdot 560^{(-1\cdot k/(maxNumIter\cdot 0.15))}$.}{
    \eIf{$k/maxNumIter<0.65$}{$LamRate = 0.44$.}{$LamRate = 0.44\cdot440^{- (((k/{maxNumIter})-0.65)/0.35)}$.}
}
\eIf{$AcceptRate > LamRate$}{$T_{k+1} = \beta\cdot T_k$.}{$T_{k+1} = T_k/\beta$.}}
\end{algorithm}

\begin{algorithm}
	\revise{\caption{The Proposed GraphReg Process}\label{alg:pipeline}
		\KwIn{Two point clouds with source $\mathcal{X}$ and target $\mathcal{Y}$.\\}
		\KwOut{Rigid transformation $\hat{\mathcal{T}}$.\\}
		Graph construction: 
		$\mathcal{G}_{\mathcal{X}}$, $\mathcal{G}_{\mathcal{Y}}$, the degree matrix $\mathbf{D}$, and the adjacency matrix $\mathbf{W}$.\\
		Point response intensity computation: $\mathbf{A}=\mathbf{D}^{-1}\mathbf{W}$, $h_{Har}(\mathbf{A})=\mathbf{I}-\mathbf{A}$,  $\mathcal{I}(x_i)=\|(h_{Har}(\mathbf{A}) \cdot \mathbf{X})_{i}\|_2^{2}$.\\
		Outlier removal based on the point response intensity:\\ {\footnotesize{$\text{MAD}(\revise{\mathcal{I}(\mathbf{X}))}=\mathop{med}_{i=1,\cdots, M}\{|\revise{\mathcal{I}(\bm{x}_i)}-\mathop{med}_{j=1,\cdots, M}(\revise{\mathcal{I}(\bm{x}_j))}|\}$}};\
		{$\mathcal{I}(\bm{x}_i)>\alpha \cdot \text{MAD}(\mathcal{I}(\mathbf{X}))$}.\\
		Resampling by $\mathcal{I}(\mathbf{X})$. Geometric invariant extraction: \\
		${\bm{s}(\bm{x}_i)}=(n_{ix}, n_{iy}, n_{iz}, c_i)$\;
		{\scriptsize{$\mathcal{V}_g(\bm{x}_i)=\sum_{e_{ij} \in \mathcal{E}}(\frac{\partial{\revise{\bm{s}}}}{\partial{e_{ij}}}(\bm{x}_i))^2=\sum_{\bm{x}_j \in\mathcal{N}_{i}}\mathbf{W}_{ij}\|\revise{\bm{s}}(\bm{x}_i)-\revise{\bm{s}}(\bm{x}_j)\|_2^2$}}.\\
		Geometry-aware force modeling: \\
		{\scriptsize{$\mathcal{C}\!=\!\{(\bm{x}_i,\bm{y}_j)\in\mathcal{X}\!\times\!\mathcal{Y}, i\in[1, M]|\min_{j\!=\!1,\cdots, N} \|\mathcal{V}_g(\bm{x}_i)\!-\!\mathcal{V}_g(\bm{y}_j)\|_2 \}$}}\;
		$k_{ij}=\frac{\sqrt{2}}{\pi}exp(-\frac{\|\mathcal{V}_g(\bm{x}_i)-\mathcal{V}_g(\bm{y}_j)\|_2^2}{2})-1\in [-1,1]$\;
		$\bm{f}=\sum_{(\bm{x}_i,\bm{y}_j)\in\mathcal{C}}k_{ij}\frac{\bm{y}_j-\bm{x}_i}{\|\bm{y}_i-\bm{x}_i\|_2^3}$\;
		Initialization: $T\!=\!1$, $AcceptRate\!=\!0.5$.\\
		Fast optimization, repeat until convergence:\\
		Solve $\bf{t}_k$, $\theta_k$, $E_k$ from Eq.~\ref{eq:displacement}--\ref{eq:rotation} and Eq.~\ref{eq:energy}\;
		Update $T$, $state$, $AcceptRate$ by Algorithm \ref{alg:ASA}\;
		\If{$state$}{$\theta_{k}=\theta_{k-1}$, $\bf{t}_{k}=\bf{t}_{k-1}$, $E_{k}=E_{k-1}$\;}
		$\theta_{k}=\theta_{k}\cdot T $, $\bf{t}_{k}=\bf{t}_{k}\cdot T$, $\mathbf{X}=\mathcal{T}_k(\mathbf{X}$).\\
		Transformation composite: $\hat{\mathcal{T}}=\mathcal{T}_{\texttt{maxNumIter}}\cdots \mathcal{T}_1$.
		
		}
\end{algorithm}

\begin{figure}[t]
	\centering
	\includegraphics[width=0.49\textwidth]{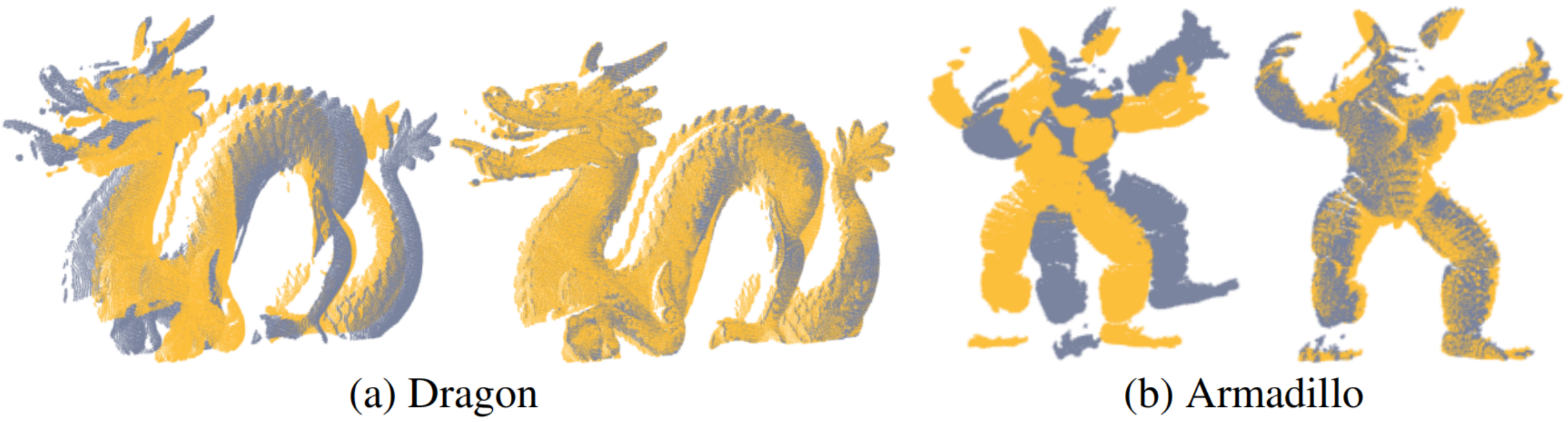}
	\vskip -0.2cm
	\caption{Misalignment of two range datasets and the registration results by the proposed method, where yellow and gray represent the target and the source point clouds, respectively.}
	\label{fig:stanford}
	\vskip -0.4cm
\end{figure}
\section{Experimental Evaluations}\label{sec:experiment}
In this section,  we perform a wide variety of experiments on benchmark 3D point cloud datasets, including \emph{range scanning} and \emph{LiDAR data}, to show the accuracy, efficiency, and robustness of the proposed method. We compare our algorithm with seven representative methods: ICP~\cite{besl1992method} optimized in MATLAB, the modified ICP comprising point-to-point (ICP-KP) and point-to-plane (ICP-KN) metrics~\cite{kjer2010evaluation}, the probabilistic registration methods including GMM~\cite{jian2010robust}, CPD~\cite{myronenko2010point}, and ECM~\cite{horaud2010rigid}, and the physics-based method PIPL~\cite{jauer2018efficient}. 
\subsection{Implementation Details}
Implementations of these compared methods are publicly available online, and their parameters are either well-tuned by ourselves or suggested by the authors. Concretely, for ICP-based methods, the \texttt{KDTreeSearcher} is used in MATLAB to speed up their convergence. The maximum iterations of ICP optimized in MATLAB and reimplemented by~\cite{kjer2010evaluation} \revise{are} set as 200 and 50, respectively. Point normals used in \textit{point2plane} minimization are found through \textit{principal component analysis} (PCA) of 4-nearest-neighborhood points. The other parameters of ICP-based methods, such as \texttt{params.Boundary} {(specifying which points are the model boundary)} and \texttt{params.Extrapolation} \revise{(if tracing out a path in the registration state space)}, are set as default. For GMM~\cite{jian2010robust}, the following values are used:\\
\texttt{config.motion='rigid3d'}, \texttt{config.normalize=1}, \texttt{config.max\_iter = 100}, \texttt{config.Lb=[-10,-10,-10,-10,-400,-400,-400]}, \texttt{config.Ub=[10,10,10,10,400,400,400]},\\
{where \texttt{normalize} means transforming both point clouds to be with zero mean and unit variance before registration; \texttt{Lb} and \texttt{Ub} are the lower and upper bounds of optimization parameters.}  
For CPD~\cite{myronenko2010point}, its parameter settings are\\
\texttt{opt.method='rigid', opt.outliers=0.5, opt.tol=1e-8, opt.normalize=1, opt.max\_it=100},\\
{here, we use the weight setting \texttt{outliers=0.5} to balance the noise and outlier distribution, and the tolerance precision \texttt{tol=1e-8} to determine the algorithm convergence.}  
For ECM~\cite{horaud2010rigid}, we use \texttt{maxNumIter=100}. For
PIPL~\cite{jauer2018efficient}, we adopt \texttt{matchingMethod='Gravitation'}, \texttt{$\beta\in[0.9, 0.98]$, Epsilon=1e-5, temperature = 1,}
{{where {$\beta$ is the cool-down rate used to regulate the system temperature}}, \texttt{Epsilon} is the lower bound of the temperature (initialized as 1) of the
simulated annealing process, and the registration terminates if \texttt{temperature} $<$ \texttt{Epsilon}. }In our proposed method, we use \texttt{maxNumIter=100} and initialize \texttt{AcceptRate = 0.5}. $\tau$ in Eq.~\ref{eq:weight} is determined by the largest Euclidean distance from the 10 nearest neighbors of each point, and the variance of graph nodes $\sigma$ in $\mathbf{W}$ satisfies $\sigma^2=\tau^2/2$. {We resample point clouds between the operations of outlier removal and surface normal estimation via GSP. The resampling rate is typically equal to $10\%-50\%$, with an aim to make the probabilistic approaches work within acceptable time and memory consumption. We keep the resampling rate fixed for the same point cloud.} 
 
The required surface normals and curvatures are estimated by the fast local plane fitting with 10-nearest-neighborhood points~\cite{hoppe1992surface}. Our approach is implemented in MATLAB and all experiments are conducted on a desktop with AMD Core Ryzen 5 3600XT and 32 GB RAM.

\subsection{Evaluation Measure}  Following~\cite{hirose2020bayesian}, we evaluate the registration quality by assessing the angular difference in degrees between two rotation matrices:
\begin{eqnarray}
	\AngErr(\hat{\bf{R}}, \bf{R})=\frac{180^\circ}{\pi}\arccos(\frac{1}{2}\{\Tr(\hat{\bf{R}}\bf{R}^T)-1\}),
\end{eqnarray} 
where $\hat{\bf{R}}, \bf{R}\in \mathbb{R}^{3\times 3}$ are the estimated and the ground truth rotation matrices of the two point clouds, respectively;  $\Tr(\cdot)$ is the trace operator of a matrix.

\begin{table}[t]
	\centering
	\caption{Average precision (degree) and runtime (second) on the Dragon and Armadillo datasets, where \textbf{green} and \textbf{red} indicate the first and the second highest accuracy, respectively.}
	\vskip -0.2cm
	\renewcommand\arraystretch{1.2}
	\setlength{\tabcolsep}{2mm}{
		\begin{tabular}{|c|cc|cc|} 
			\hline
			\multirow{2}{*}{\diagbox{\textbf{Method}}{\textbf{Dataset}}} & \multicolumn{2}{c|}{Dragon} & \multicolumn{2}{c|}{Armadillo}\\ \cline{2-5}
			&AngErr ($^\circ$)&Time (s)&AngErr ($^\circ$)&Time (s) \\ \cline{1-5}
			ICP~\cite{besl1992method}&0.4870&0.1921&1.8736&0.1867\\ \cline{1-5}
			ICP-KP~\cite{kjer2010evaluation}&0.6818&1.6527&1.9022&0.9715\\ \cline{1-5}
			ICP-KN~\cite{kjer2010evaluation}&0.3438&1.7171&1.6444&1.0268\\ \cline{1-5}
			GMM~\cite{jian2010robust}&13.0921&34.2532&24.8091&23.2408\\ \cline{1-5}
			CPD~\cite{myronenko2010point}&0.6532&13.8807&0.6417&11.8751\\ \cline{1-5}
			ECM~\cite{horaud2010rigid}&\cellcolor{second}{0.3438}&15.6381&179.0493&10.8415\\ \cline{1-5}
			PIPL~\cite{jauer2018efficient}&0.4412&1.1700&\cellcolor{second}{0.3380}&1.1419\\ \cline{1-5}
			Ours&\cellcolor{best}{0.3323}&0.3070&\cellcolor{best}{0.2521}&0.3091\\ \cline{1-5}
		\end{tabular}
	}
	\label{table:stanford}
	\vskip -0.6cm
\end{table}
\begin{figure*}[t]
	\centering
	\subcaptionbox{Misalignment}{
		\includegraphics[width=0.181\textwidth]{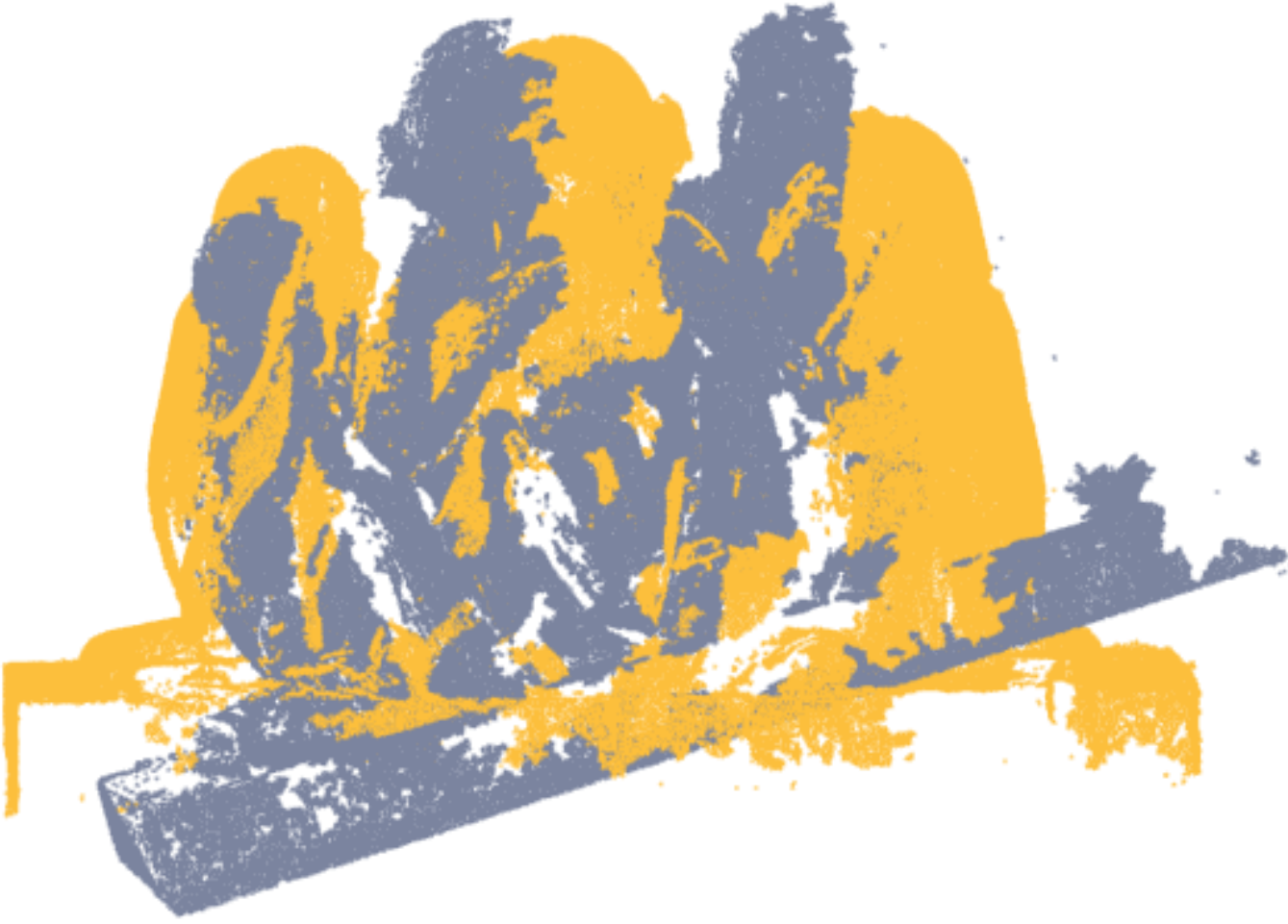}
	}
	\subcaptionbox{ICP (2.2052)}{
		\includegraphics[width=0.181\textwidth]{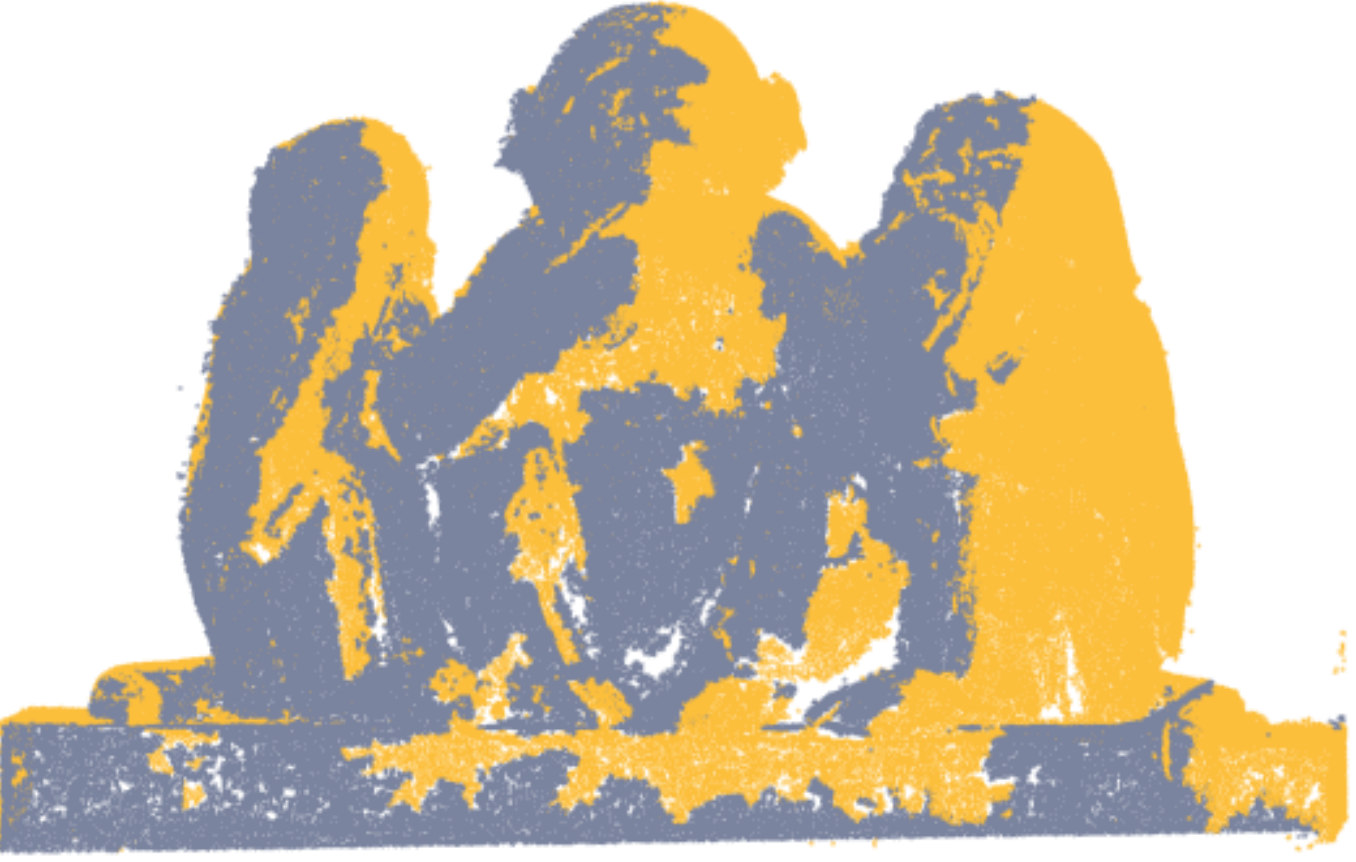}
	}
	\subcaptionbox{ICP-KP (2.2052)}{
		\includegraphics[width=0.181\textwidth]{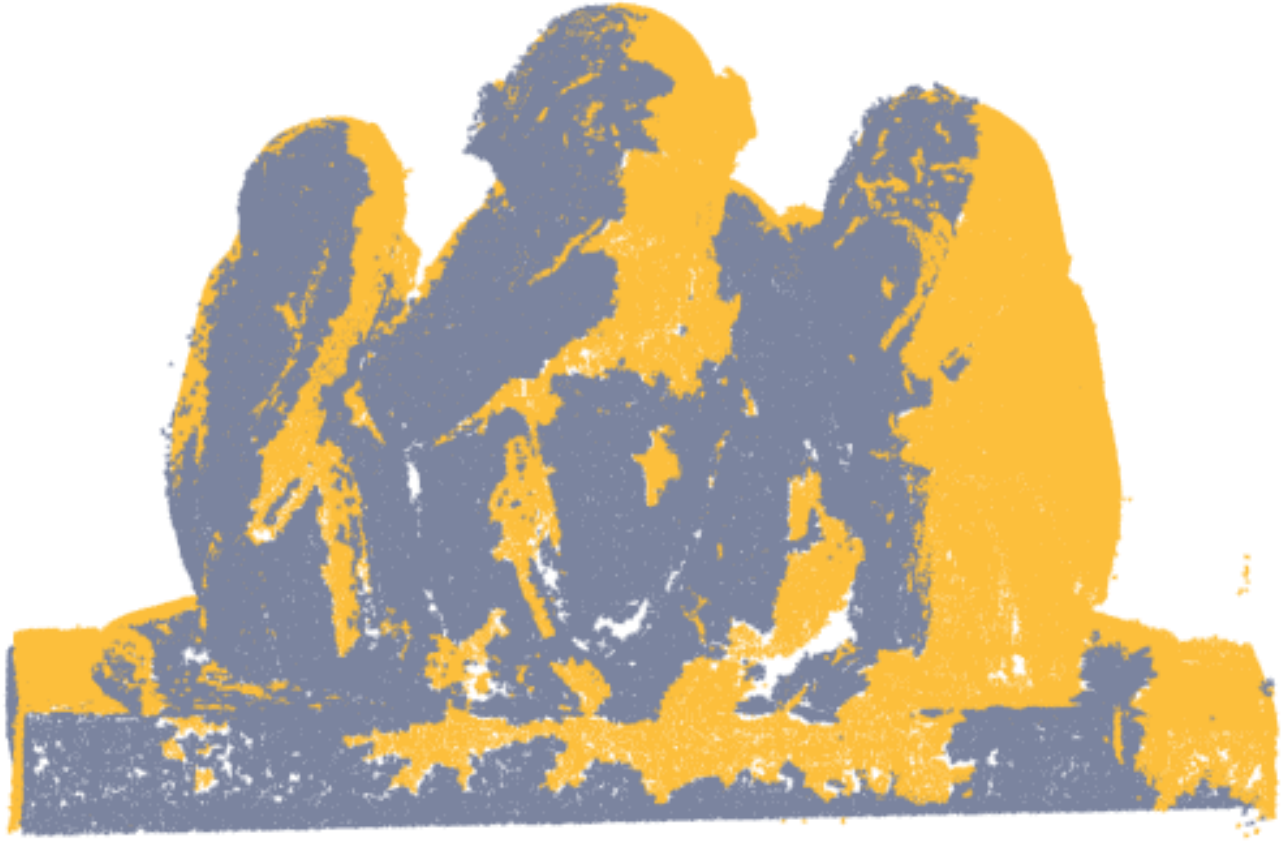}
	}
	\subcaptionbox{ICP-KN (60.4347)}{
		\includegraphics[width=0.181\textwidth]{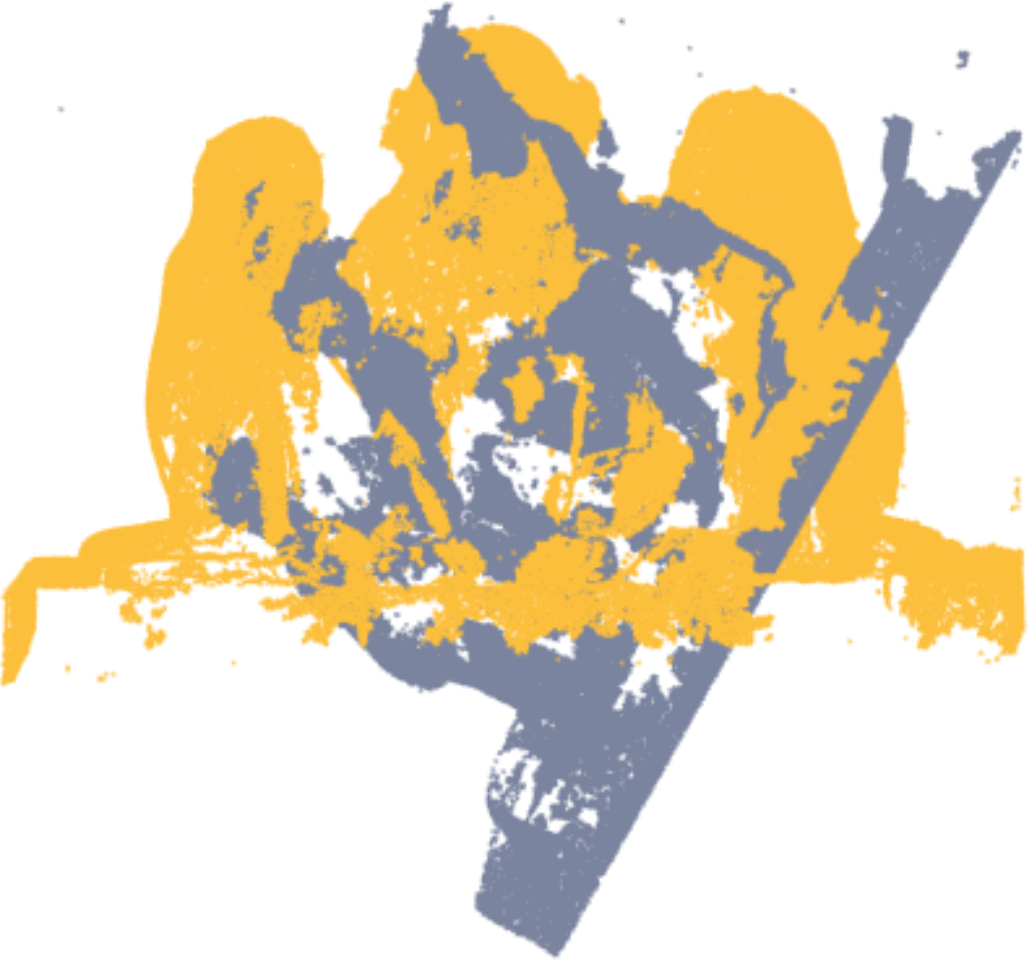}
	}
	\subcaptionbox{GMM (7.8044)}{
		\includegraphics[width=0.181\textwidth]{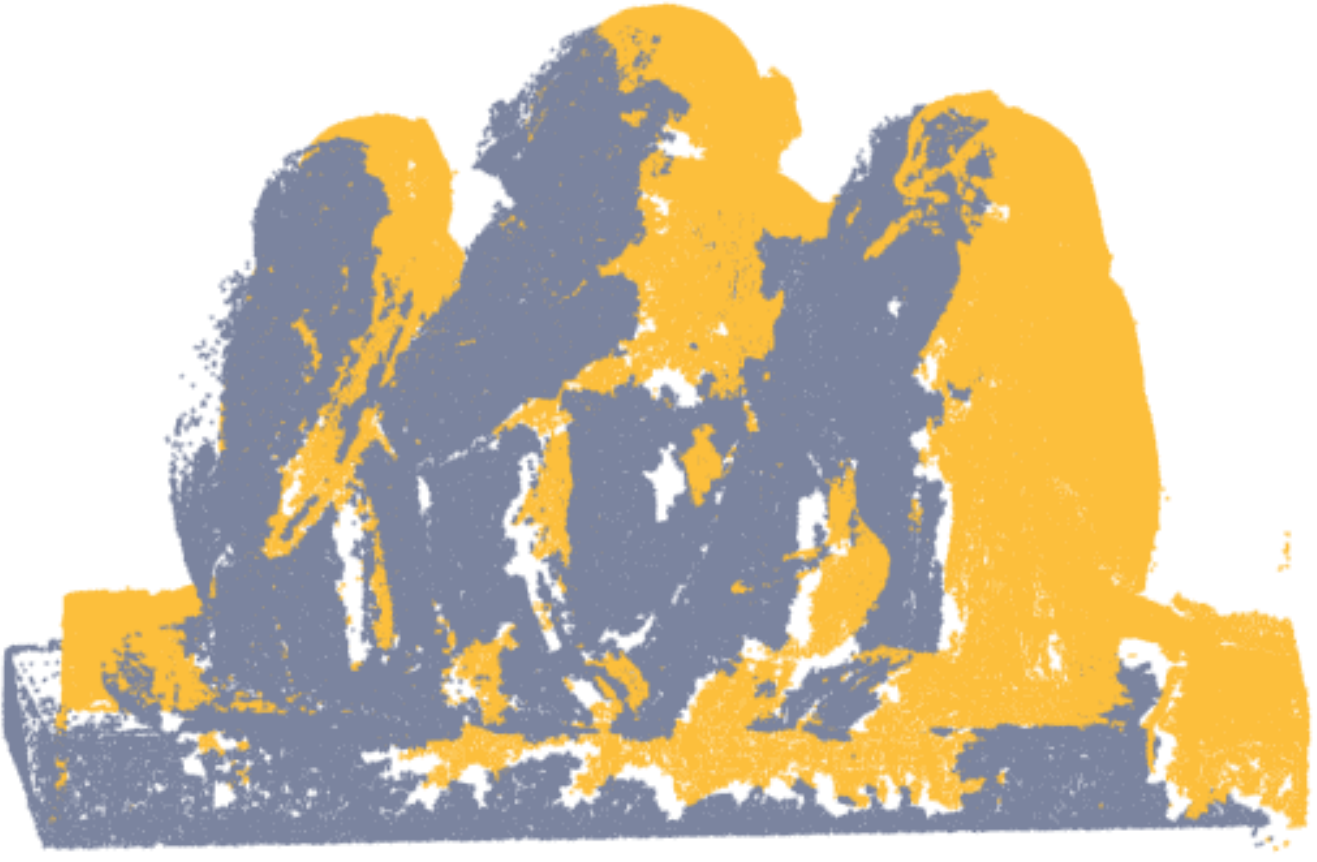}
	}
	
	\subcaptionbox{CPD (6.3719)}{
		\includegraphics[width=0.181\textwidth]{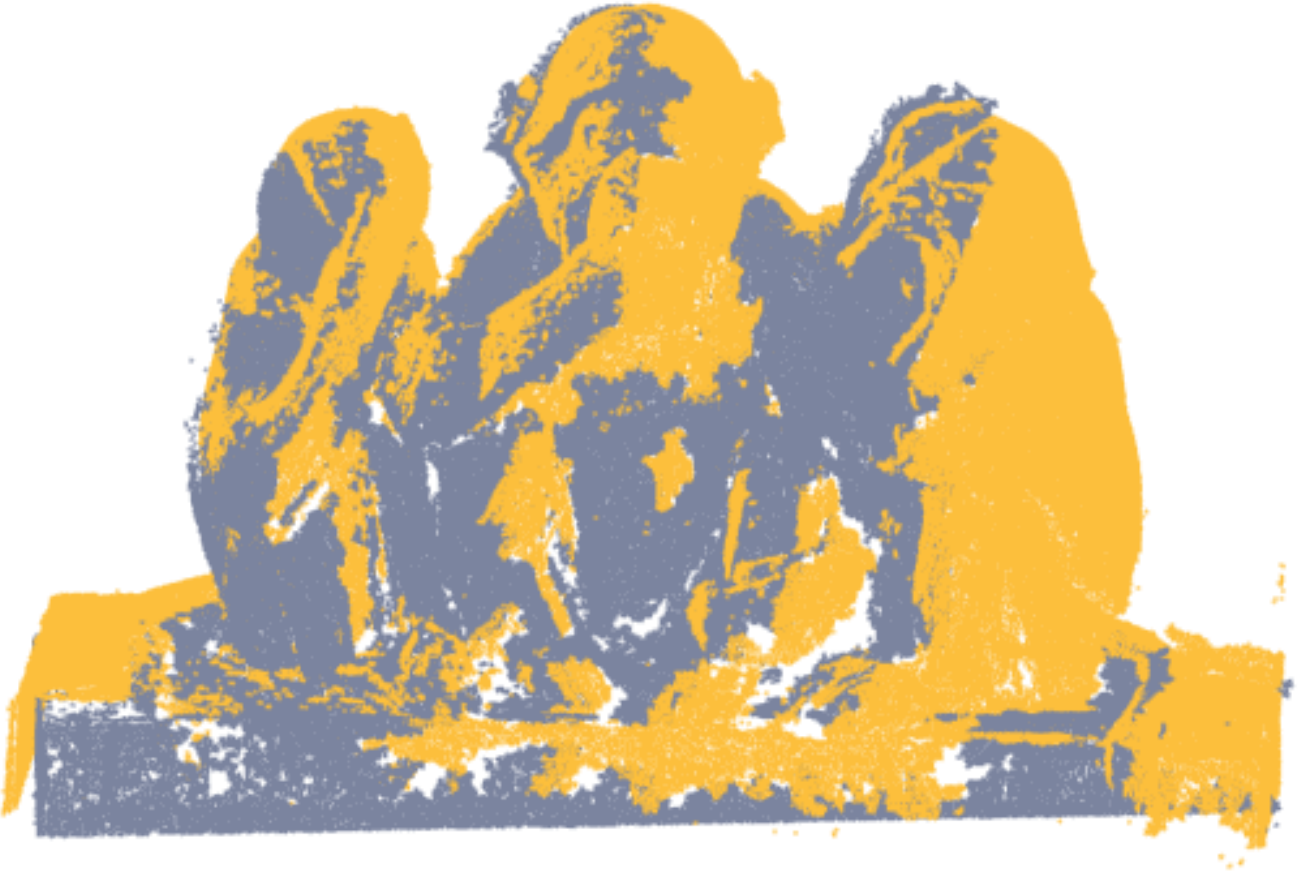}
	}
	\subcaptionbox{ECM (2.3550)}{
		\includegraphics[width=0.181\textwidth]{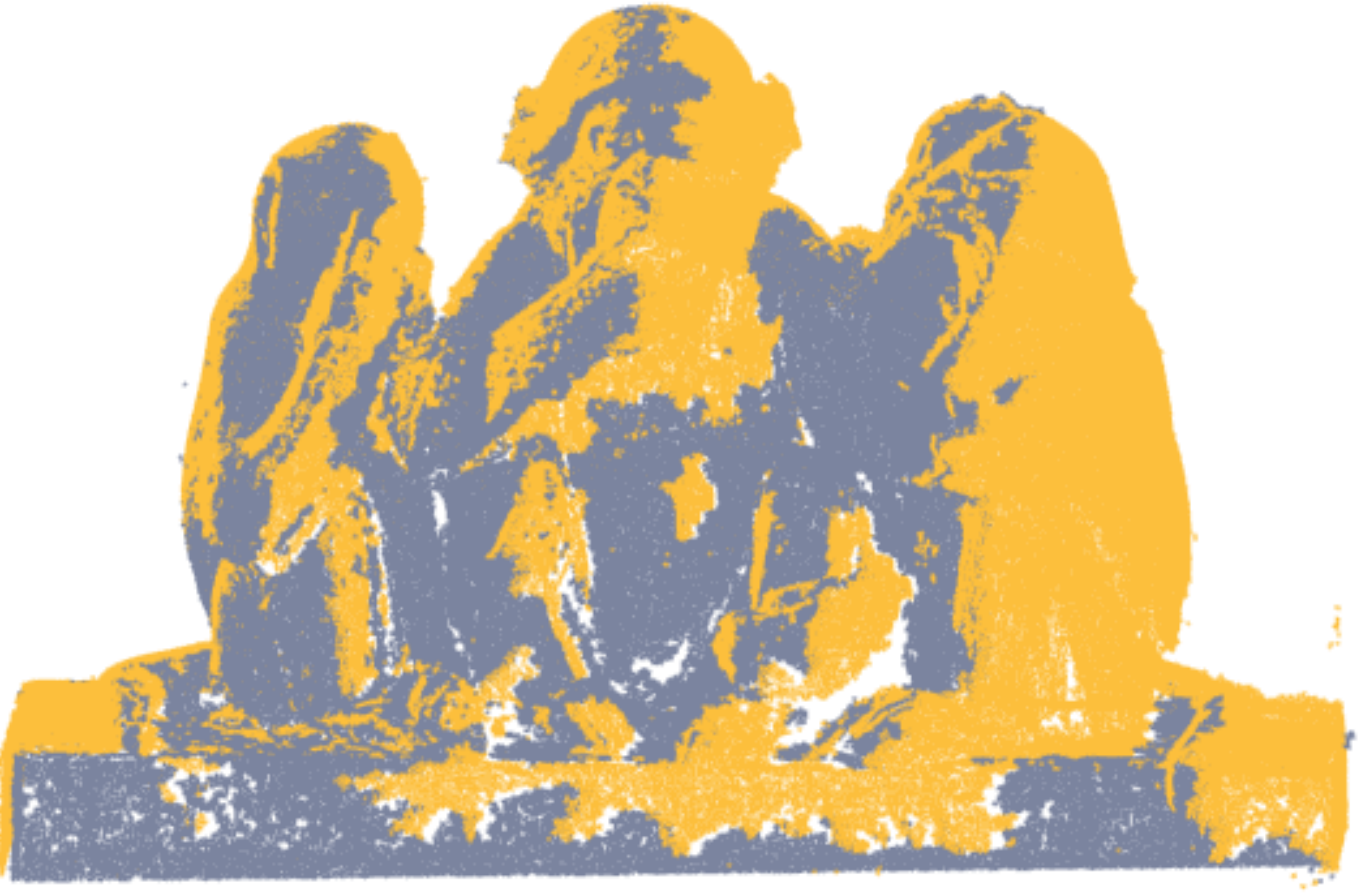}
	}
	\subcaptionbox{PIPL (16.6440)}{
		\includegraphics[width=0.181\textwidth]{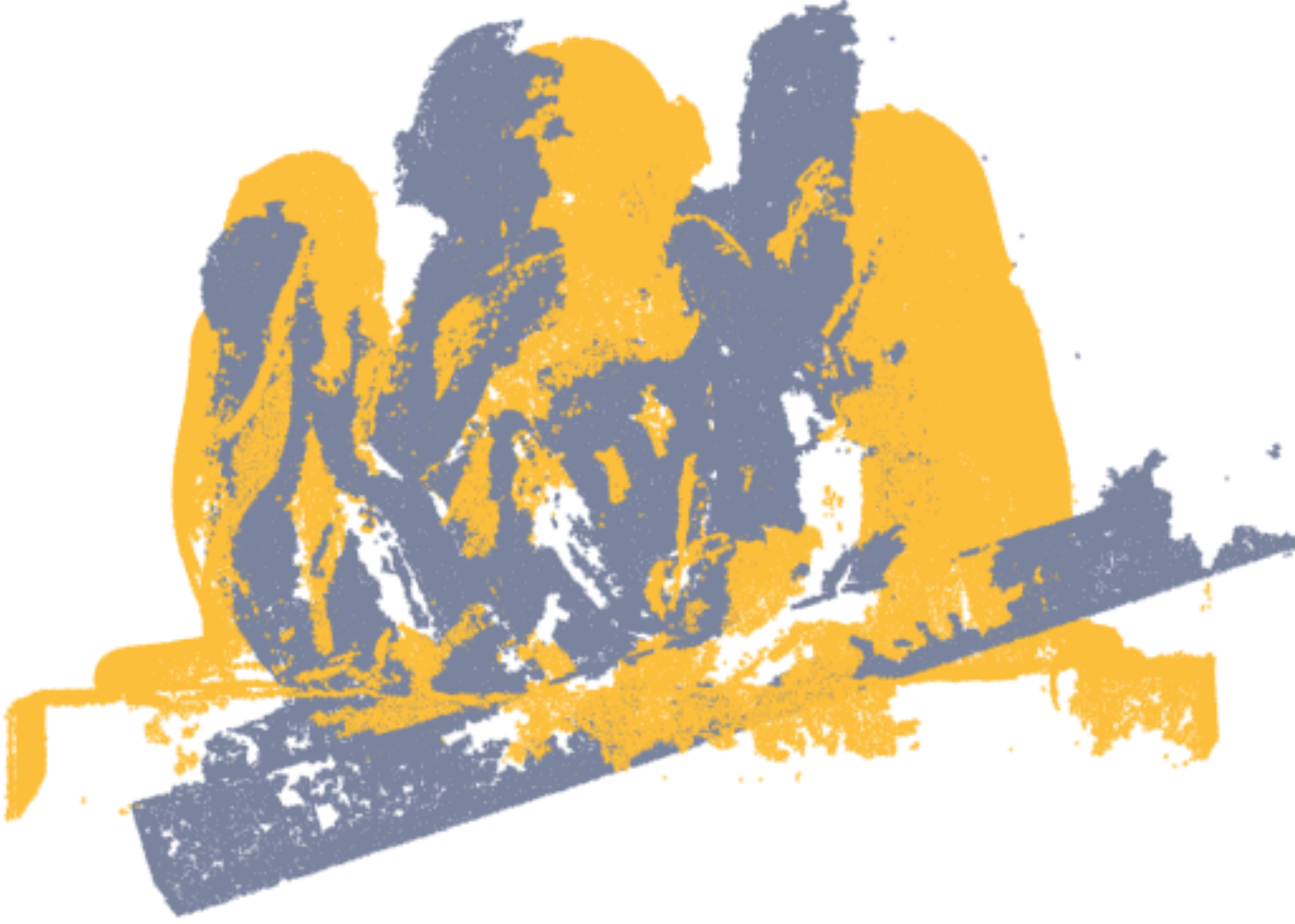}
	}
	\subcaptionbox{Ours (\textbf{1.8940})}{
		\includegraphics[width=0.181\textwidth]{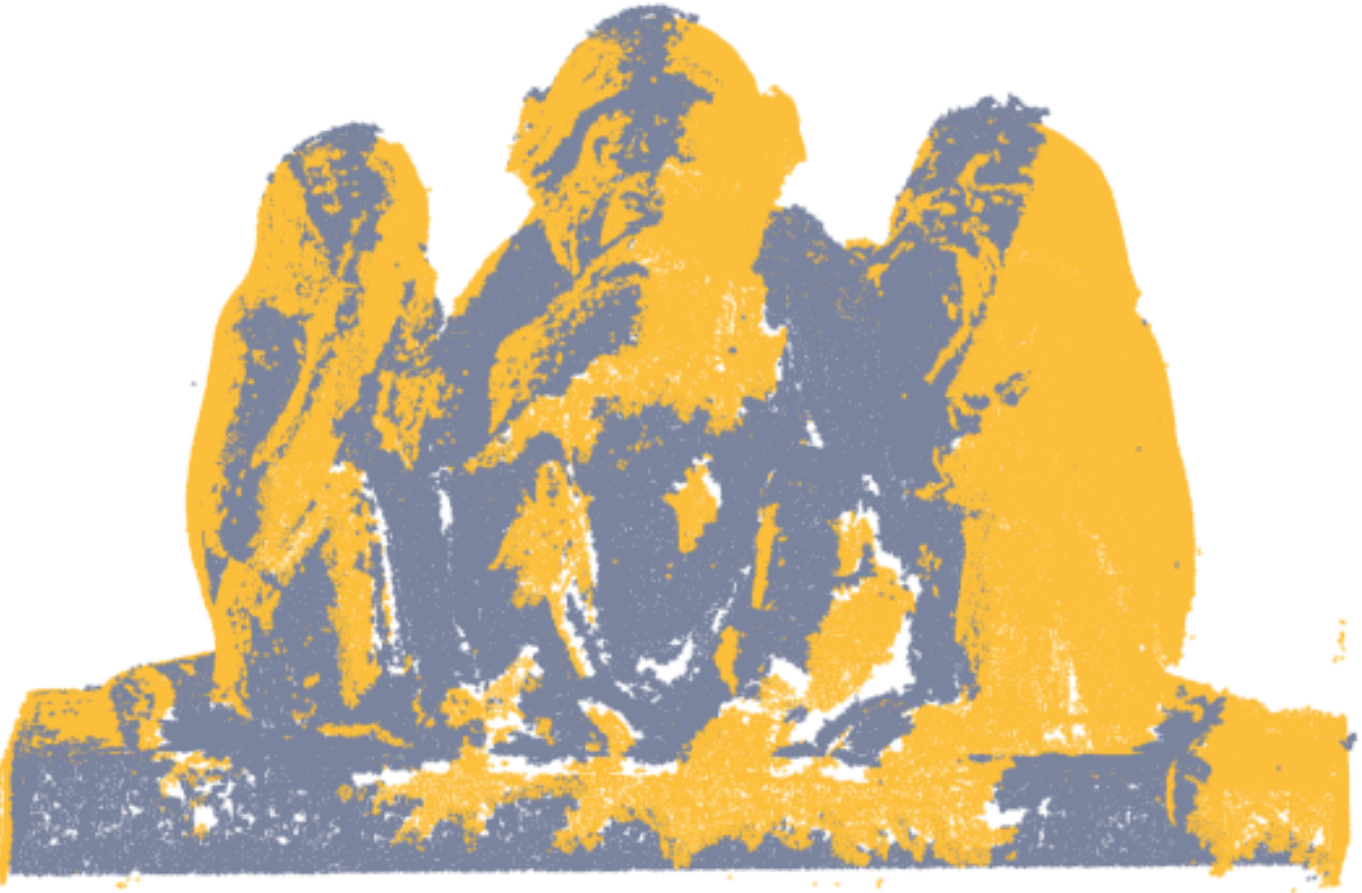}
	}
	\caption{Registration results on the Monkey dataset~\cite{DataEPFL}. Values inside the parentheses are the corresponding registration errors. 
	}
	\label{monkey_ex}
\end{figure*}
\subsection{Registration on Range Datasets}
\subsubsection{Accuracy test}
To assess the accuracy of the proposed method, we use two datasets from the \textit{Stanford 3D Scanning Repository}~\cite{DataStanford}, namely, Dragon and Armadillo, which are captured by a Cyberware 3030 MS laser scanner. 
Both datasets are scanned from two different perspectives with angle differing $24^\circ$ and $30^\circ$; \revise{therefore}, they are disparate and only have partial overlappings, as illustrated in Fig.~\ref{fig:stanford}(a) and (b). To be statistically representative, we perform 100 registrations for each dataset.

The average accuracy and time consumption are reported in Table \ref{table:stanford}. {Considering that we use the proposed resampling operation to reduce the point quantity of input point clouds for all algorithms, we mainly focus on the direct registration process after the resampling step}. From  Table \ref{table:stanford}, we conclude that the proposed method attains the highest accuracy on both datasets, and its registration speed is fast with a runtime below 1s, which is comparable to 
the optimized ICP in MATLAB. The baseline method PIPL attains satisfactory results but consumes more than triple time as ours. As for the probabilistic registration framework, except for CPD, GMM and ECM tend to get trapped in the local minima. \revise{In addition}, probabilistic approaches suffer from long registration time, \revise{whereas} ICP-based methods are more efficient. However, as angle \revise{difference} increases (such as from $24^\circ$ to $30^\circ$), indicating poor initialization and low overlapping, ICP and its variants start to show more significant deviations. \revise{By contrast}, the proposed method is relatively more stable. Fig.~\ref{fig:stanford} exhibits the registration results of our proposed method. Further tests on Monkey dataset from \emph{EPFL Statue Model Repository}~\cite{DataEPFL} is presented in Fig.~\ref{monkey_ex}. 

\begin{figure}[h]
	\vskip -0.45cm
	\centering
	
	\subcaptionbox{Initialization}{
		\includegraphics[width=0.15\textwidth]{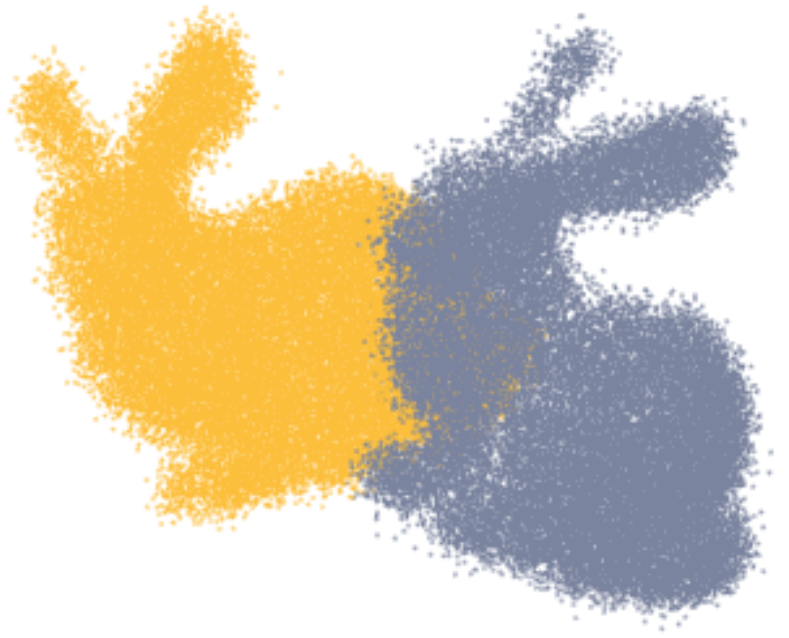}
	}
	\subcaptionbox{PIPL (6.2240)}{
		\includegraphics[width=0.14\textwidth]{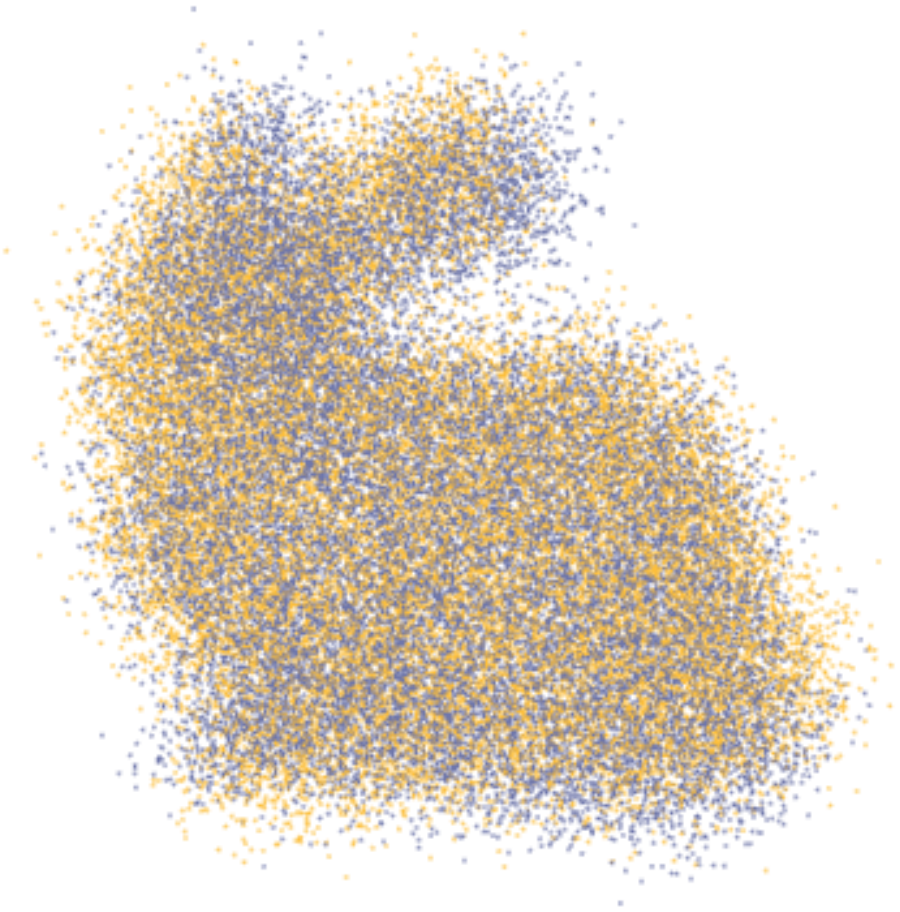}
	}
	\subcaptionbox{Ours (\textbf{1.4161})}{
		\includegraphics[width=0.145\textwidth]{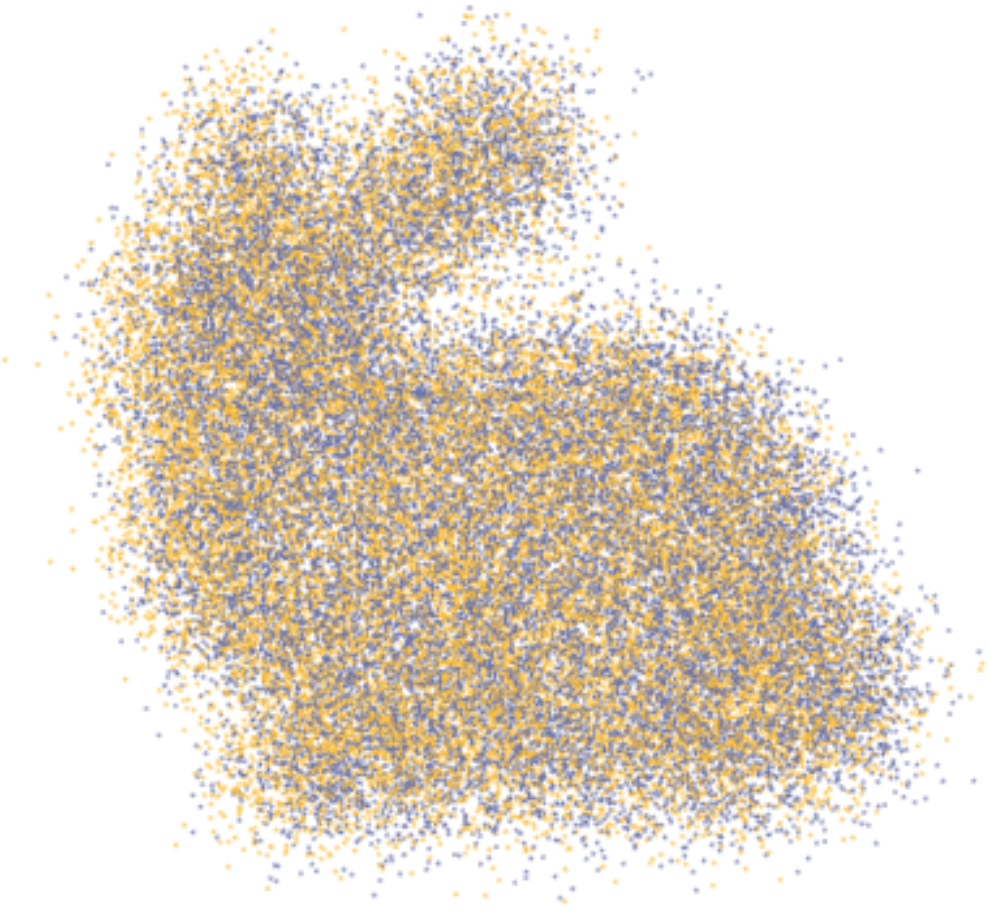}
	}
	\subcaptionbox{{\color{black}{AngErr}}}{
		\includegraphics[width=0.227\textwidth]{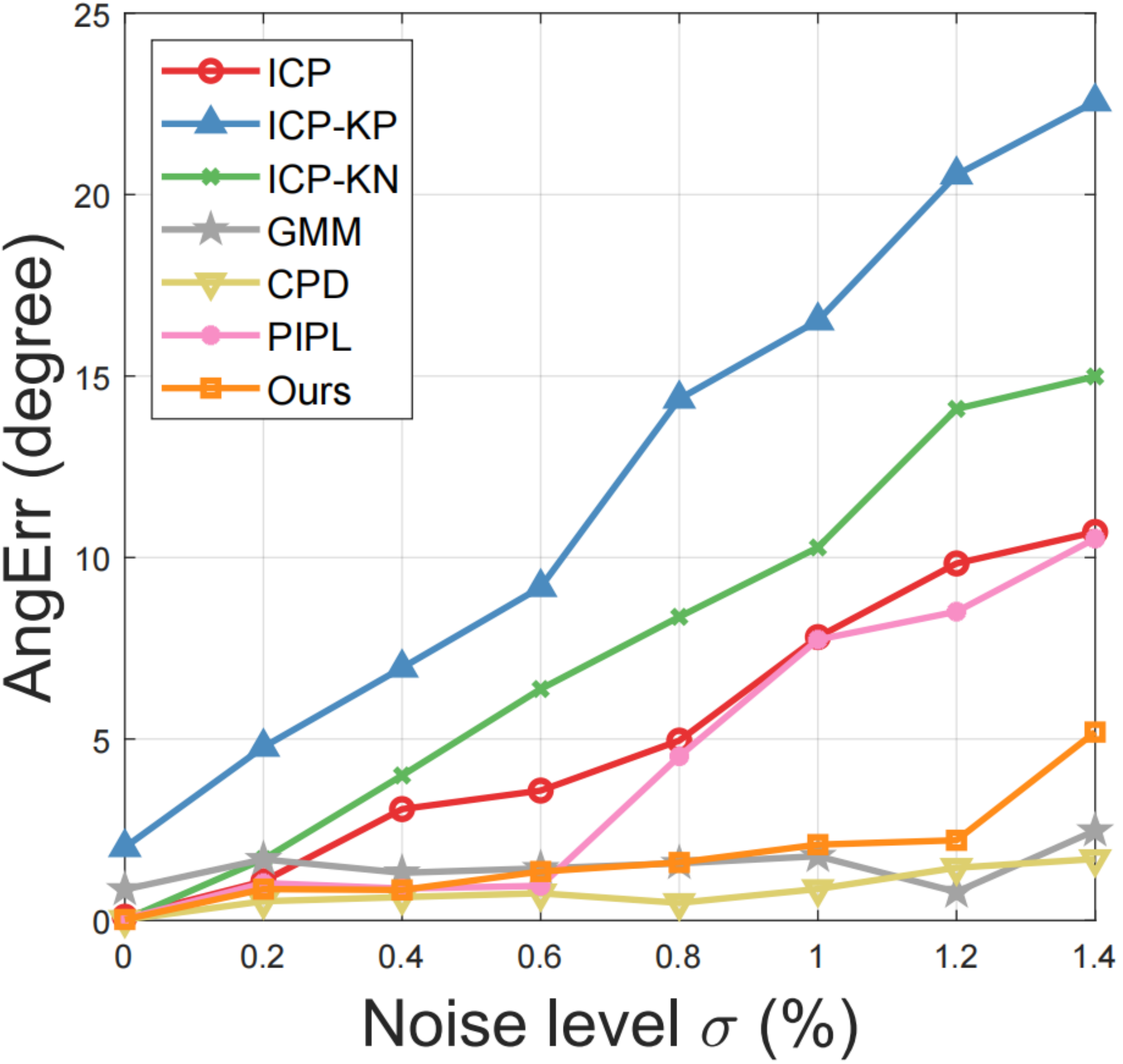}
	}
	\subcaptionbox{{\color{black}{Runtime}}}{
		\includegraphics[width=0.227\textwidth]{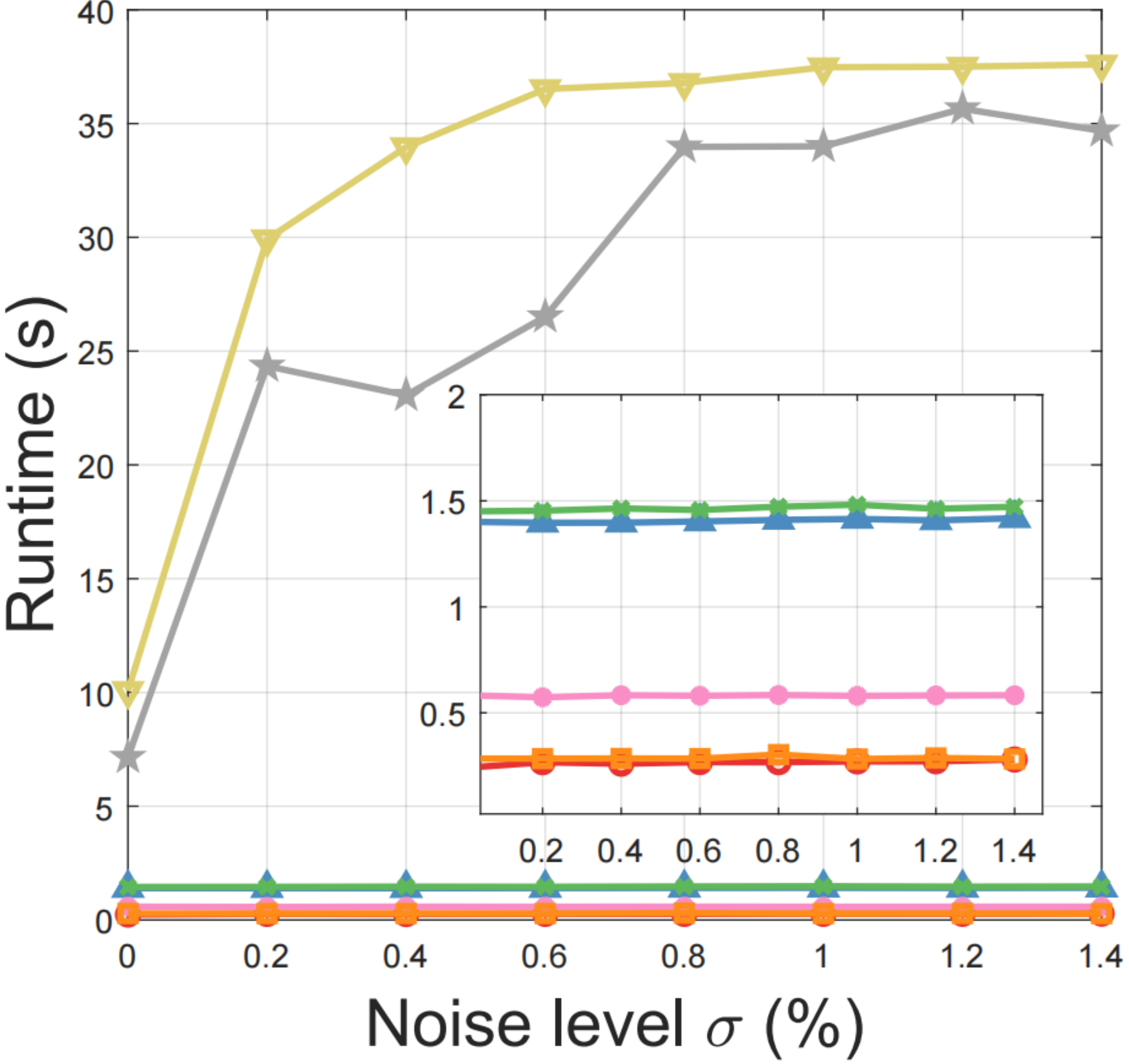}
	}	

	\vskip -0.2cm
	\caption{\revise{Comparisons of different methods with respect to the spatial noise}. Top: (a) shows two noisy point clouds with zero mean and standard deviation $\sigma=0.8\%$; (b) and (c) are the alignment results by the baseline approach PIPL and the proposed method, where values inside the parenthesis are the corresponding AngErr. \revise{Bottom: Statistic results of compared methods under $\sigma\in [0\%, 1.4\%]$, where our proposed method attains high precision and runs significantly fast.}}%
	\label{tab:bunny}
	\vskip -0.3cm
\end{figure}

\begin{figure*}
	\begin{center}
		\begin{minipage}{0.18\textwidth}
			\subcaptionbox*{Input}{
				\includegraphics[width=\textwidth]{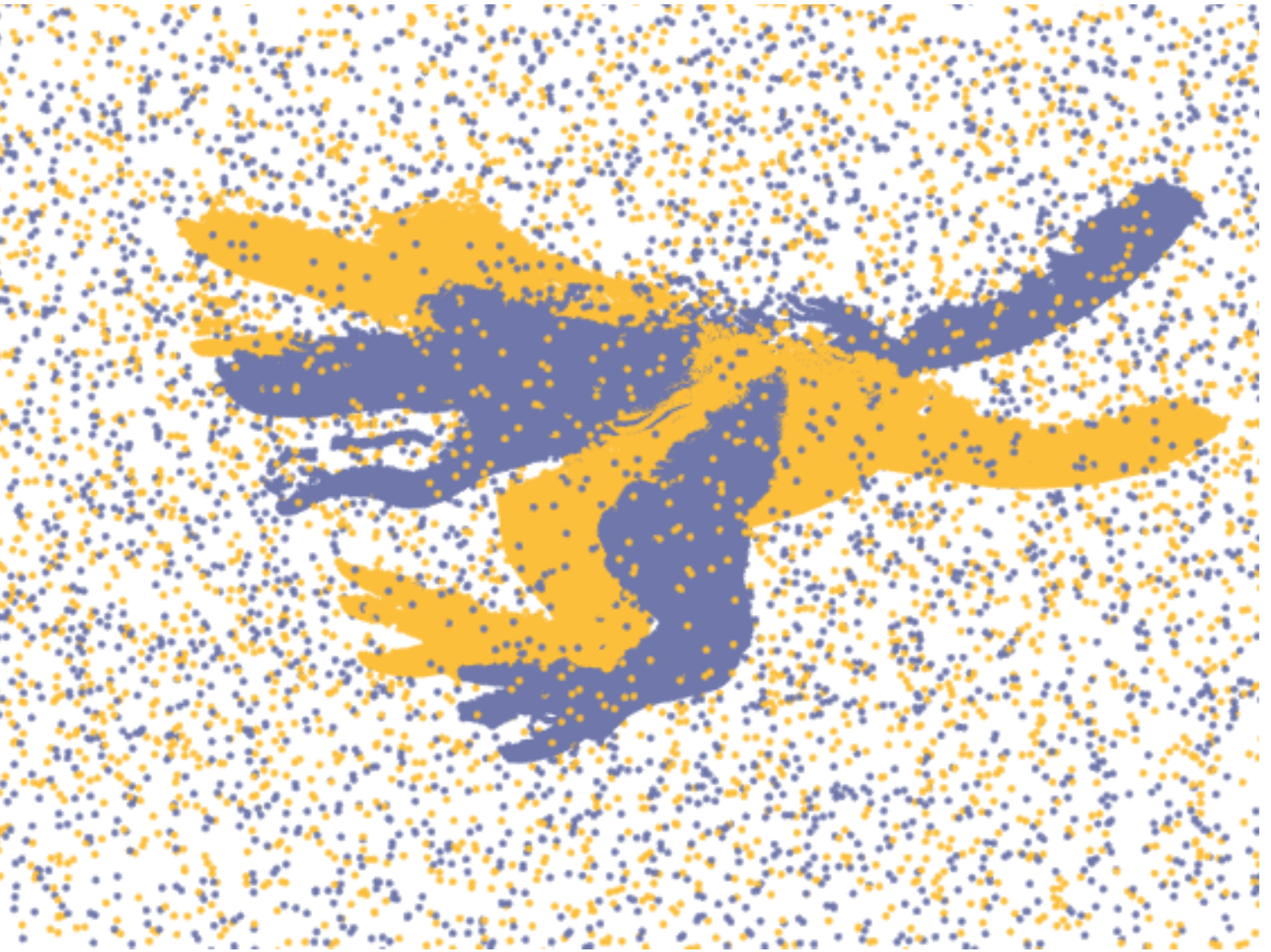}
			}\\
			\subcaptionbox*{ECM (0.8942)}{
				\includegraphics[width=\textwidth]{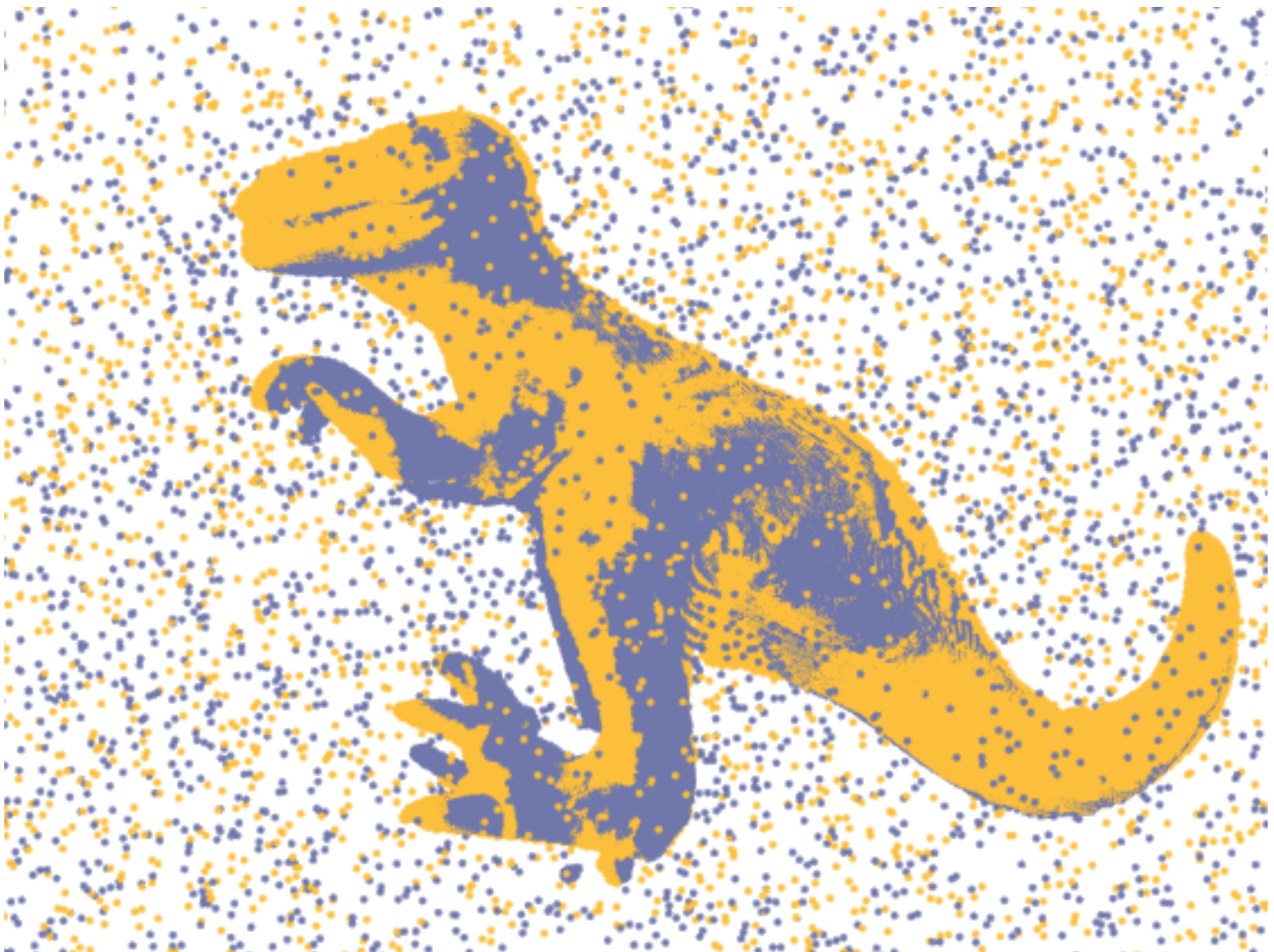}
			}\\
			\subcaptionbox*{Input}{
				\includegraphics[width=\textwidth]{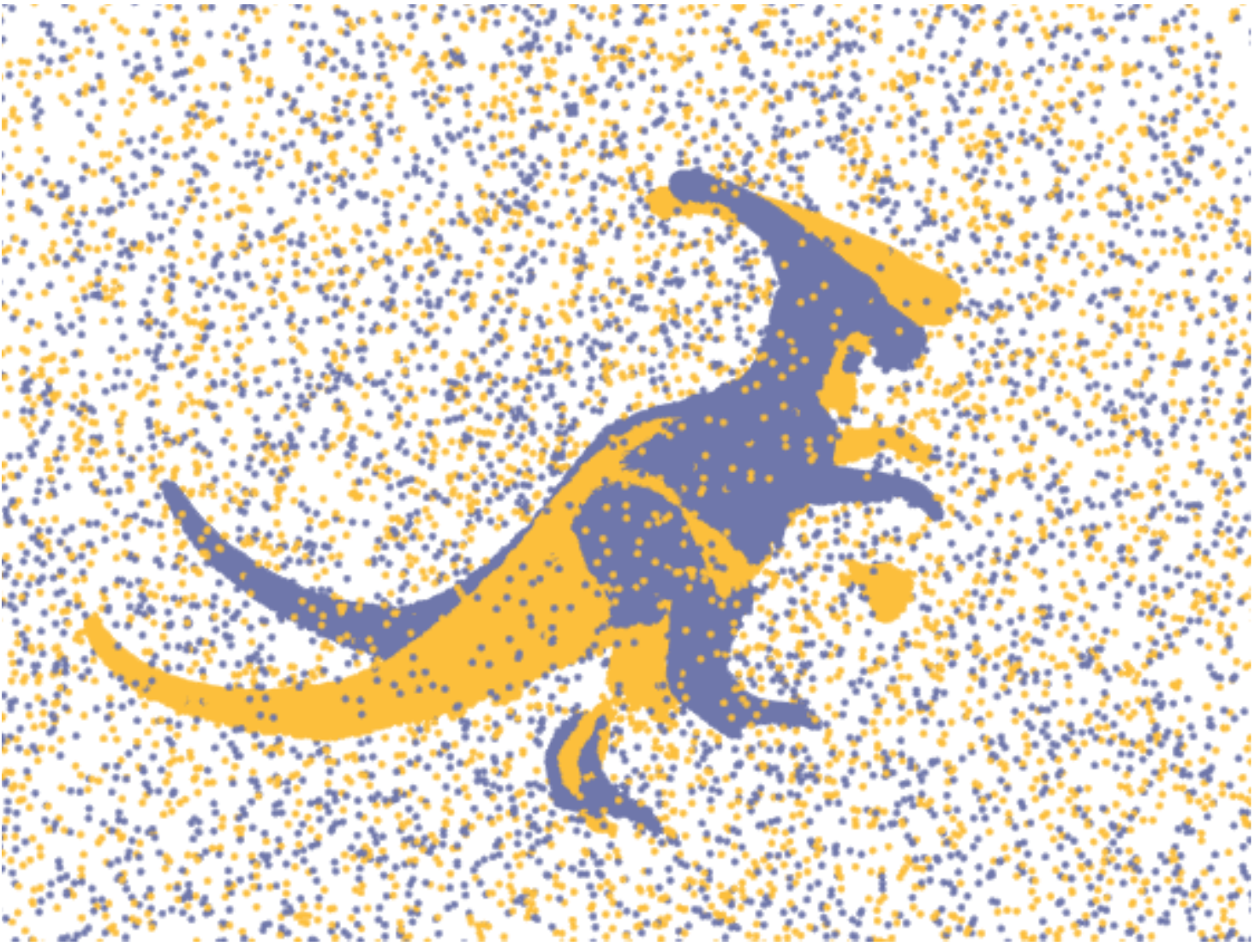}
			}\\
			\subcaptionbox*{ECM (0.5220)}{
				\includegraphics[width=\textwidth]{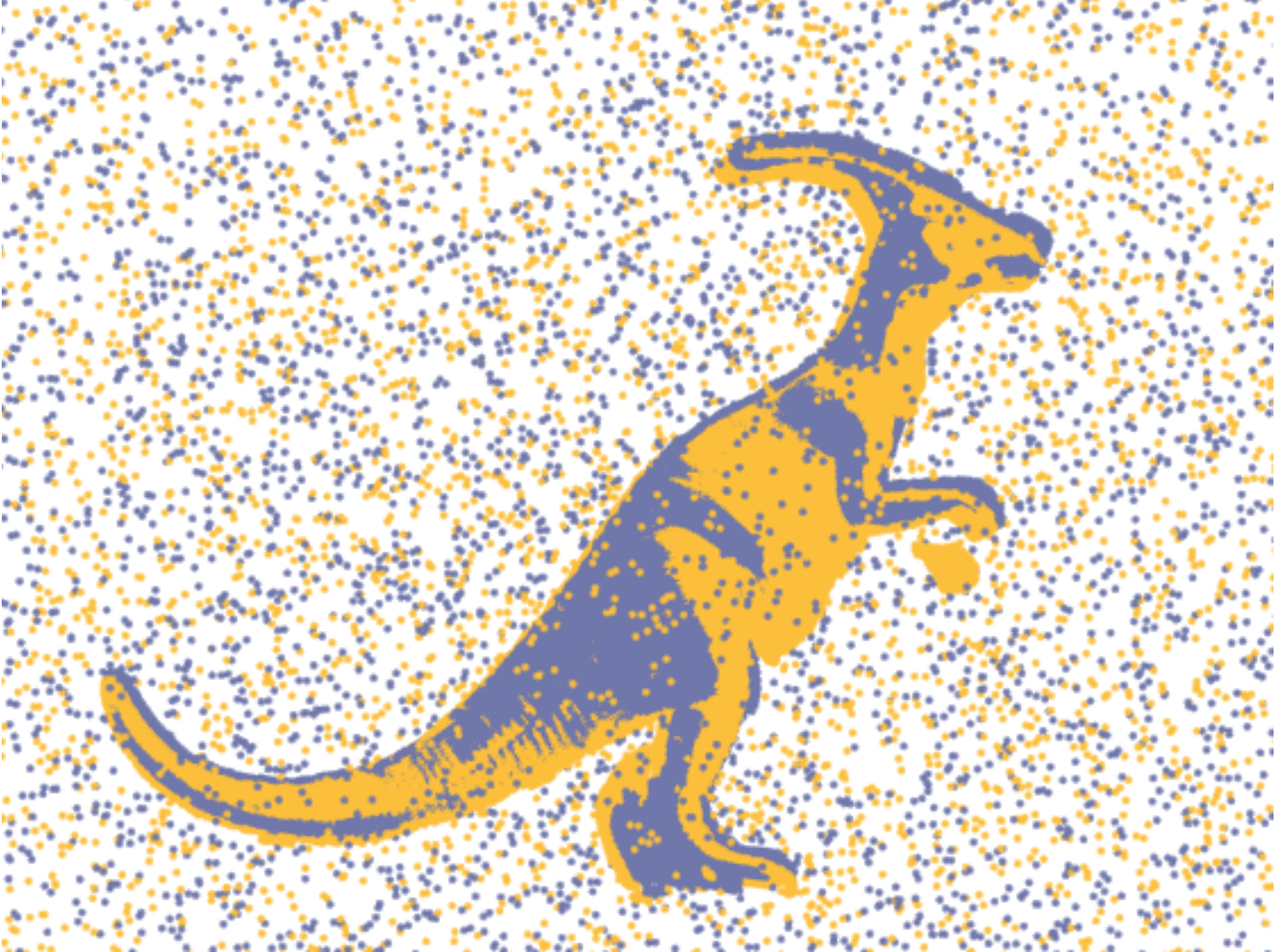}
			}
		\end{minipage}
		\begin{minipage}{0.18\textwidth}
			\subcaptionbox*{GT}{
				\includegraphics[width=\textwidth]{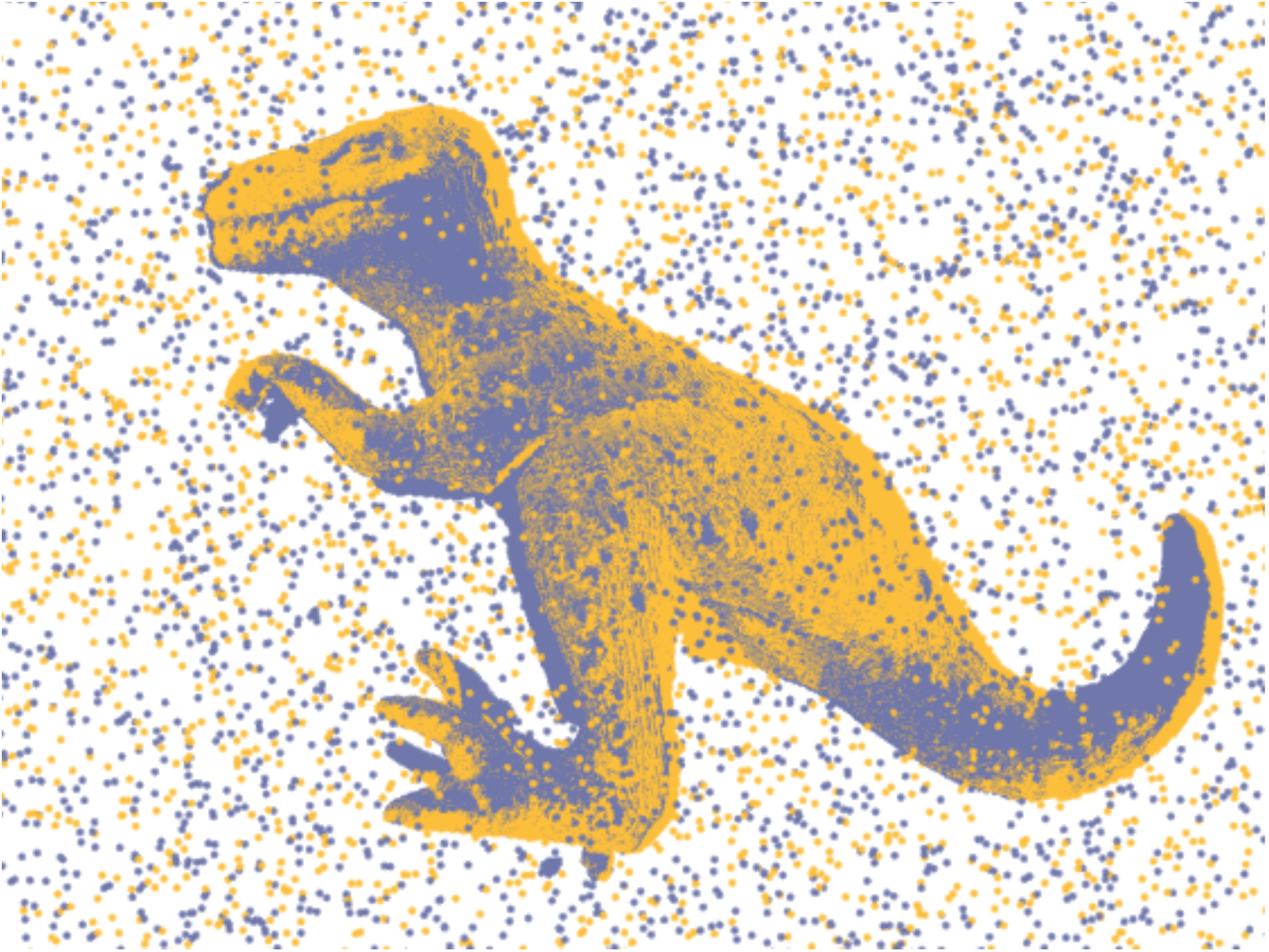}
			}\\
			\subcaptionbox*{FGR (1.3374)}{
				\includegraphics[width=\textwidth]{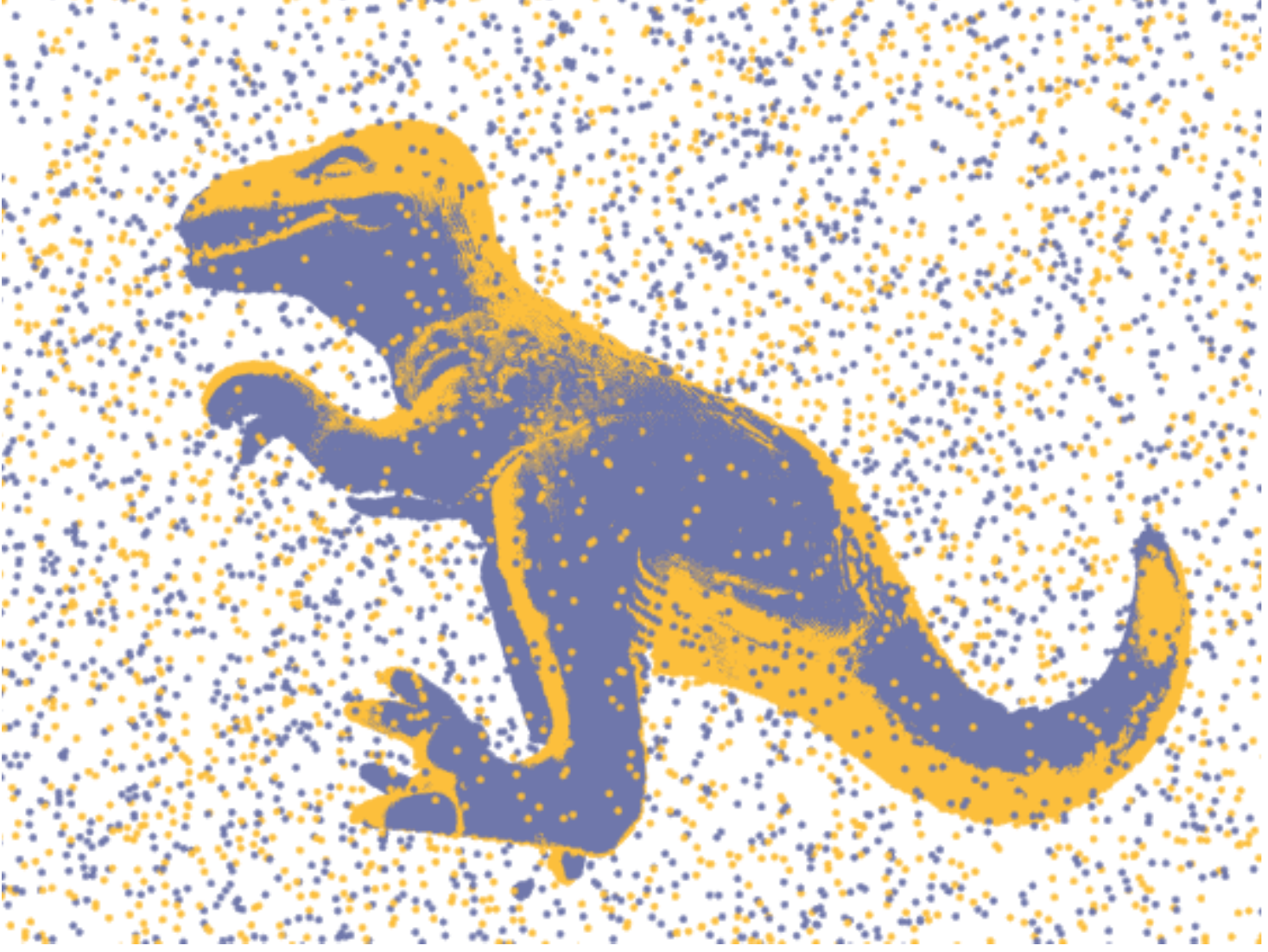}
			}\\
			\subcaptionbox*{GT}{
				\includegraphics[width=\textwidth]{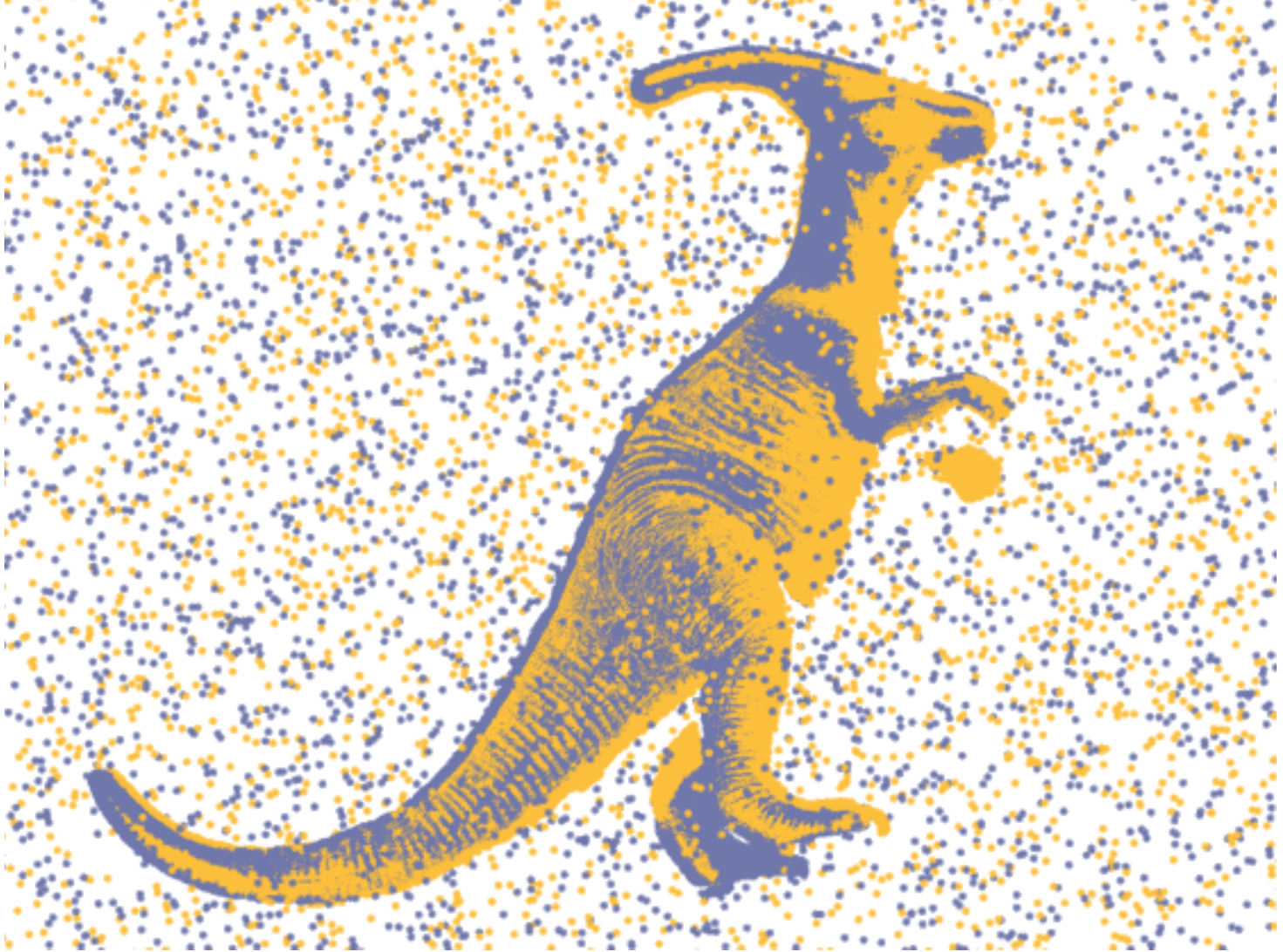}
			}\\
			\subcaptionbox*{FGR (15.5914)}{
				\includegraphics[width=\textwidth]{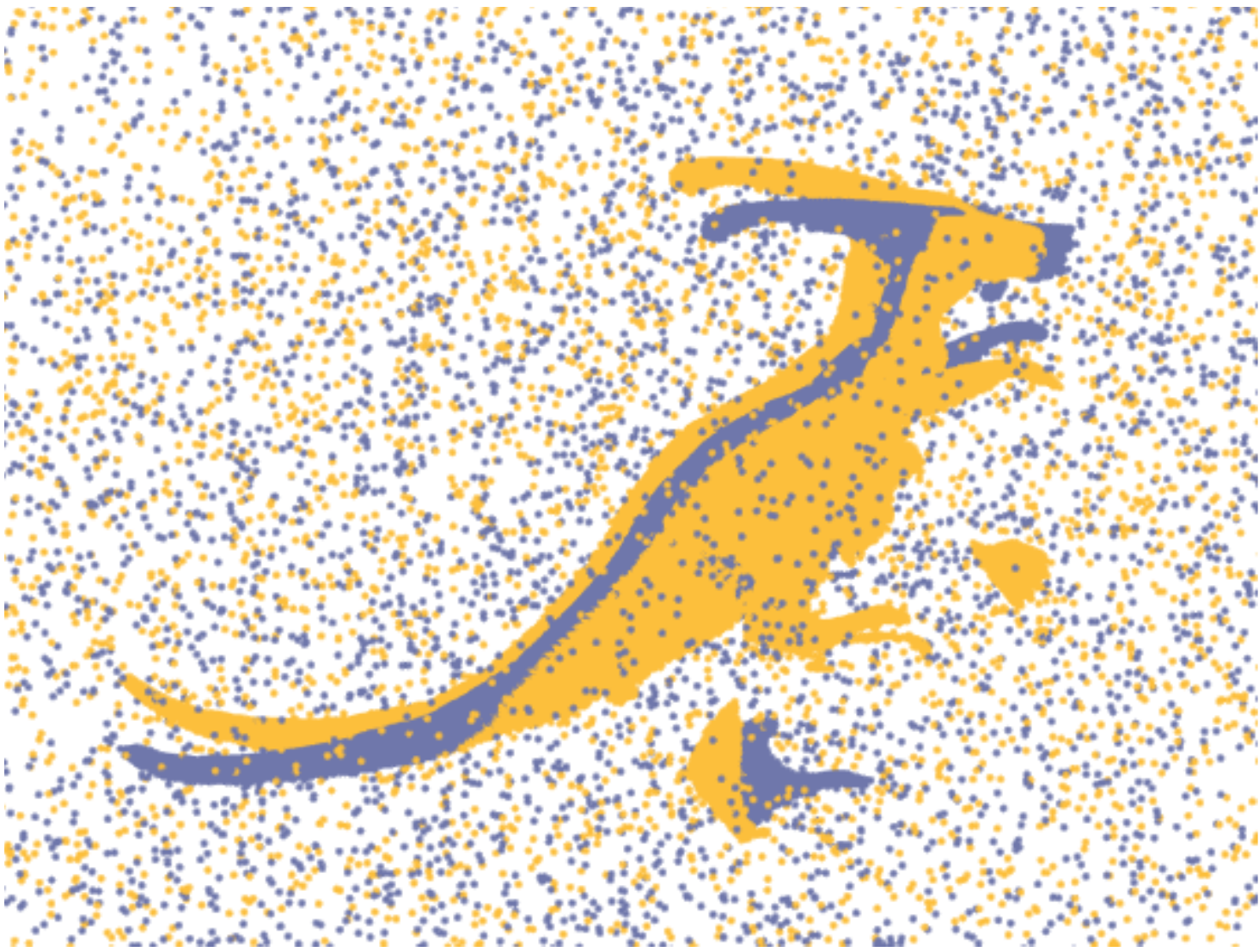}
			}
		\end{minipage}
		\begin{minipage}{0.18\textwidth}
			\subcaptionbox*{TrICP (0.0115)}{
				\includegraphics[width=\textwidth]{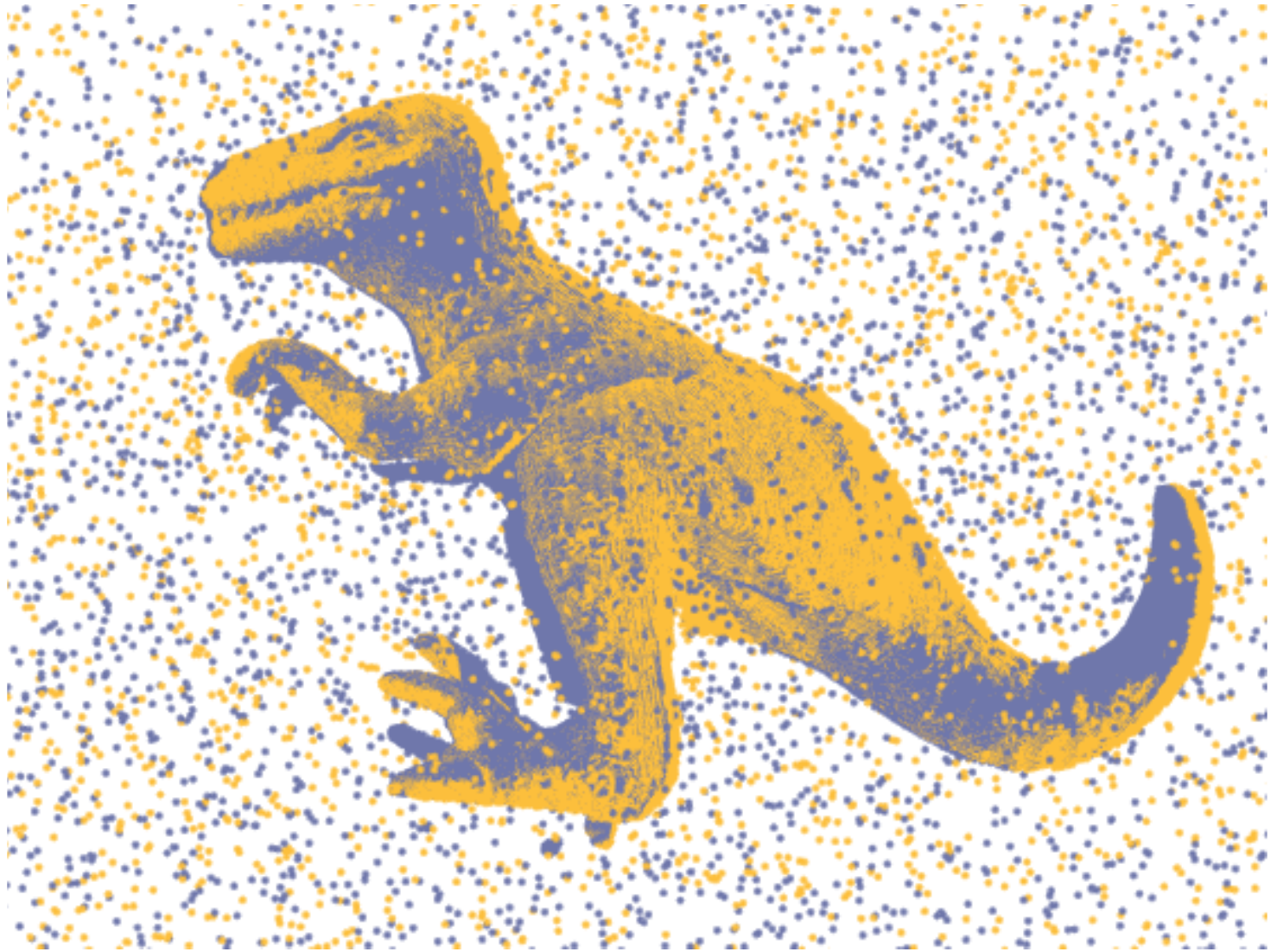}
			}\\
			\subcaptionbox*{TEASER++ (0.8172)}{
				\includegraphics[width=\textwidth]{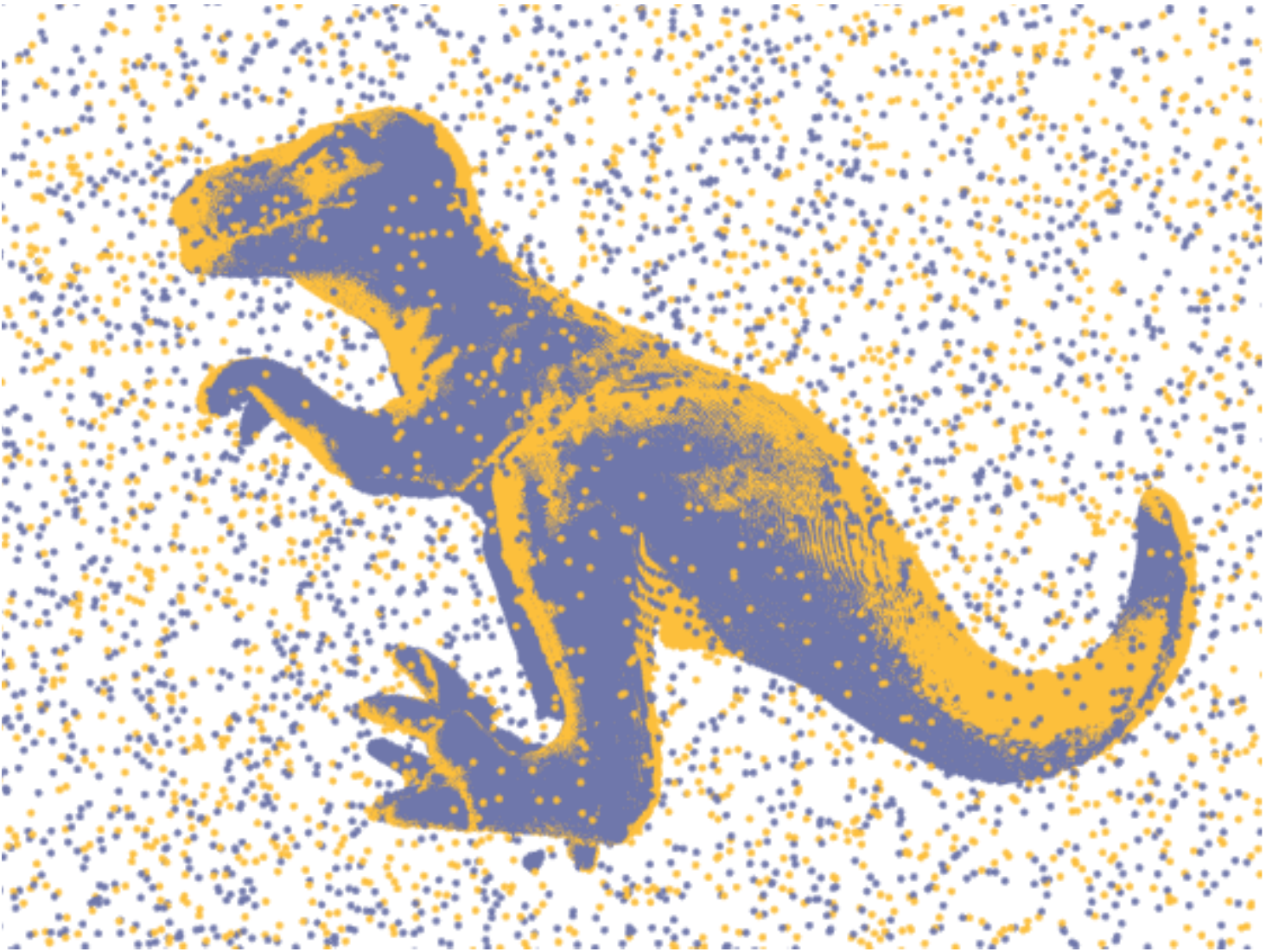}
			}\\
			\subcaptionbox*{TrICP (2.4634)}{
				\includegraphics[width=\textwidth]{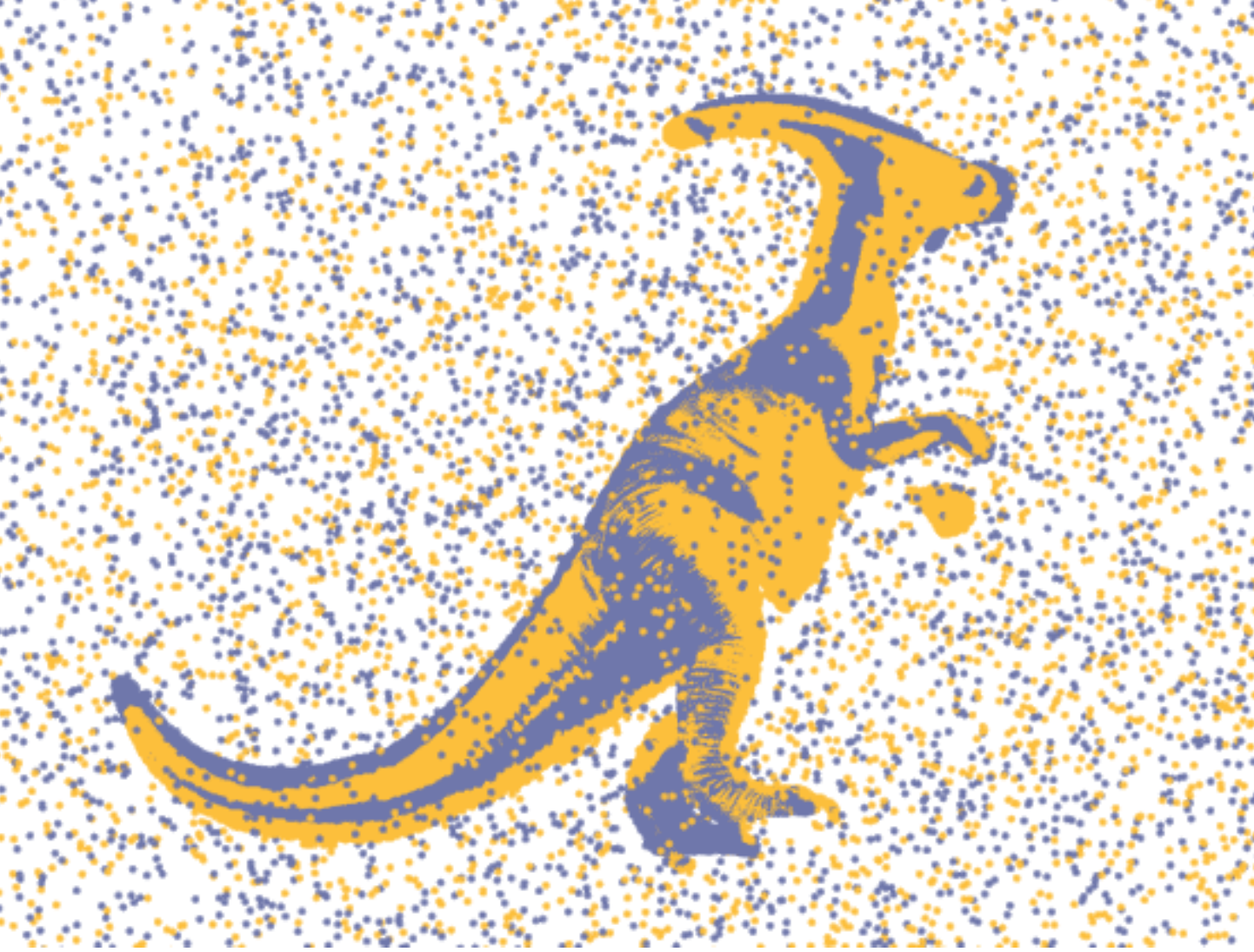}
			}\\
			\subcaptionbox*{TEASER++ (1.5666)}{
				\includegraphics[width=\textwidth]{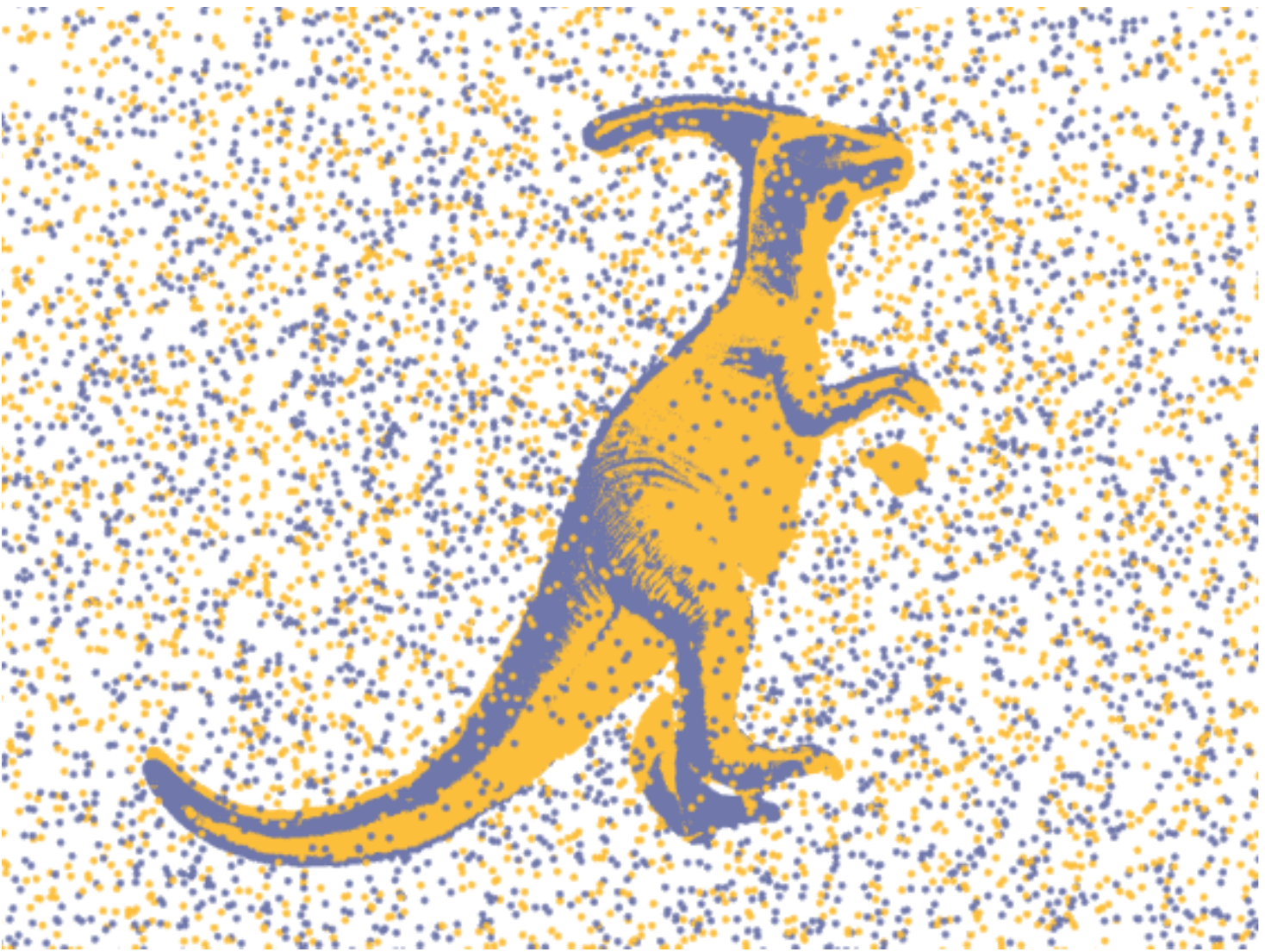}
			}
		\end{minipage}
		\begin{minipage}{0.18\textwidth}
			\subcaptionbox*{GMM (21.8886)}{
				\includegraphics[width=\textwidth]{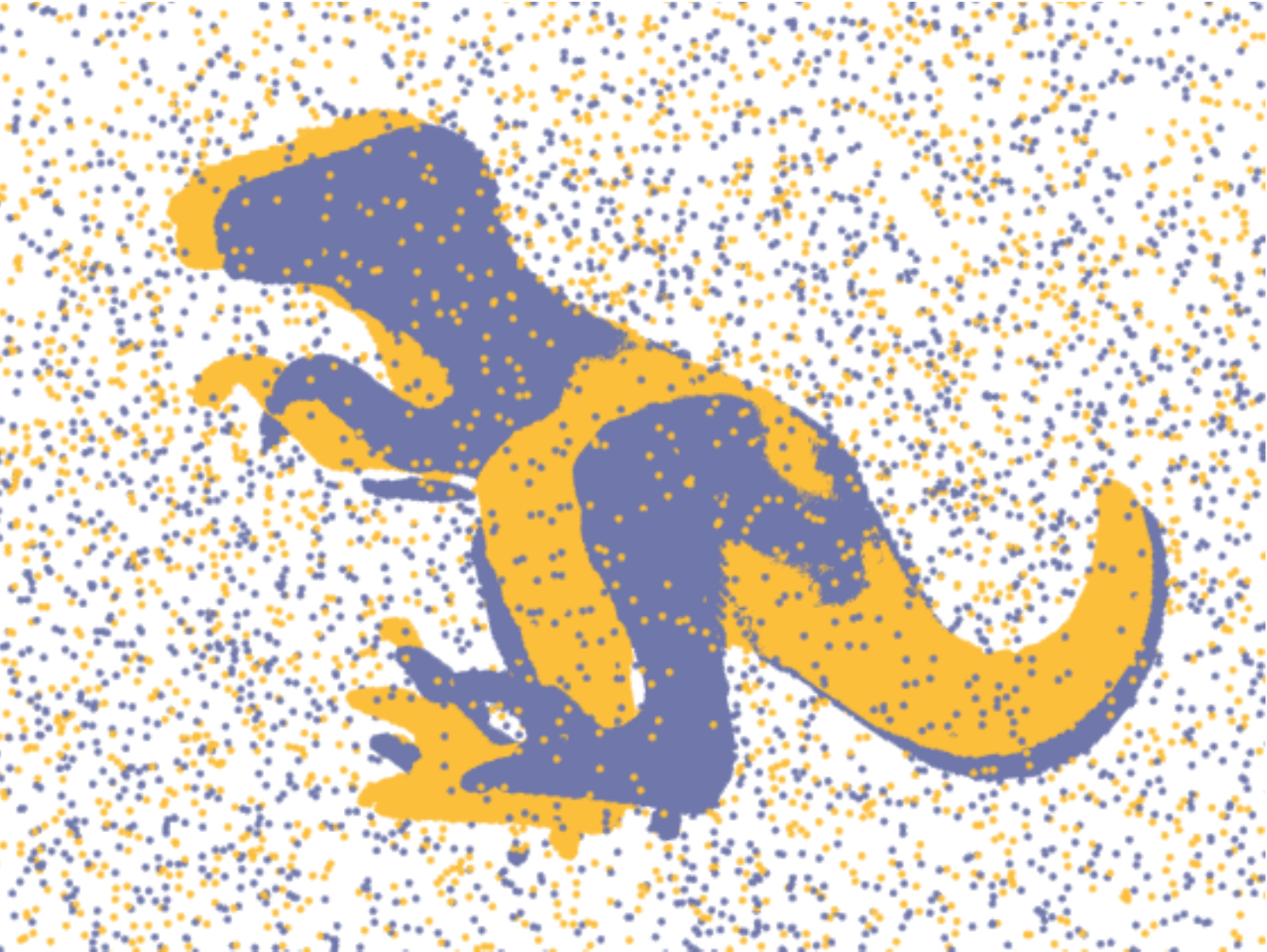}
			}\\
			\subcaptionbox{FRICP (0.4695)}{
				\includegraphics[width=\textwidth]{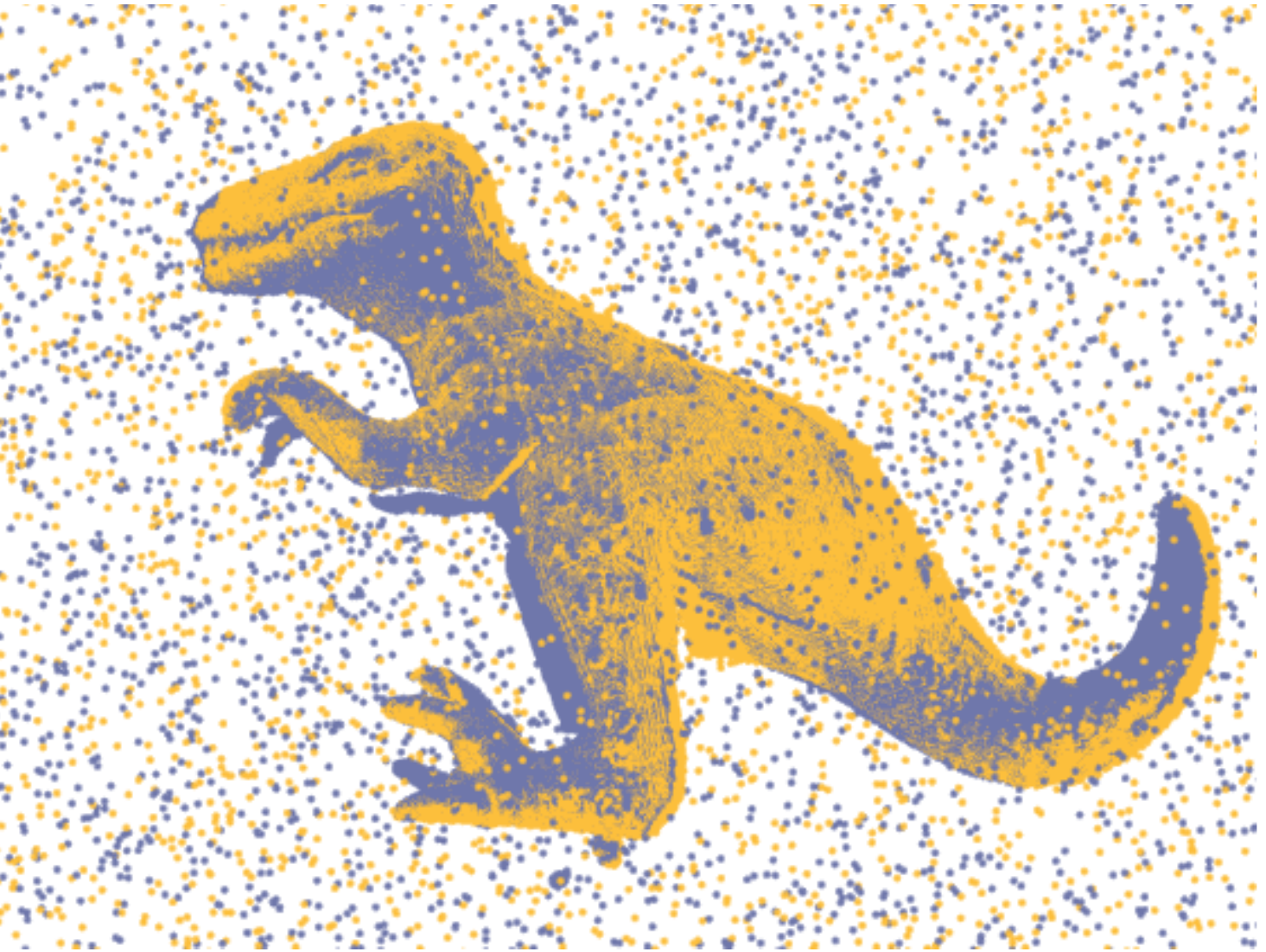}
			}\\
			\subcaptionbox*{GMM (2.7801)}{
				\includegraphics[width=\textwidth]{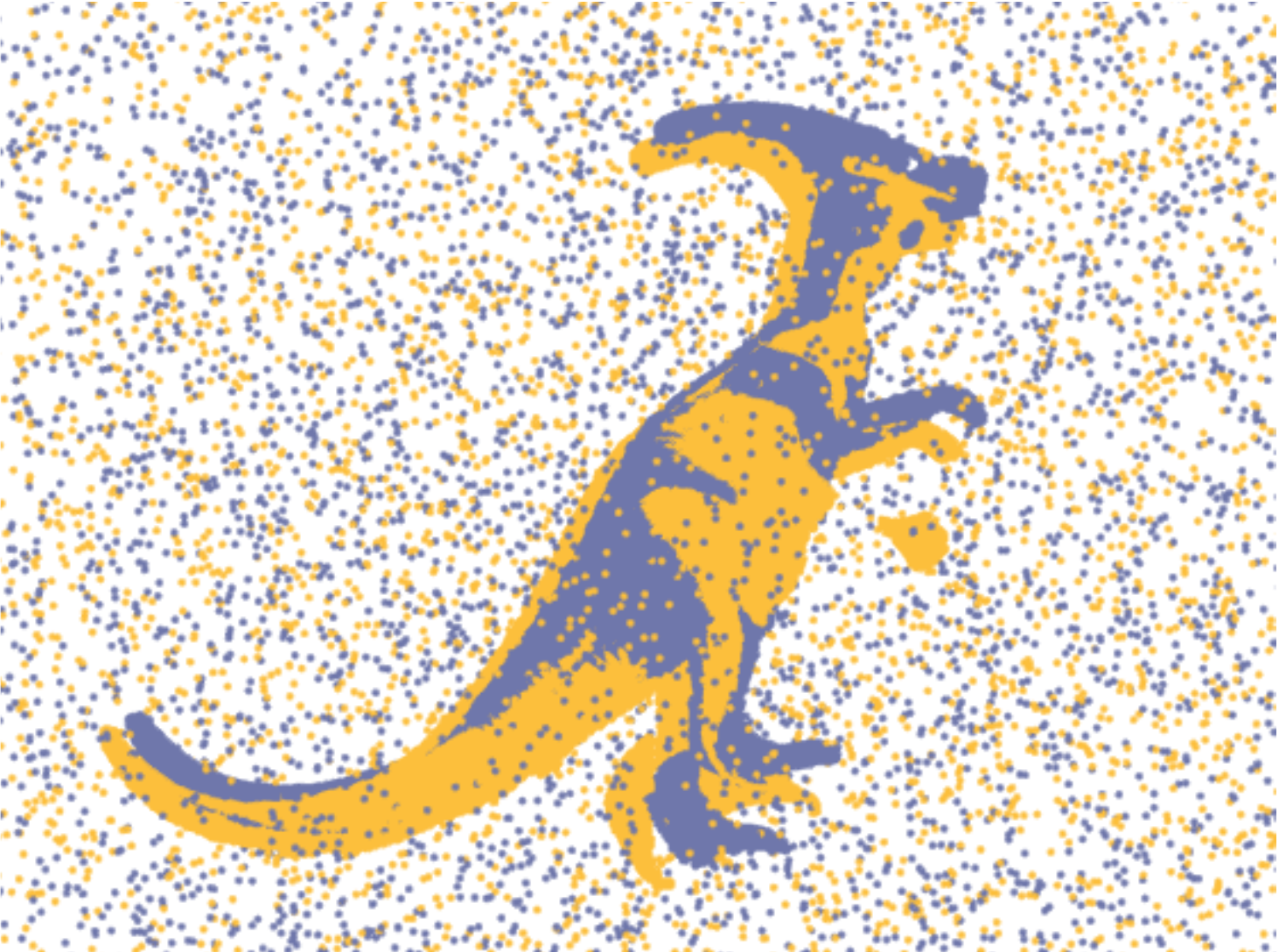}
			}\\
			\subcaptionbox*{FRICP (0.6008)}{
				\includegraphics[width=\textwidth]{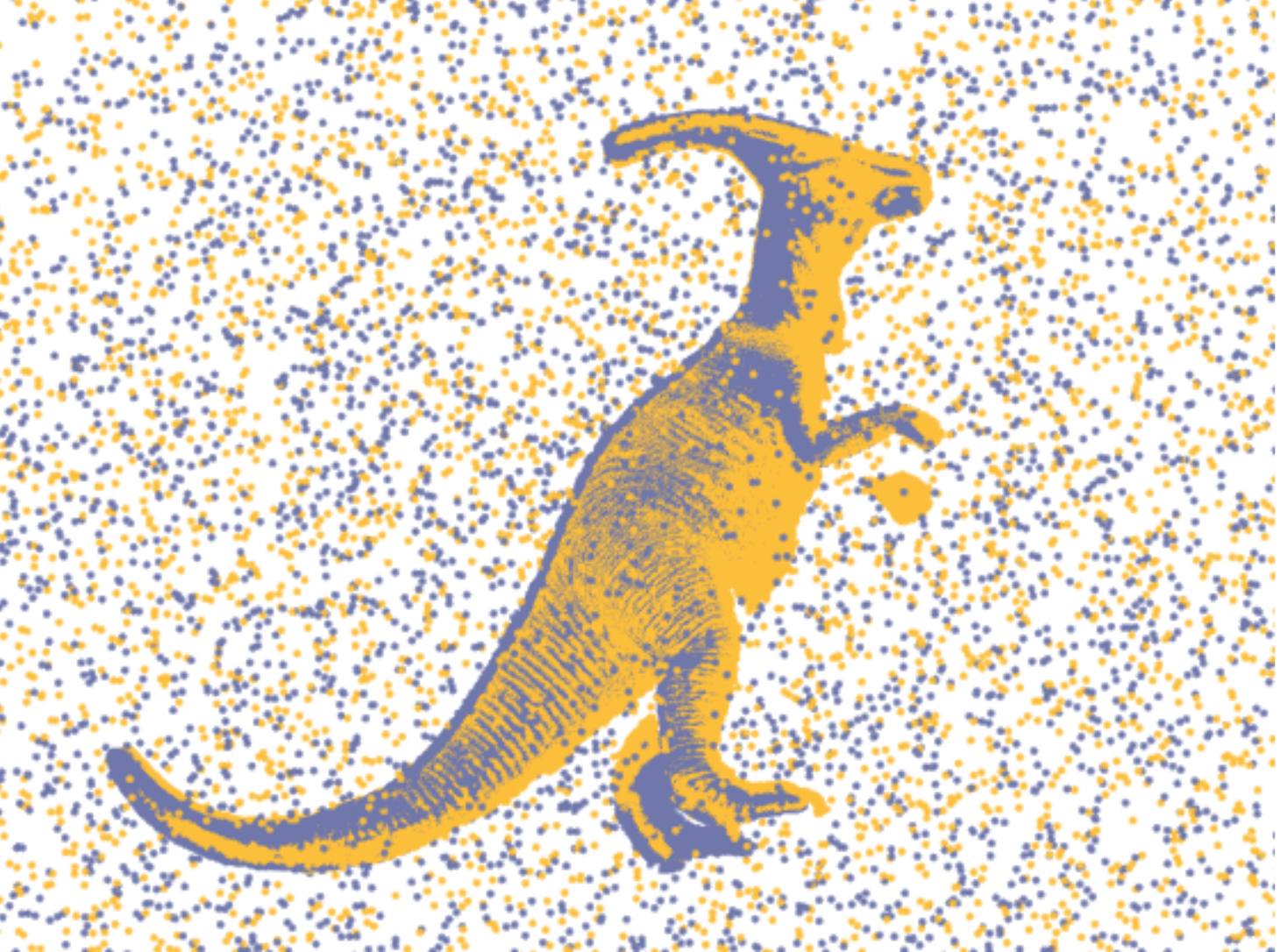}
			}
		\end{minipage}
		\begin{minipage}{0.18\textwidth}
			\subcaptionbox*{CPD (0.2030)}{
				\includegraphics[width=\textwidth]{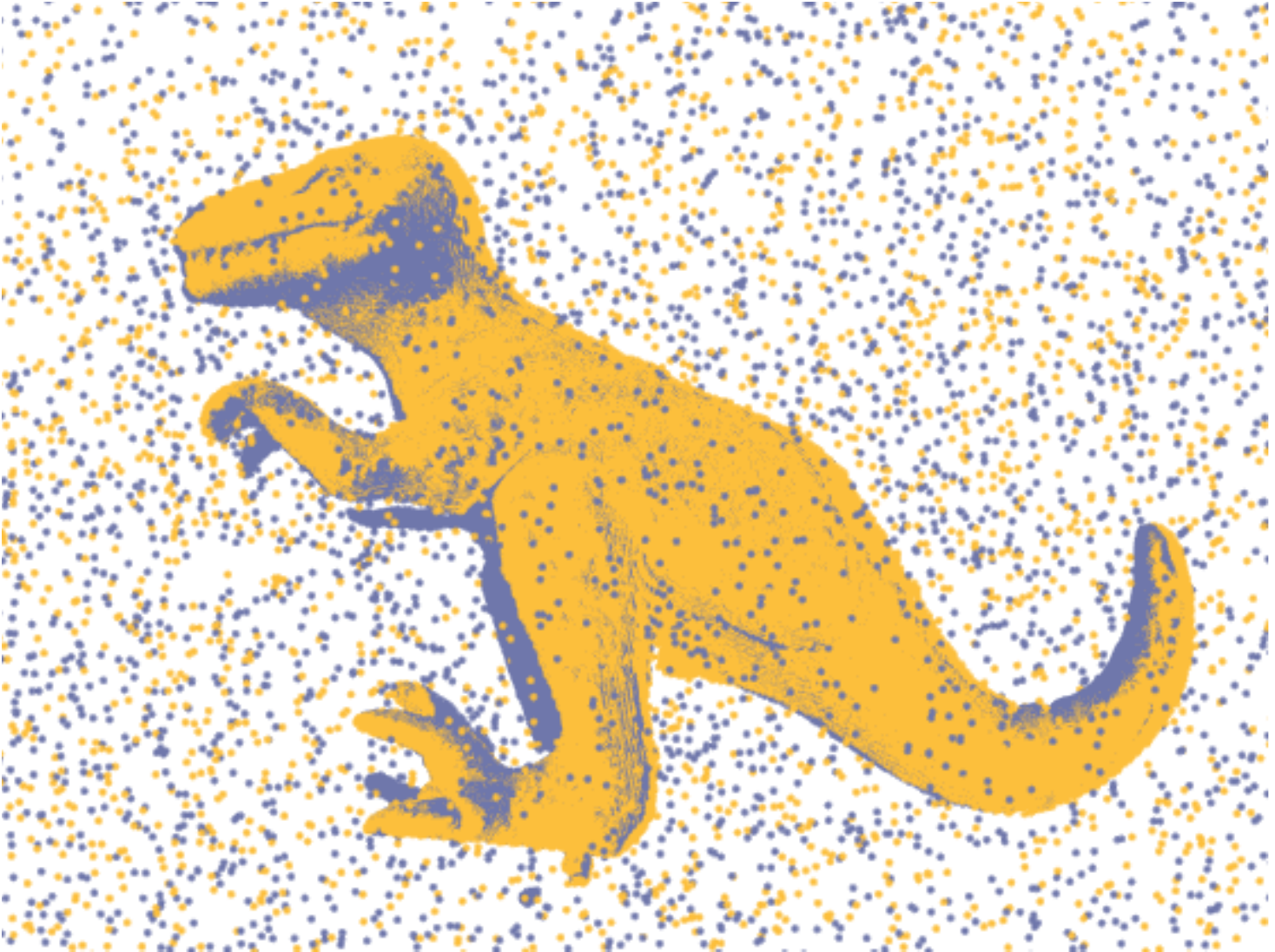}
			}\\
			\subcaptionbox*{Ours (0.3540)}{
				\includegraphics[width=\textwidth]{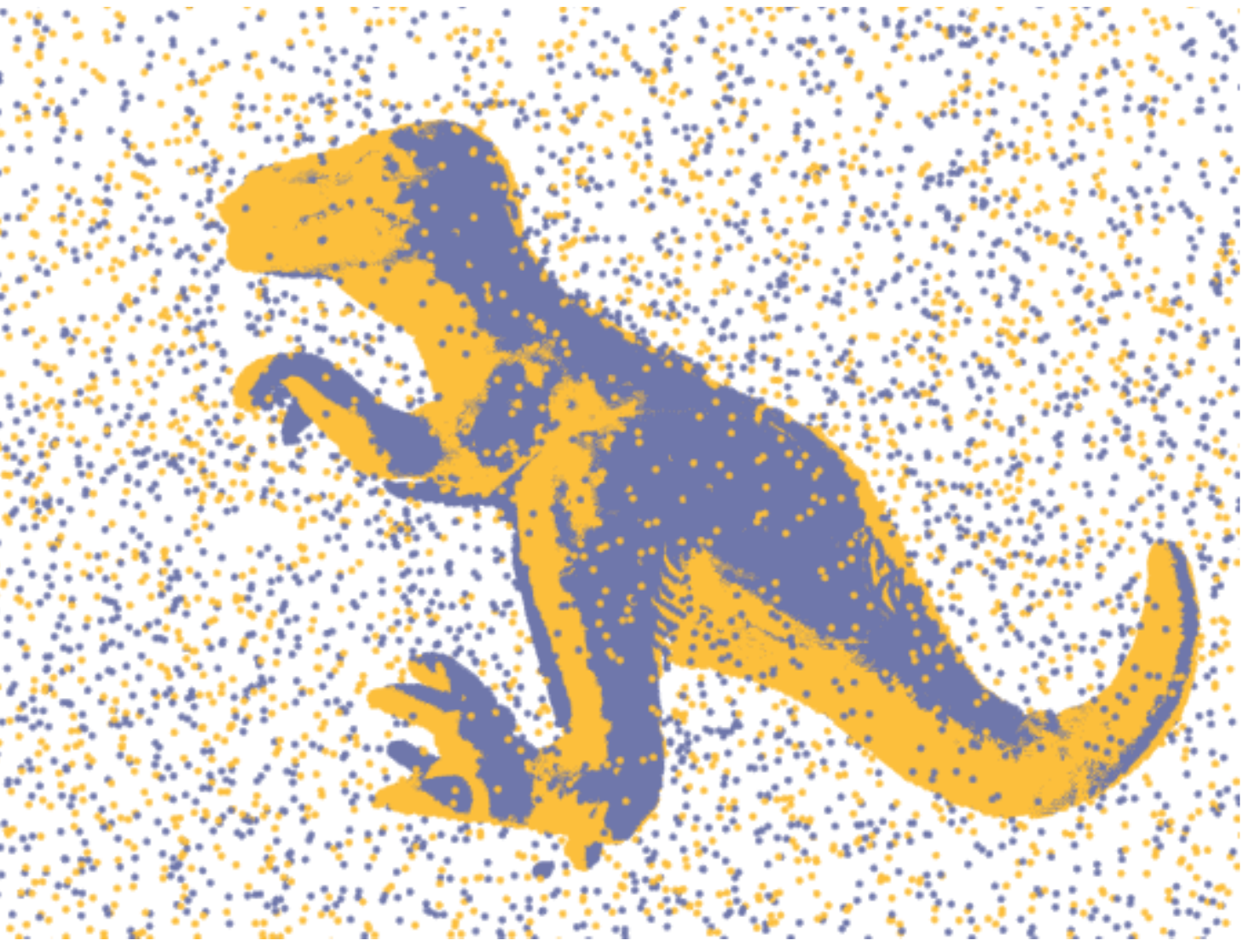}
			}\\
			\subcaptionbox*{CPD (0.3683)}{
				\includegraphics[width=\textwidth]{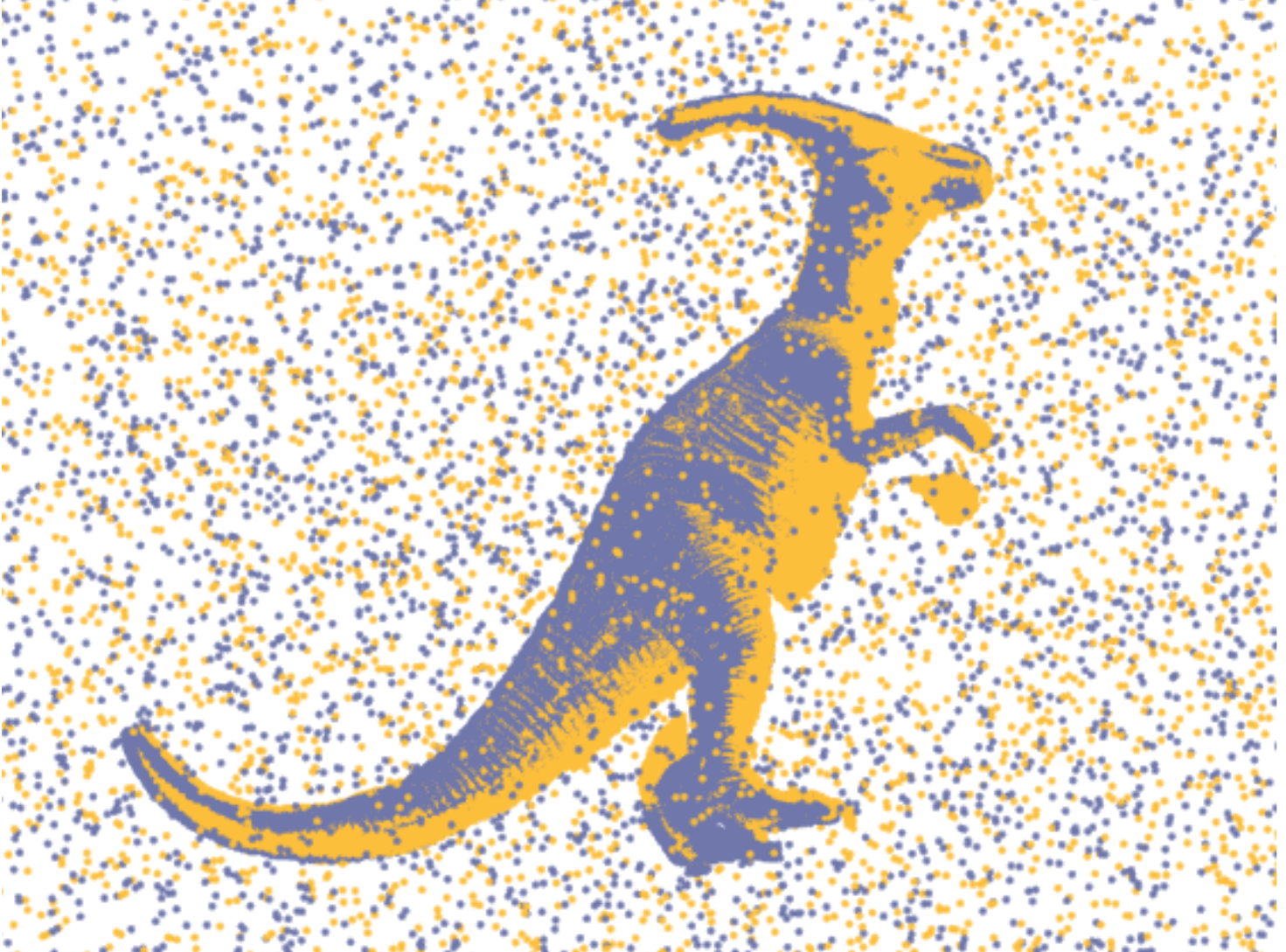}
			}\\
			\subcaptionbox*{Ours (0.3001)}{
				\includegraphics[width=\textwidth]{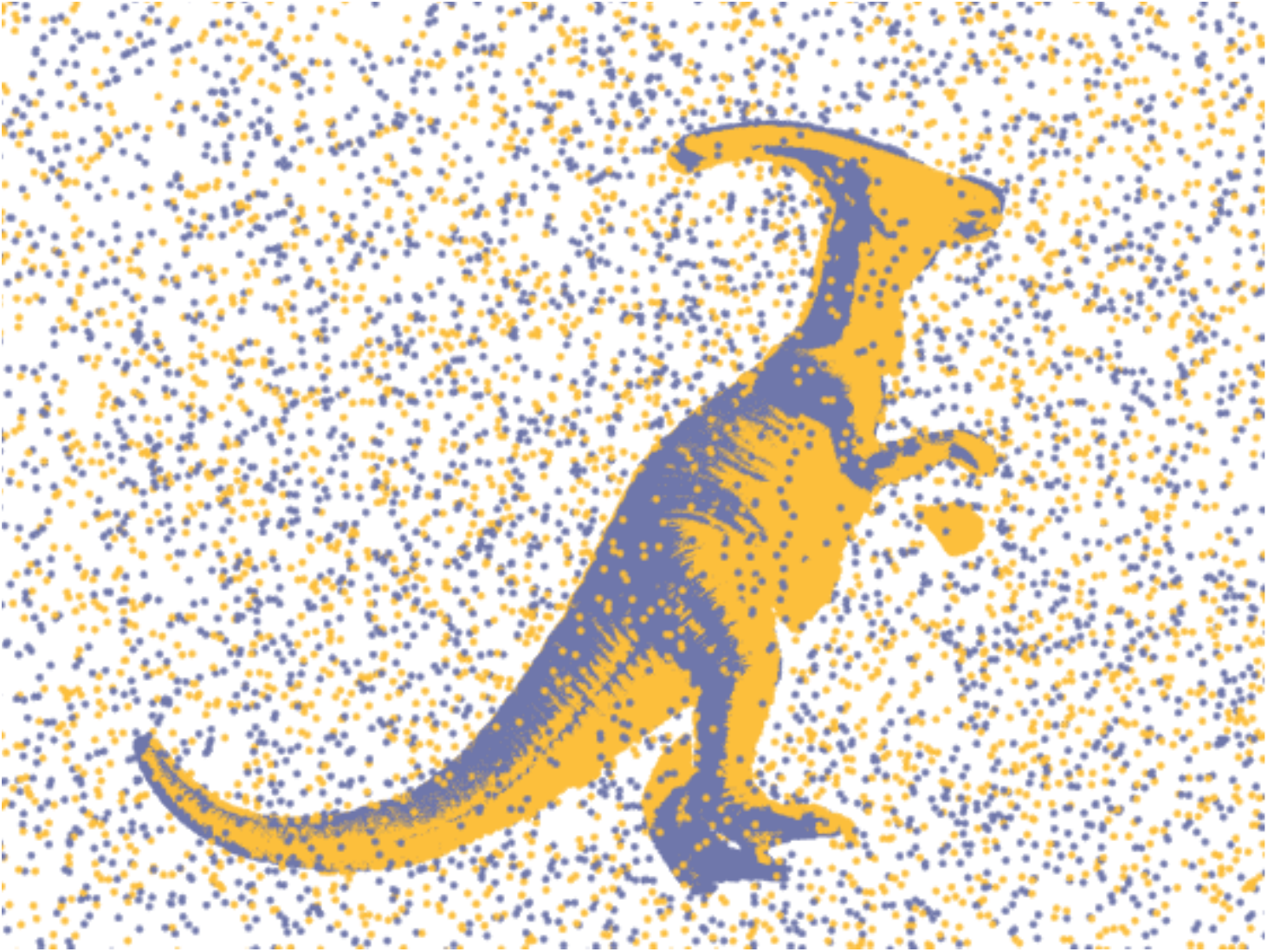}
			}
		\end{minipage}
		\vskip -0.2cm	
		\caption{Registration results of outlier-contaminated point clouds. Our method attains the highest or comparable robustness.} 
		\label{outlier_test}
	\end{center}
	\vskip -0.3cm		
\end{figure*}

\subsubsection{Robustness test}
We contaminate the Bunny dataset from~\cite{DataStanford} \revise{with} spatial noise and conduct 10 registrations to test the robustness of the proposed method. Bunny is first randomly perturbed about $30^\circ$ along the $X,$ $Y,$ $Z$ axes. \revise{Then,} Gaussian noise with zero mean and different standard deviations $\sigma\in[0\%, 1.4\%]$ relative to the three directions is added to the original 40,256 points. Fig.~\ref{tab:bunny}(a)-(c) show two sample point clouds with $\sigma=0.8\%$, and the corresponding registration results by PIPL and our method; our proposed algorithm shows higher robustness with lower AngErr, as indicated in the parenthesis. Fig.~\ref{tab:bunny}(d) and (e) summarize the registration error and running time of compared approaches (we omit ECM due to its instability and high amount of errors), in which GMM, CPD, and the proposed method attain the overall highest accuracy and outperform the others \revise{in terms of} robustness. \revise{This result is} attributed to the soft correspondence manner with probability. However, as observed in Fig.~\ref{tab:bunny}(e), GMM and CPD are much slower than \revise{our method}. ICP-based methods \revise{remain} efficient. However,  ICP and its variants are more sensitive to noise and \revise{lead to} significant errors due to the hard correspondence scheme between points. 

\begin{table}[t]
	\vskip -0.3cm
	\centering
	\caption{Robustness test against 30\% outliers on the T-rex and Parasaurolophus datasets, where the first two best results are indicated by \textbf{green} and \textbf{red}, respectively.}
	\vskip -0.2cm
	\renewcommand\arraystretch{1.2}
	\setlength{\tabcolsep}{2mm}{
		\begin{tabular}{|c|cc|cc|} 
			\hline
			\multirow{2}{*}{\diagbox{\textbf{Method}}{\textbf{Dataset}}} & \multicolumn{2}{c|}{T-rex} & \multicolumn{2}{c|}{Parasaurolophus}\\ \cline{2-5}
			&AngErr ($^\circ$)&Time (s)&AngErr ($^\circ$)&Time (s) \\ \cline{1-5}
			TrICP~\cite{chetverikov2002trimmed}&1.6420&0.9655&1.4566&1.3820\\ \cline{1-5}
			GMM~\cite{jian2010robust}&10.6462&48.5120&3.0283&39.3940\\ \cline{1-5}
			CPD~\cite{myronenko2010point}&\cellcolor{best}0.2492&41.7030&\cellcolor{second}{0.3905}&52.4800\\ \cline{1-5}
			ECM~\cite{horaud2010rigid}&{0.4965}&64.3260&0.9547&82.1900\\ \cline{1-5}
			TEASER++~\cite{yang2020teaser}&0.6169&0.1362&0.7035&0.1189\\ \cline{1-5}
			FGR~\cite{zhou2016fast}&1.3326&0.2255&14.2750&0.1973\\ \cline{1-5}
			FRICP~\cite{zhang2021fast}&\cellcolor{second}{0.4695}&1.2454&0.6008&1.2221\\ \cline{1-5}
			Ours&{0.5392}&0.6180&\cellcolor{best}{0.2455}&0.5770\\ \cline{1-5}
		\end{tabular}
	}
	\label{table:UWA}
	\vskip -0.3cm
\end{table}
Subsequently, we test the robustness of the proposed method against outliers. We use TrICP~\cite{chetverikov2002trimmed}, GMM, CPD, ECM, and three representative robust registration approaches, including FGR~\cite{zhou2016fast}, TEASER++~\cite{yang2020teaser}, and FRICP~\cite{zhang2021fast} (all implemented in C++), for comparison, where 
FGR is optimized by the Open3D library~\cite{Open3D}. We leave out PIPL due to its sensitivity to outliers. We use two partially overlapping objects from the \emph{UWA} dataset~\cite{mian2006three} for assessment: T-rex and Parasaurolophus, which are acquired by a Hokuyo UTM-30LX scanning rangefinder. 
We randomly add 30\% outliers \revise{sampled from zero mean Gaussian distribution with the standard variance relative to the bounding box size of each model}, as illustrated in Fig.~\ref{outlier_test}. Table~\ref{table:UWA} reports the average results after 10 tests, where we conclude that CPD, ECM, TEASER++, FRICP, and our method are quite robust, with deviations below $1^\circ$. The robustness of CPD and ECM is attributed to an additional uniform distribution except to the Gaussian mixture models, which require careful tuning of the trade-off weight. TEASER++ uses the truncated least-squares and FRICP adopts the Welsch's function to depress spurious correspondence, \revise{thus guaranteeing} high robustness. They suppress correspondence deviations \revise{during} the registration process. By contrast, our method directly removes outliers according to \revise{point response intensity} and the robust MAD criterion, which is free of hyperparameter tuning and can be used as preprocessing for most registration algorithms. Moreover, the proposed method with runtime below 1s is substantially faster than most competitors, especially for probabilistic-based approaches. We present several  registration results in Fig.~\ref{outlier_test}.

\begin{figure}[t]
	\centering
	\begin{minipage}{0.23\textwidth}
		\includegraphics[width=\textwidth]{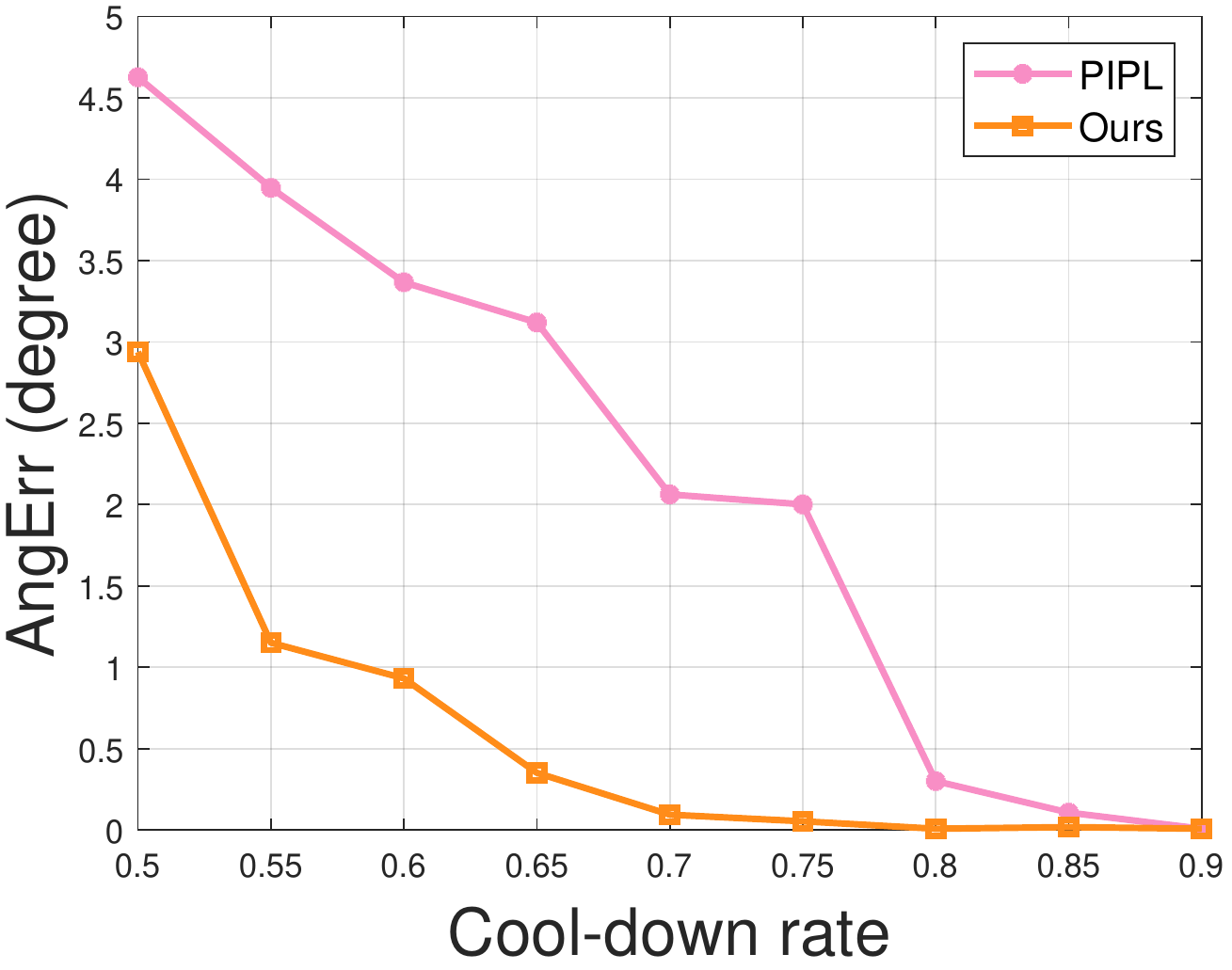}  
	\end{minipage}
	\begin{minipage}{0.23\textwidth}
		\includegraphics[width=0.5\textwidth]{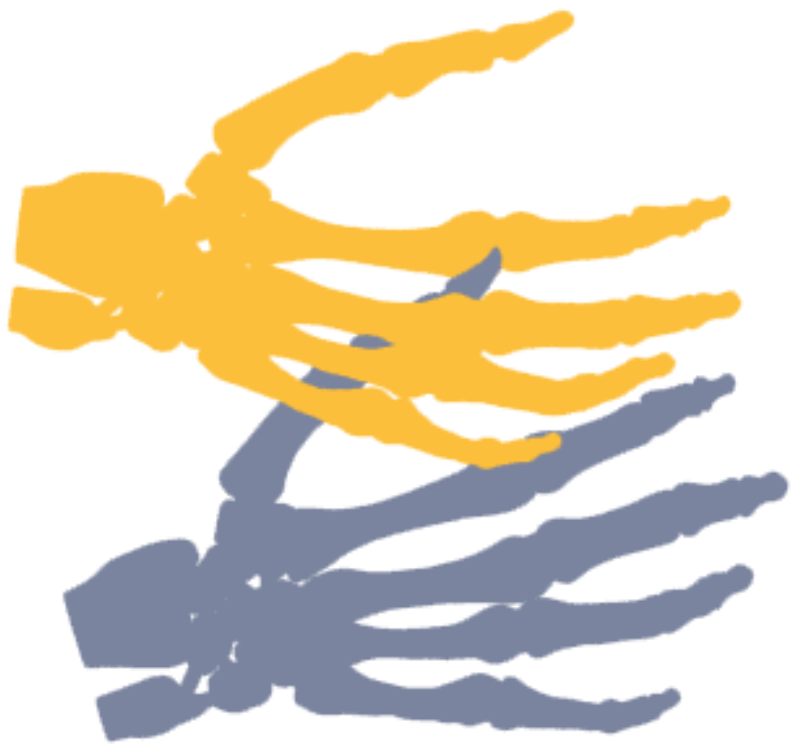}
		\includegraphics[width=0.48\textwidth]{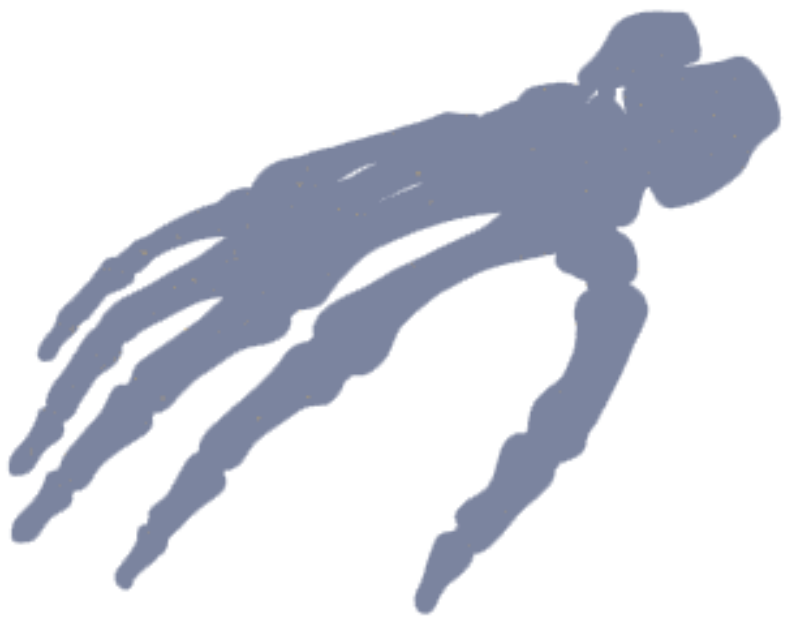}
	\end{minipage}
	\vskip -0.2cm
	\caption{Left: Investigation of the cool-down rate $\beta$ for point cloud registration. Right: The misalignment point clouds and the registration result by the proposed method with $\beta=0.9$.}
	\label{fig:hand}
	\vskip -0.3cm
\end{figure}

\subsection{Effect of the Cool-down Rate} We investigate the influence of the cool-down rate $\beta$ of the (adaptive) simulated annealing for point cloud registration. We adopt the medical dataset Skeleton Hand from the \textit{Large Geometric Model Archive} provided by Georgia Institute of Technology~\cite{DataGeo} for this purpose, as illustrated in the right panel of Fig.~\ref{fig:hand}. The target point cloud (yellow) has 327,323 points, and the source data (gray) is attained by a random rigid transformation of the target. We vary $\beta\in[0.5, 0.9]$ with $\Delta \beta=0.05$ and set the same value for PIPL and the proposed method. As observed in the left panel of Fig.~\ref{fig:hand}, with $\beta$ increasing, PIPL and \revise{our method} achieve higher registration accuracy. However, the proposed method outperforms PIPL by a large margin, especially under small $\beta_s$, due to the incorporation of local surface variation and geometric features into the force modeling. Meanwhile, our method has a broader convergence basin ($\beta\in [0.7, 0.9]$) than PIPL ($\beta \in [0.85, 0.9]$), suggesting a faster registration process.

\subsection{Validity of the Proposed Geometric Invariant}
\begin{figure}[t]
	\centering
	\subcaptionbox{Initialization}{
		\includegraphics[width=0.115\textwidth]{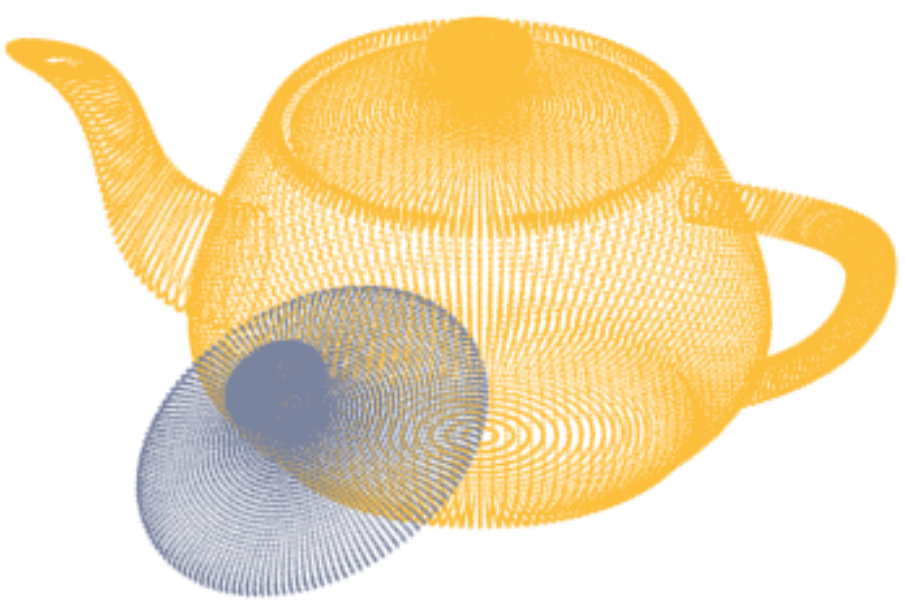}
	}
	\subcaptionbox{Super4PCS}{
		\includegraphics[width=0.10\textwidth]{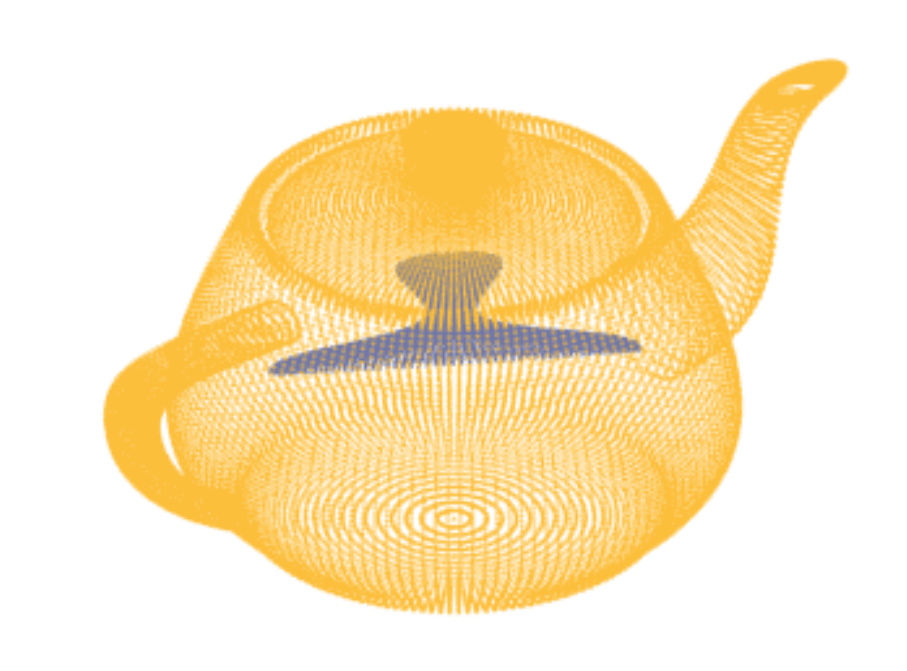}
	}
	\subcaptionbox{PIPL}{
		\includegraphics[width=0.10\textwidth]{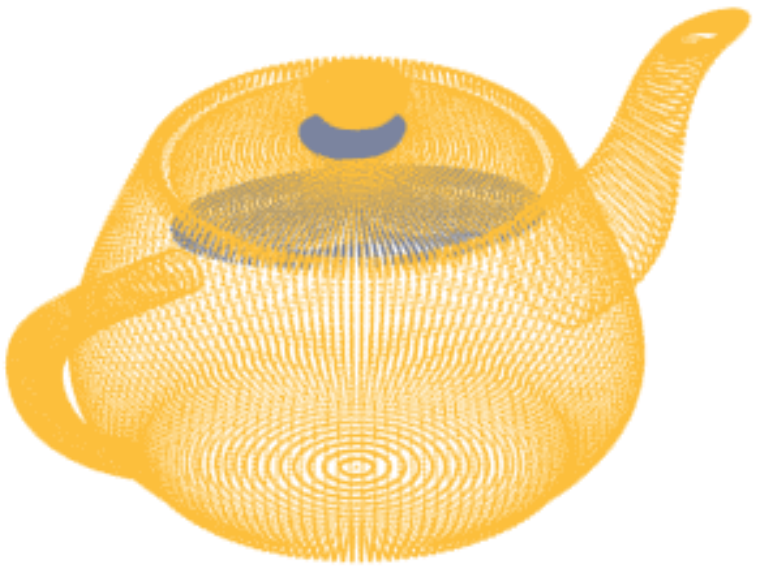}
	}
	\subcaptionbox{Ours}{
		\includegraphics[width=0.10\textwidth]{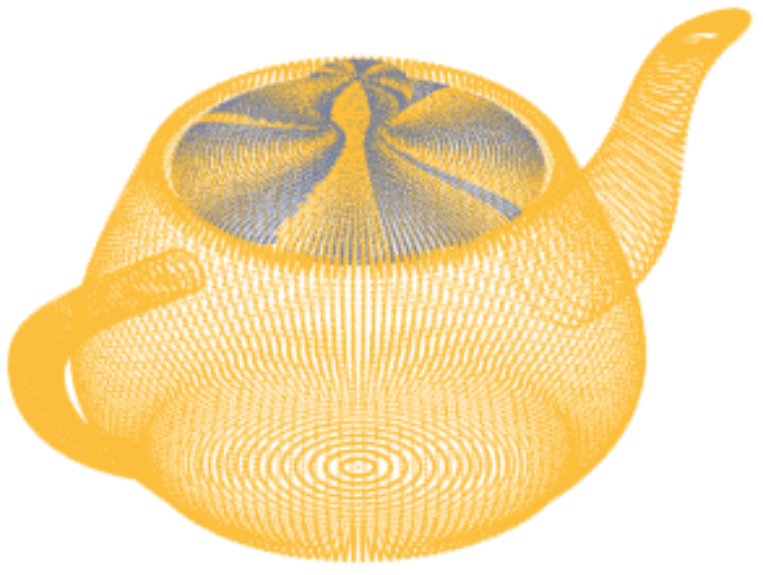}
	}
	\vskip -0.2cm
	\caption{Verification of the proposed geometric invariant $\mathcal{V}_g(\bm{x})$. Our proposed method attains higher registration accuracy and speed than Super4PCS and PIPL. }
	\label{fig:teaPot}
\end{figure}

To evaluate the validity of the proposed geometric invariant $\mathcal{V}_g(\bm{x})$ in our method, we utilize the \textit{Utah Teapot} dataset 
and compare it with the baseline methods, \ie,  PIPL and Super4PCS (implemented from the OpenGR C++ library~\cite{OpenGR}). As shown in Fig.~\ref{fig:teaPot}(a), our objective is to align the pot cover (source) to the teapot (target), where the pot body is seen as outliers. {Considering that the overlap ratio between the source and the target is relatively small, we make the overlap setting $O$ (a hyperparameter) of Super4PCS belong to $\{0.20, 0.50\}$. $O=0.20$ represents a relatively conservative overlap rate, whereas $O=0.50$ gives a more confident estimation.} To be statistically representative, we vary $\beta\in\{0.9, 0.95\}$, perform 100 trials and measure the average deviation and runtime, as reported in Table~\ref{tab:teaPot}. Super4PCS and PIPL show significant errors as the ambiguous outliers disturb their correspondence between point pairs. \revise{By contrast}, our method attains substantially accurate registration because the geometric invariant $\mathcal{V}_g(\bm{x})$ provides more insights \revise{into} the point clouds' feature distribution than the single point location, hence enabling more correct correspondence. \revise{In addition}, with the priori overlap threshold $O$ increasing, \revise{Super4PCS consumes more time but without accuracy enhancement},  whereas our method outperforms Super4PCS and PIPL \revise{in terms of} speed. We present a test sample in Fig.~\ref{fig:teaPot}(c) and  provide a demo in Supplemental Material to exhibit this dynamical registration process.

\begin{table}[t]
	\centering
	\renewcommand\arraystretch{1.2}

	\setlength{\tabcolsep}{2mm}{
		\begin{tabular}{|c|c|c|c|c|}
			\cline{1-5}
			\multirow{2}{*}{\diagbox{\textbf{Method}}{\textbf{Metric}}} & \multicolumn{2}{c|}{$\beta=0.90$ $(O=0.20)$} & \multicolumn{2}{c|}{$\beta=0.95$ $(O=0.50)$}\\ \cline{2-5}
			&AngErr ($^\circ$)&Time (s)&AngErr ($^\circ$)&Time (s)\\
			\cline{1-5}
			Super4PCS~\cite{mellado2014super}&40.4707&0.6690&40.4646&1.3258\\ \cline{1-5}
			PIPL~\cite{jauer2018efficient}&50.8282&0.4460&49.9012&0.6085\\ \cline{1-5}
			Ours&\cellcolor{best}{0.0004}&\cellcolor{best}{0.3210}&\cellcolor{best}{0.0003}&\cellcolor{best}{0.2960}\\
			\cline{1-5}
		\end{tabular}
	}
	\caption{Statistic of the average angle error and runtime on the Utah TeaPot dataset.}
	\label{tab:teaPot}
	\vskip -0.3cm
\end{table}
\begin{figure}[!htbp]
	\centering
	\includegraphics[width=0.4\textwidth]{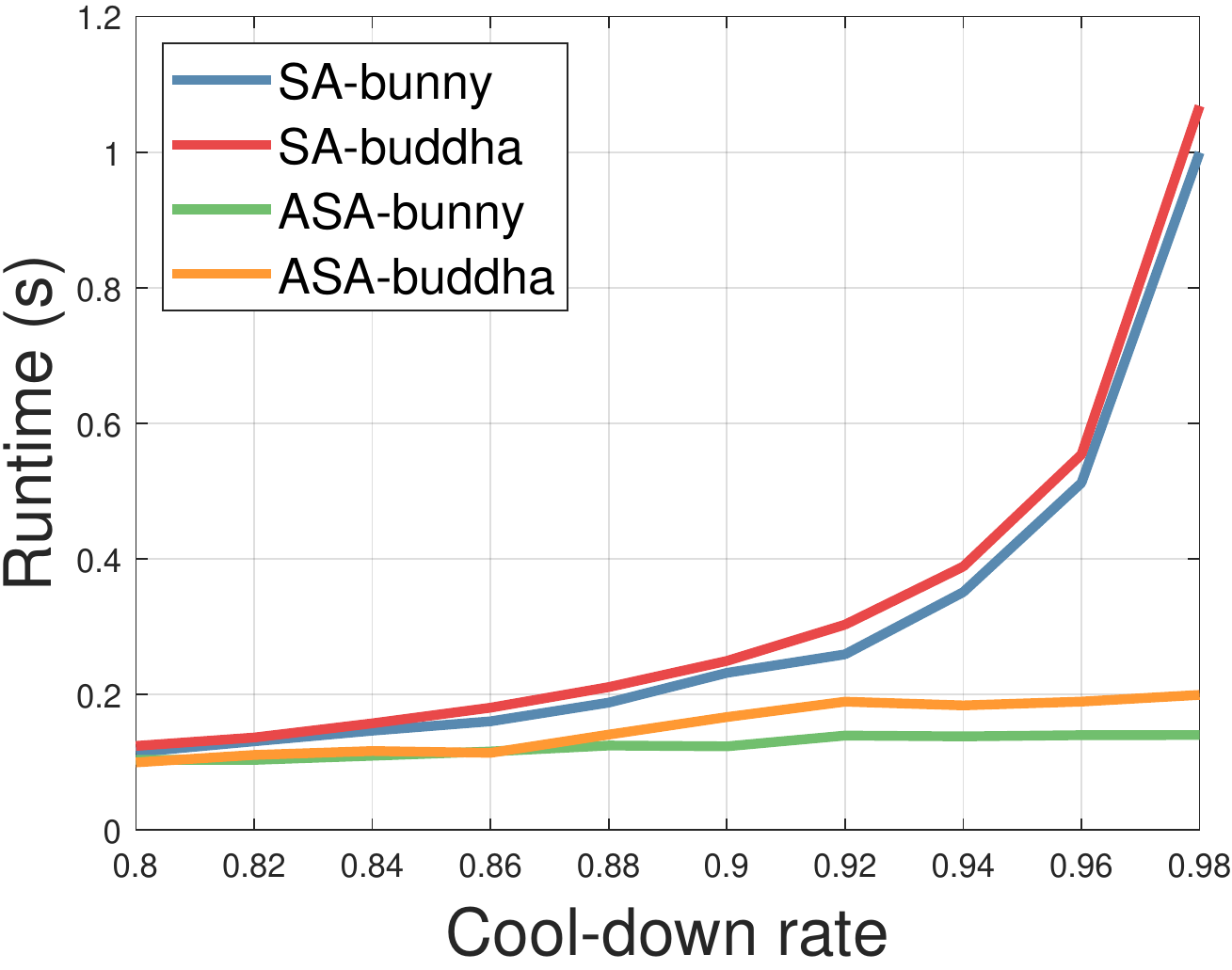}
	\vskip -0.2cm
	\caption{Runtime comparison of the simulated annealing (SA) and the adaptive simulated annealing (ASA) algorithms for point cloud registration. ASA is much faster and more stable than SA, especially under large cool-down rates.}
	\vskip -0.3cm
	\label{SA}
\end{figure}

\begin{figure}[b]
	\includegraphics[width=0.23\textwidth]{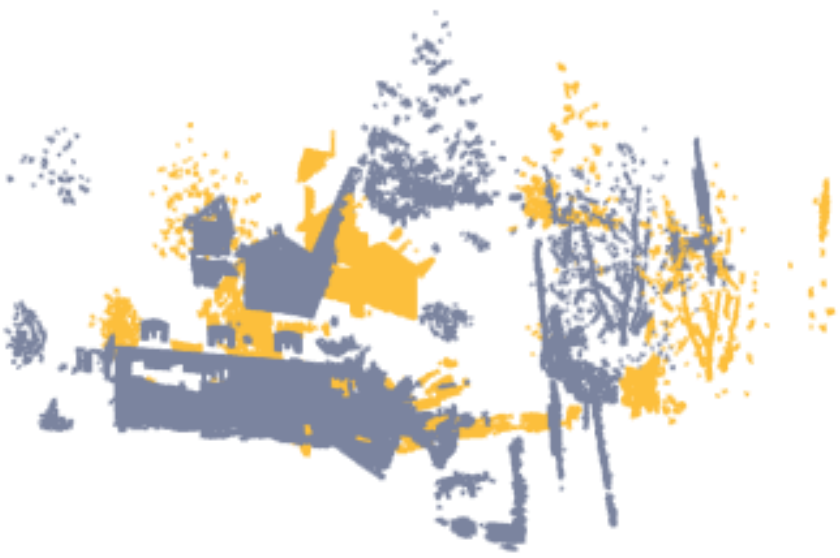}
	\includegraphics[width=0.236\textwidth]{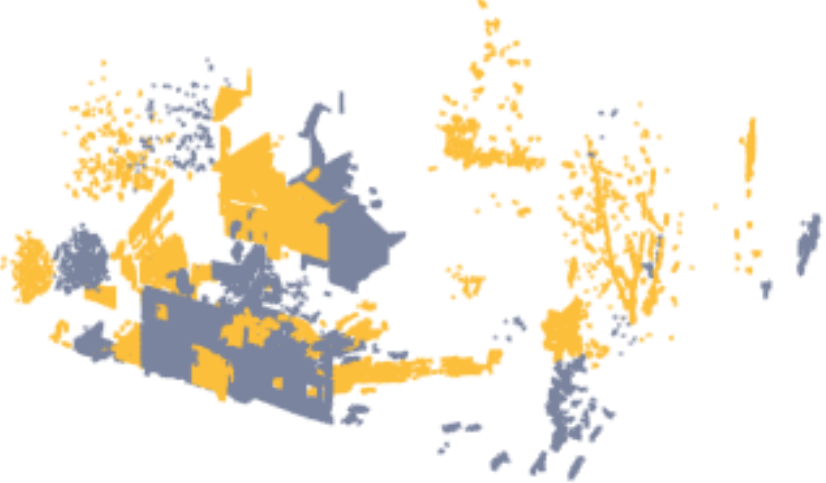}

	\caption{Examples of the large-scale terrestrial laser scanning point clouds captured from different perspectives.}
	
	\label{fig:ETHZ_example}
	\vskip -0.3cm
\end{figure}

\subsection{Comparison of SA \& ASA}
To test the acceleration effect of the introduced adaptive simulated annealing (ASA) method for point cloud registration, we further compare it with the simulated annealing (SA) algorithm. For this purpose, we replace the global optimization part \revise{with} SA and keep the others fixed in our method. SA requires a low or cooling temperature $\texttt{Epsilon}$ to judge the convergence and stop the algorithm iteration. We adopt the typical default setting {\color{black}{{{$\texttt{Epsilon=1e-5}$}}}} for the test. \revise{By contrast}, ASA uses the \texttt{maxNumIter} as the termination condition, where we set \texttt{maxNumIter=100}.

\begin{table*}[t]

	\centering
	\renewcommand\arraystretch{1.3}

	\caption{Quantitative results on the ETH Facade dataset. \textbf{Green} and \textbf{red} fonts indicate the first two best accuracy.}
	\setlength{\tabcolsep}{0.65mm}{
		\begin{tabular}{|c|l|c|c|c|c|c|c|c|c|c|c|c|}
			\cline{1-13}
			\diagbox{\textbf{Dataset}}{\textbf{Method}}&Metric&ICP\cite{besl1992method} & ICP-KP\cite{kjer2010evaluation}&ICP-KN\cite{kjer2010evaluation}&GMM\cite{jian2010robust}&CPD\cite{myronenko2010point}&ECM\cite{horaud2010rigid}&TEASER++\cite{yang2020teaser}&FGR\cite{zhou2016fast}&FRICP\cite{zhang2021fast}&PIPL\cite{jauer2018efficient}&Ours\\
			\hline
			\multirow{2}{*}{Pair 1}&AngErr&11.6826	&12.7025	&9.1673	&20.1567	&{2.6700}	&5.9759	&\cellcolor{second}{1.8652}&4.2284&{2.4306}&4.1883	&\cellcolor{best}{1.6100}\\ \cline{2-13}
			&Time&0.3285&3.8016&3.9138&38.4033&47.0350&39.6100&1.193&0.5711&0.8287&0.4504&0.3584\\ \cline{1-13}
			\multirow{2}{*}{Pair 2}&AngErr&2.8705&	2.9106&	4.0623&	4.7727&	2.5726&	0.3724&0.8146&1.6893&2.8869&	\cellcolor{second}{0.1031}&	\cellcolor{best}{0.0839}\\ \cline{2-13}
			&Time&0.2226&2.8928&2.9664&42.4362&26.9683&30.8803&0.7606&0.3654&0.7960&0.5380&0.3357\\ \cline{1-13}
			\multirow{2}{*}{Pair 3}&AngErr&5.4316&	5.4488&	9.4825&	6.9385&	13.4760&	11.9003&{2.3320}&\cellcolor{second}{1.6398}&\cellcolor{best}{0.4207}&	{3.9591}&	{1.9209}\\ \cline{2-13}
			&Time&0.2301&2.9445&3.0095&68.5586&24.2632&31.8921&0.9601&0.4571&0.7540&0.5303&0.3415\\ \cline{1-13}
			\multirow{2}{*}{Pair 4}&AngErr&{2.0054}&	{2.0283}&	3.3633&	5.9180&	9.9122&	12.2728&5.3606&\cellcolor{best}{0.8089}&2.7742&	10.1872&	\cellcolor{second}{1.6625}\\ \cline{2-13}
			&Time&0.2156&2.3750&2.4604&38.6081&37.4666&26.2225&0.7401&0.3609&0.5212&0.4461&0.3419\\ \cline{1-13}
			\multirow{2}{*}{Pair 5}&AngErr&6.0332&	6.0734&	7.2193&	20.9932&	8.1589	&7.3281&{4.2966}&{2.1304}&\cellcolor{second}{1.8182}&	{5.9874}&	\cellcolor{best}{1.3293}\\ \cline{2-13}
			&Time&0.2349&2.2919&2.3802&23.7734&18.4693&25.5012&0.8913&0.4588&0.7080&0.4495&0.3412\\ \cline{1-13}
		\end{tabular}
	}

	\label{tab:ETHZ}
	\vskip -0.3cm
\end{table*}

\begin{figure*}[t]
	\centering

	\subcaptionbox{ICP~(10.2077)}{
		\includegraphics[width=0.223\textwidth]{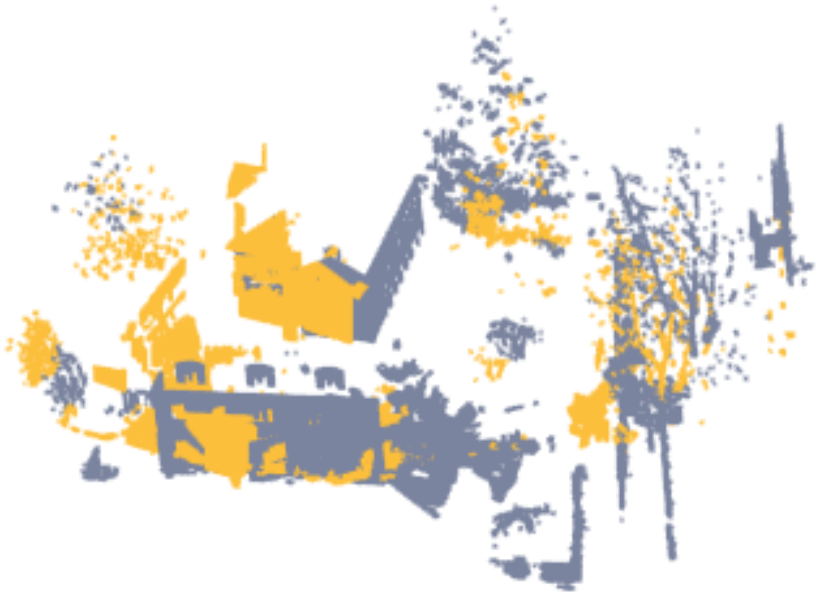}
	}
	\subcaptionbox{ICP-KP~(5.7161)}{
		\includegraphics[width=0.23\textwidth]{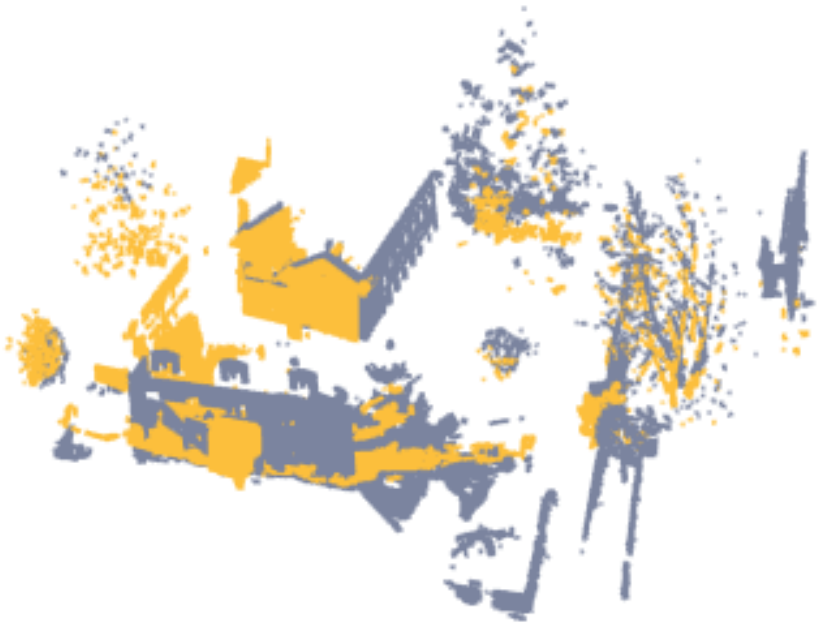}
	}
	\subcaptionbox{ICP-KN~(26.4846)}{
		\includegraphics[width=0.23\textwidth]{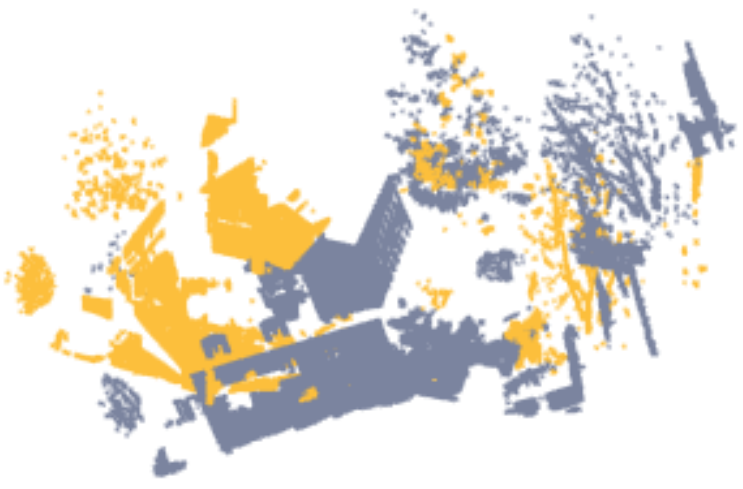}
	}
	\subcaptionbox{GMM~(15.3665)}{	
		\includegraphics[width=0.23\textwidth]{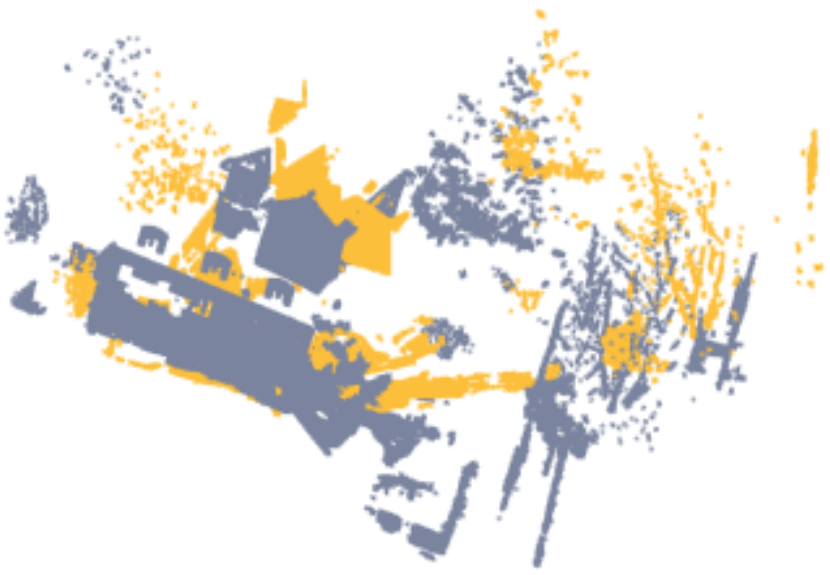}
	}

	\subcaptionbox{CPD~(2.5504)}{
		\includegraphics[width=0.23\textwidth]{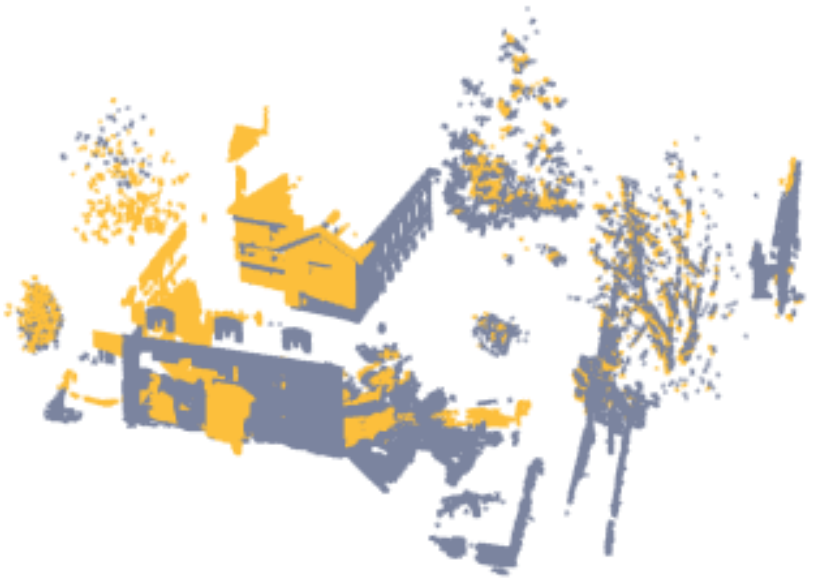}
	}
	\subcaptionbox{ECM~(2.9128)}{
		\includegraphics[width=0.223\textwidth]{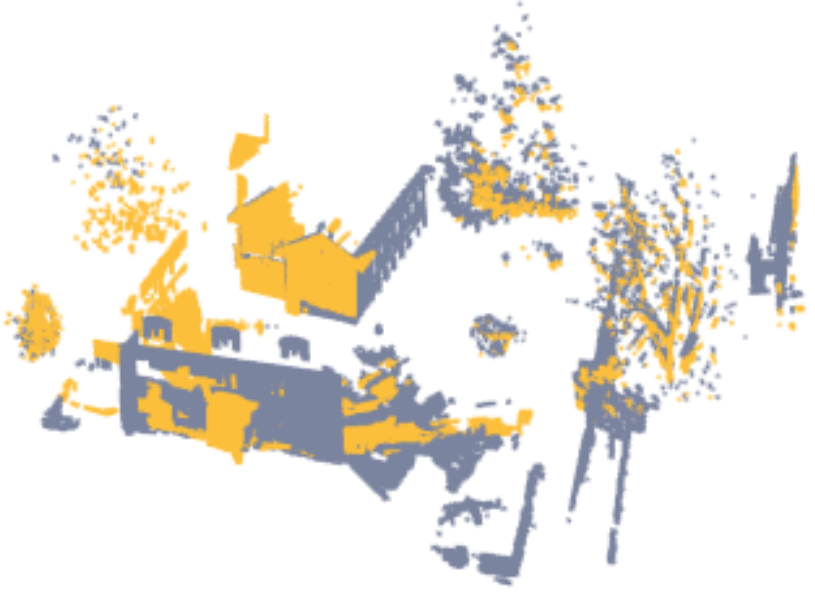}
	}
	\subcaptionbox{\revise{TEASER++~(3.0316)}}{
		\includegraphics[width=0.23\textwidth]{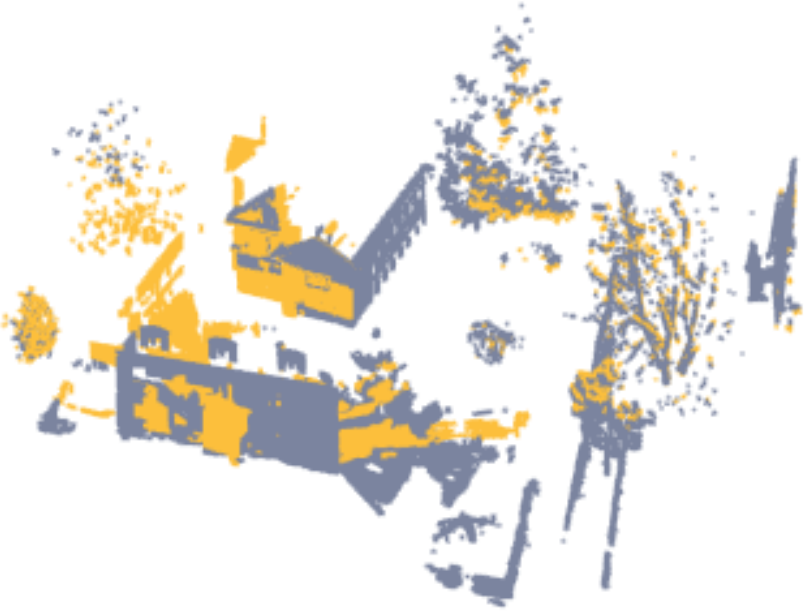}
	}
	\subcaptionbox{\revise{FGR~(3.1425)}}{
		\includegraphics[width=0.23\textwidth]{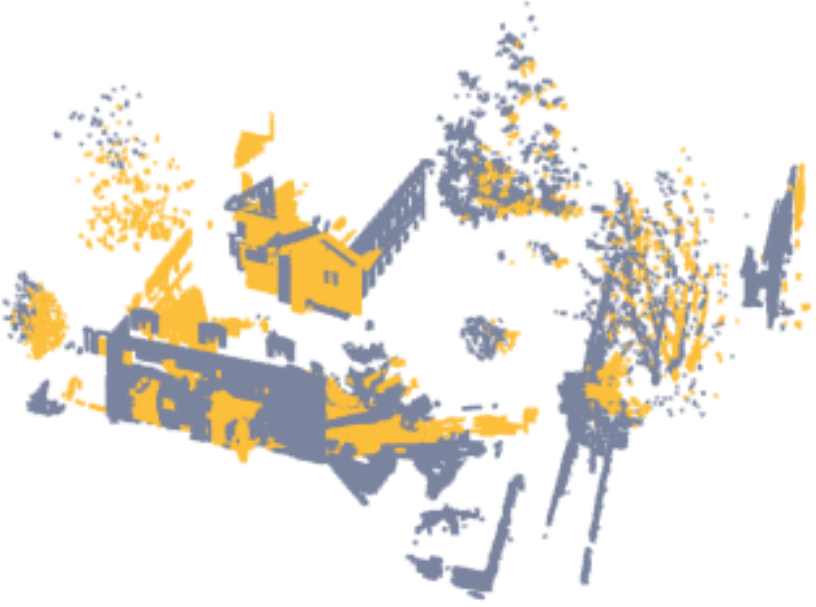}
	}
	\subcaptionbox{\revise{FRICP~(2.4328)}}{
		\includegraphics[width=0.238\textwidth]{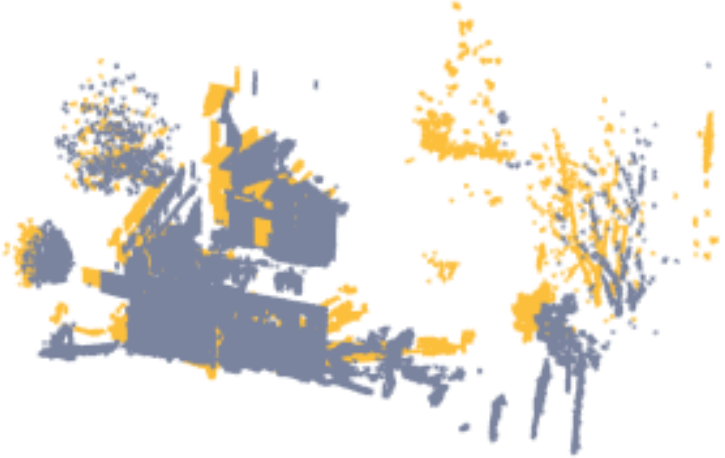}
	}
	\subcaptionbox{PIPL~(5.8498)}{	
		\includegraphics[width=0.23\textwidth]{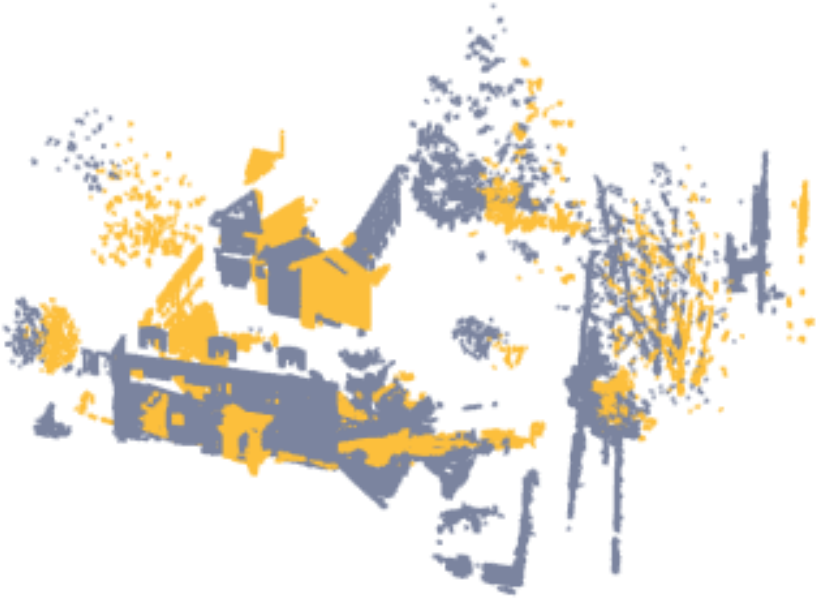}
	}
	\subcaptionbox{Ours (2.3639)}{
		\includegraphics[width=0.23\textwidth]{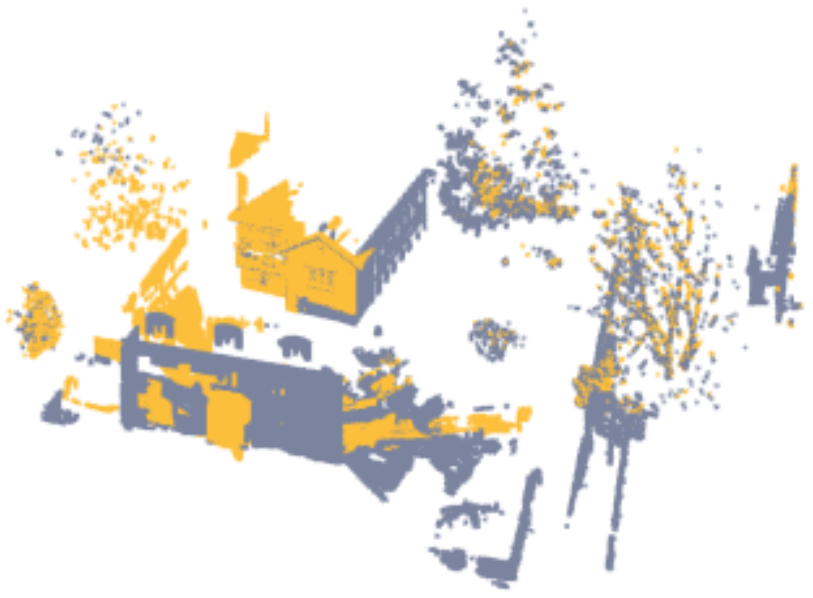}
	}

	\caption{\revise{Qualitative results on the TLS Facade dataset}. The proposed method attains the highest precision than competitors.}
	\vskip -0.3cm
	\label{fig:ETHZEx}
\end{figure*}

\begin{figure}[!htbp]

	\centering
	\subcaptionbox{}{
		\includegraphics[width=0.21\linewidth]{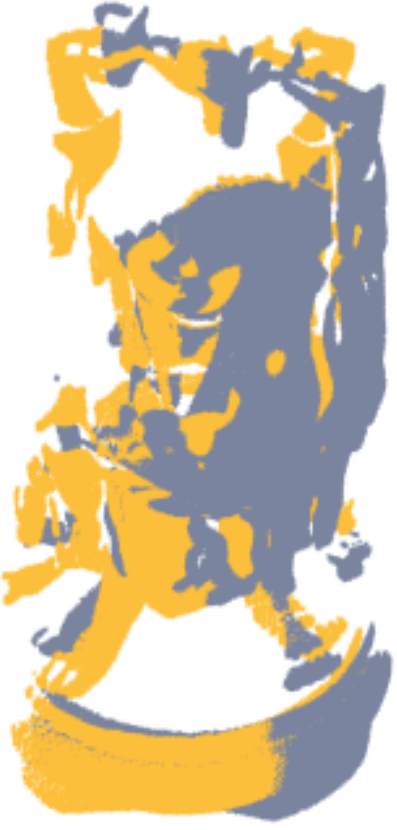}
	}
	\subcaptionbox{}{
		\includegraphics[width=0.21\linewidth]{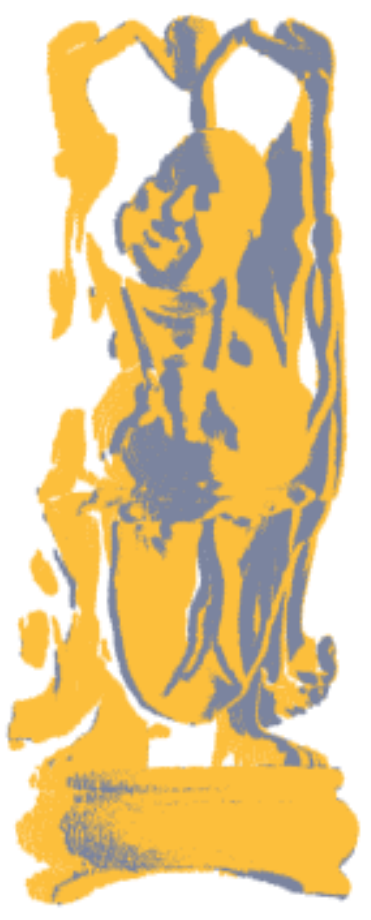}
	}
	\subcaptionbox{}{
		\includegraphics[width=0.21\linewidth]{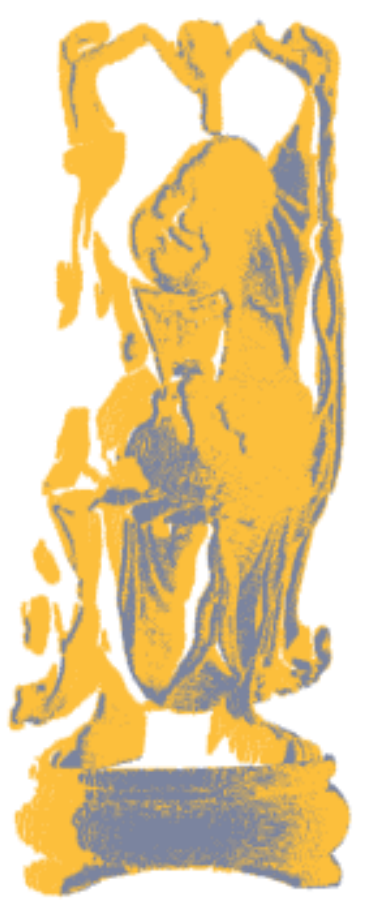}
	}
	\subcaptionbox{}{
		\includegraphics[width=0.21\linewidth]{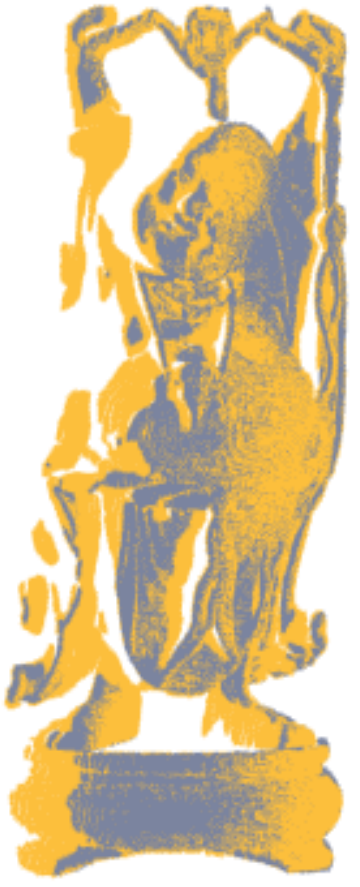}
	}
	\vskip -0.2cm
	\caption{(a) The input two misalignment point clouds. (b)--(c) Registration results of SA and ASA with AngErr=\textbf{5.2386} and AngErr=\textbf{0.4301}, separately. (d) The ground truth alignment.}

	\label{fig:Buddha}
	\vskip -0.3cm
\end{figure}

We use the Bunny and the Buddha datasets from~\cite{DataStanford} for comparison, where the point number of the source and target point clouds of Buddha is 78,056 and 75,582, respectively. To be fair, we vary the cool-down rate $\beta\in [0.8, 0.98]$ with interval $\Delta \beta=0.02$ and report the average results of 10 trials. The average runtime is reported in Fig.~\ref{SA}. \revise{As shown in the figure}, SA consumes more time than ASA for both datasets, and with $\beta$ increasing, its runtime grows significantly. However, ASA \revise{remains} relatively stable due to the adaptive tuning scheme of the temperature, and its time consumption \revise{remains} below $0.2s$ even at large $\beta$. We present the Buddha registration results after algorithm convergence in Fig.~\ref{fig:Buddha}.

\begin{table*}[t]

	\centering
	\renewcommand\arraystretch{1.3}
	
	\caption{Statistic of the registration results on KITTI odemetry datasets, where the average deviations of the relative rotation (Rel. Rot.), relative translation (Rel. Tran.), absolute rotation (Abs. Rot.), absolute translation (Abs. Tran.), last frame rotation (Last Rot.), and last frame translation (Last Tran.) are reported. Values inside the parenthesis indicate the maximum. {\color{blue}{Blue}} and {\color{orange}{orange}} fonts indicate the first two highest accuracy.
	}
	\setlength{\tabcolsep}{2.7mm}{
		\begin{tabular}{|c|c|c|c|c|c|c|}
			\cline{1-7}
			\diagbox{\textbf{Method}}{\textbf{Metric}}&Rel. Rot. ($^\circ$)&Rel. Tran. (m) & Abs. Rot. ($^\circ$)&Abs. Tran. (m)&Last Rot. ($^\circ$)&Last Tran. (m)\\ \cline{1-7}
			ICP\cite{besl1992method}&0.2850 (2.6499)&0.2978 (5.4468)&21.9215 (39.8268)&79.6718 (213.1728)&39.8268&213.1728\\ \cline{1-7}
			ICP-KP~\cite{kjer2010evaluation}&0.2830 (2.5197)&0.2930 (3.6881)&21.9874 (38.9154)&79.7293 (212.3571)&38.7033 &212.3571\\ \cline{1-7}
			ICP-KN~\cite{kjer2010evaluation}&1.4763 (9.3142)&0.3019 (5.0016)&139.0011 (179.7096)&190.6753 (496.8816)&126.6452&496.8816\\ \cline{1-7}
			GMM~\cite{jian2010robust}&0.3326 (1.0954)&2.9138 (4.7576)&25.8011 (40.7604)&227.1595 (497.6106)&40.093&497.6106\\ \cline{1-7}
			CPD~\cite{myronenko2010point}&{0.2092} (\second{1.0069})&0.2161 (\second{1.0037})&13.7057 (24.7993)&35.6197 (95.2179)&24.4202&94.8900\\ \cline{1-7}
			ECM~\cite{besl1992method}&\second{0.1264} ({2.5036})&0.1912 (3.6157)&{5.3706} ({12.2503})&19.6340 (41.4195)&12.2503&41.2022\\\cline{1-7}
			TEASER++~\cite{yang2020teaser}&0.3707 ({5.6199})&{0.0976} (1.3854)&{46.4931} ({179.3500})&37.2031 (171.8366)&{173.8888}&{171.7321}\\\cline{1-7}
			FGR~\cite{zhou2016fast}&0.4218 ({2.4082})&{0.2392} (1.0075)&{12.1925} ({25.4476})&48.4584 (123.3828)&{24.0846}&{123.3828}\\\cline{1-7}
			FRICP~\cite{zhang2021fast}&0.1527 ({7.9674})&{0.2191} (30.0883)&{4.1370} ({10.6699})&19.0893 (50.8703)&\second{5.3265}&{50.8703}\\\cline{1-7}
			PIPL~\cite{jauer2018efficient}&0.1419 ({2.1346})&\second{0.0799} (6.7271)&\second{4.1124} (\second{6.6670})&\second{14.5802} (\second{36.1614})&{6.1962}&\second{36.1614}\\\cline{1-7}
			Ours&\best{0.1017} (\best{0.6565})&\best{0.0422} (\best{0.5347})&\best{2.0542} (\best{3.2864})&\best{5.6745} (\best{13.2796})&\best{2.9099}&\best{13.2796}\\
			\cline{1-7}
		\end{tabular}
	}
	
	\label{tab:kitti}
	\vskip -0.3cm
\end{table*}

\begin{figure*}[t]
	\centering
	\subcaptionbox{ICP}{
		\includegraphics[width=0.223\textwidth]{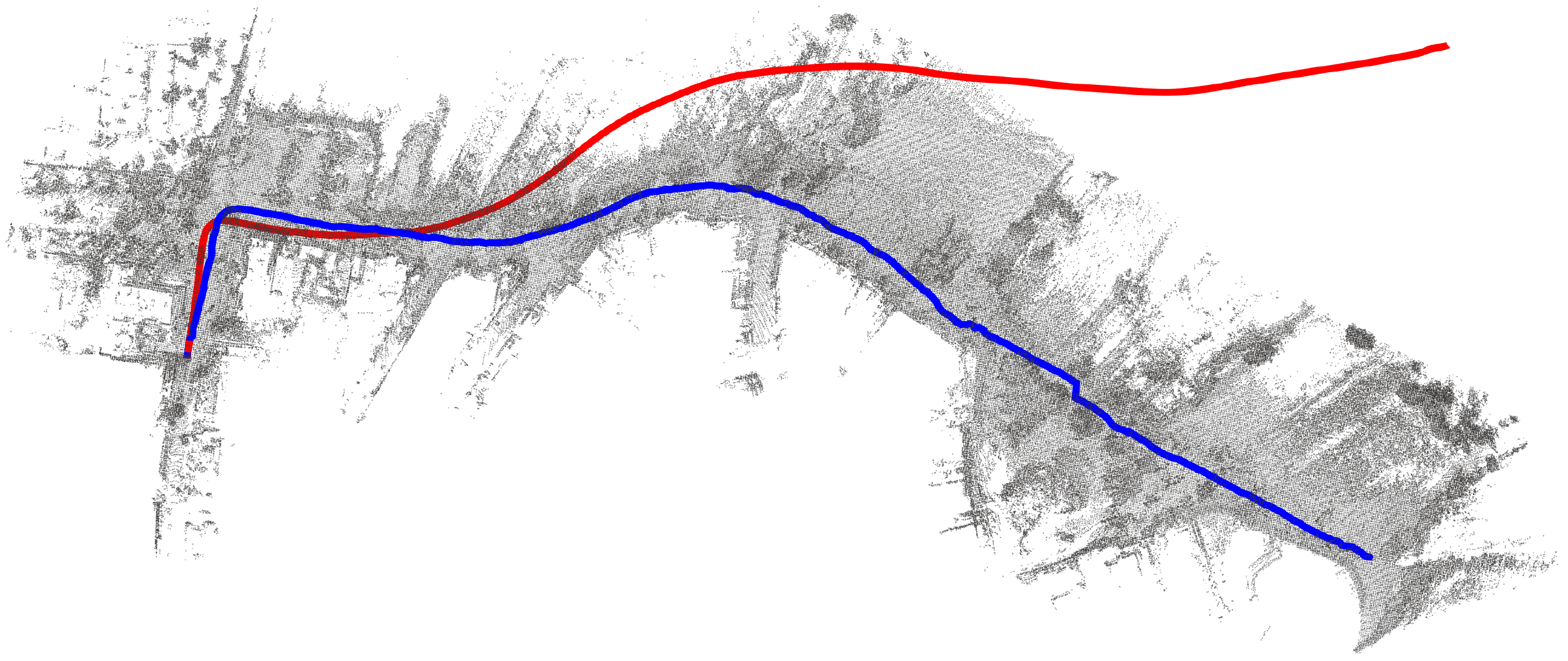}
	}
	\subcaptionbox{ICP-KP}{ 
		\includegraphics[width=0.19\textwidth]{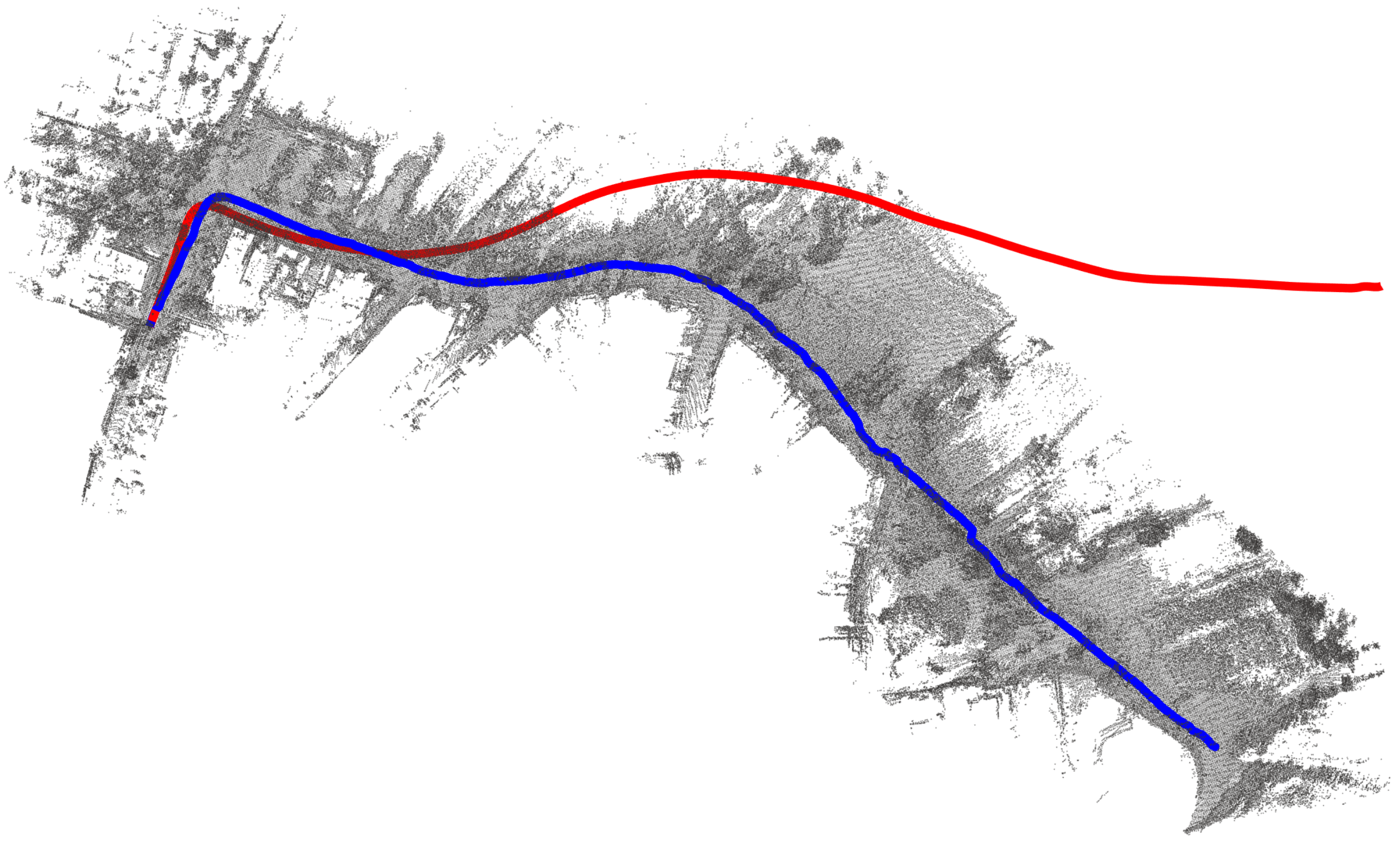}
	}
	\subcaptionbox{ICP-KN}{
		\includegraphics[width=0.23\textwidth]{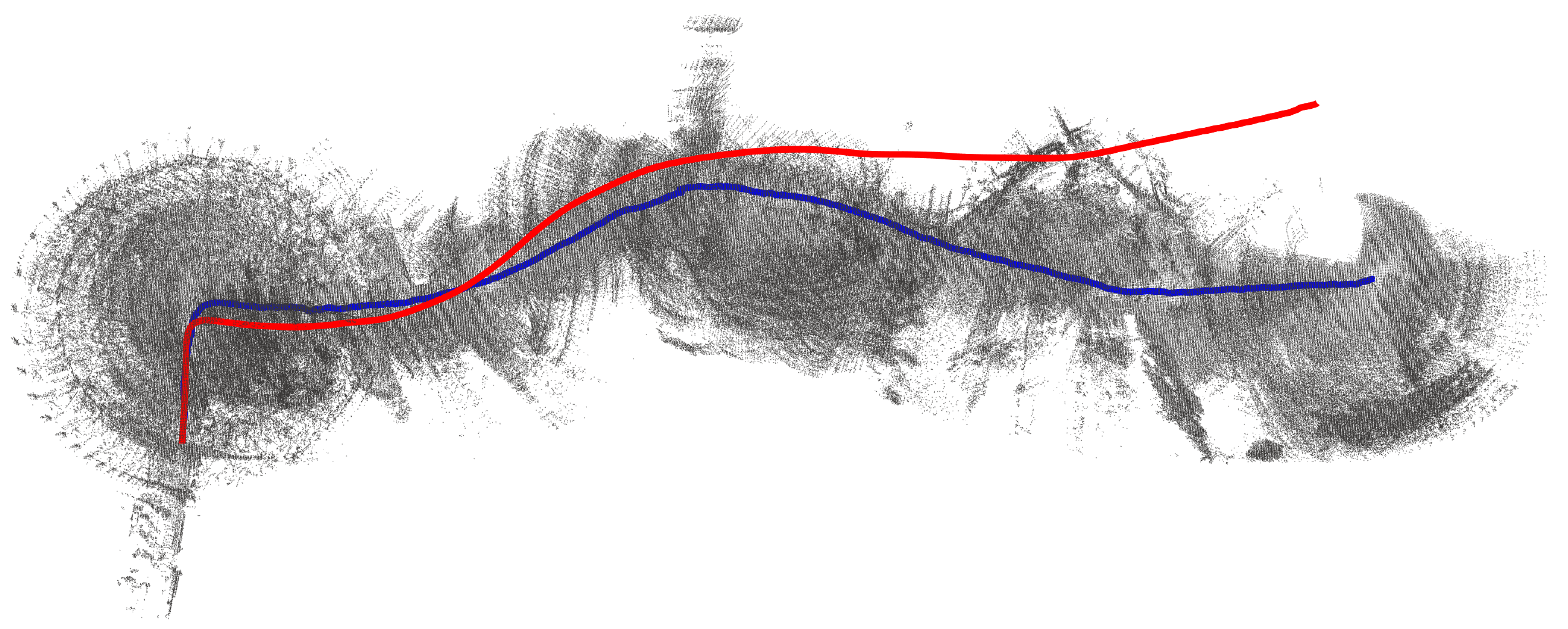}
	}
	\subcaptionbox{GMM}{	
		\includegraphics[width=0.23\textwidth]{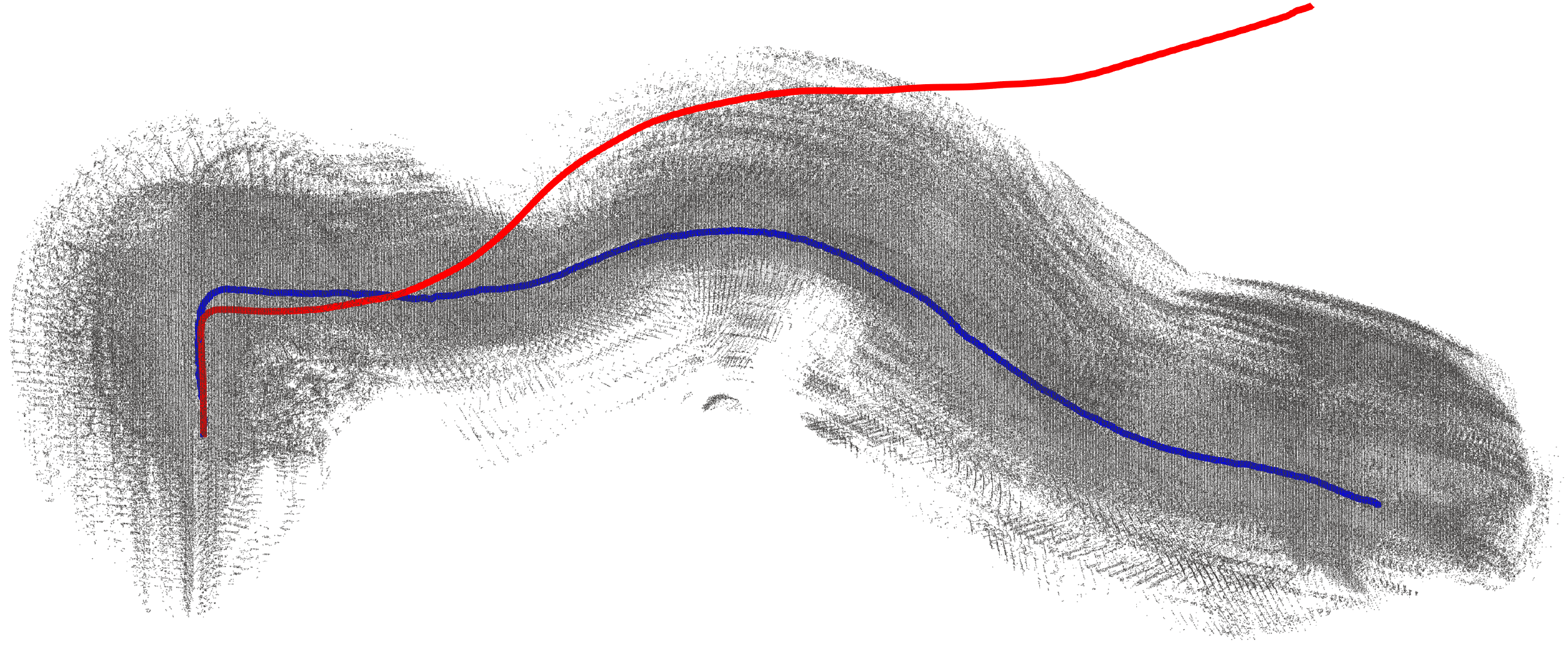}
	}

	\subcaptionbox{CPD}{
		\includegraphics[width=0.23\textwidth]{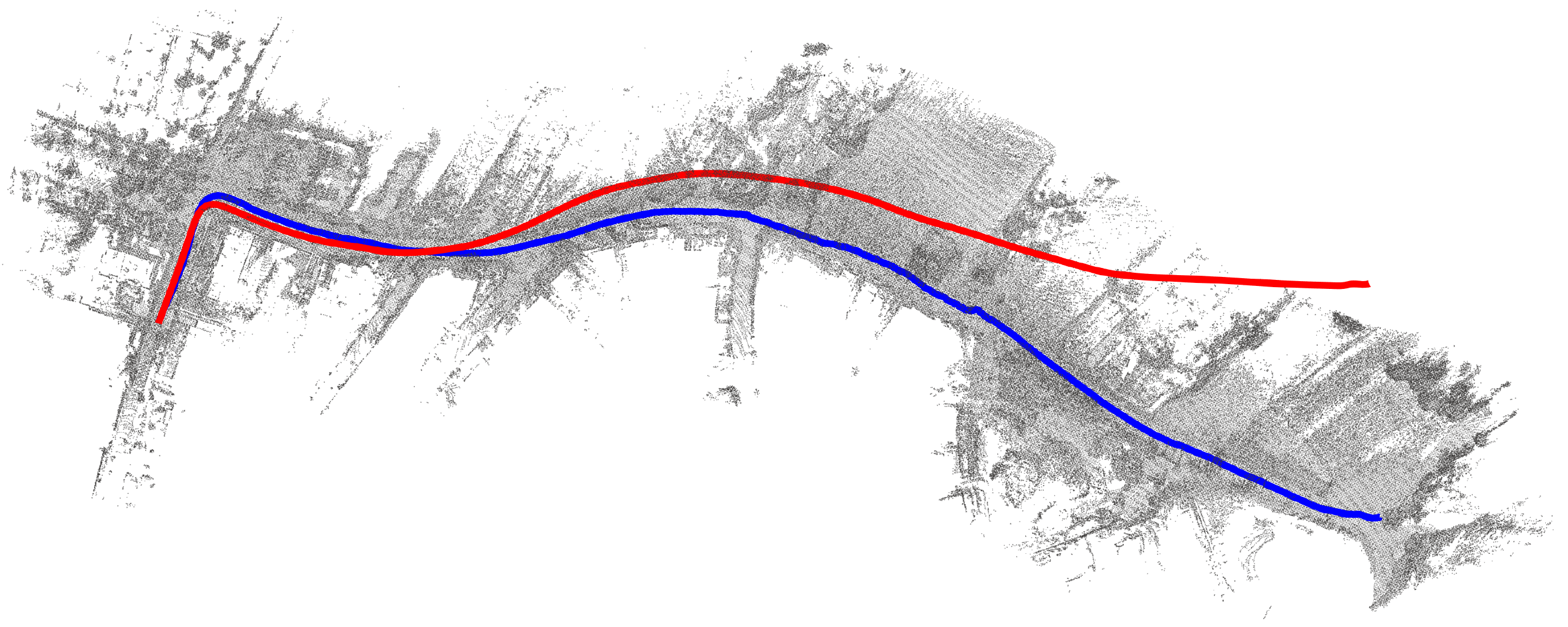}
	}
	\subcaptionbox{ECM}{
		\includegraphics[width=0.223\textwidth]{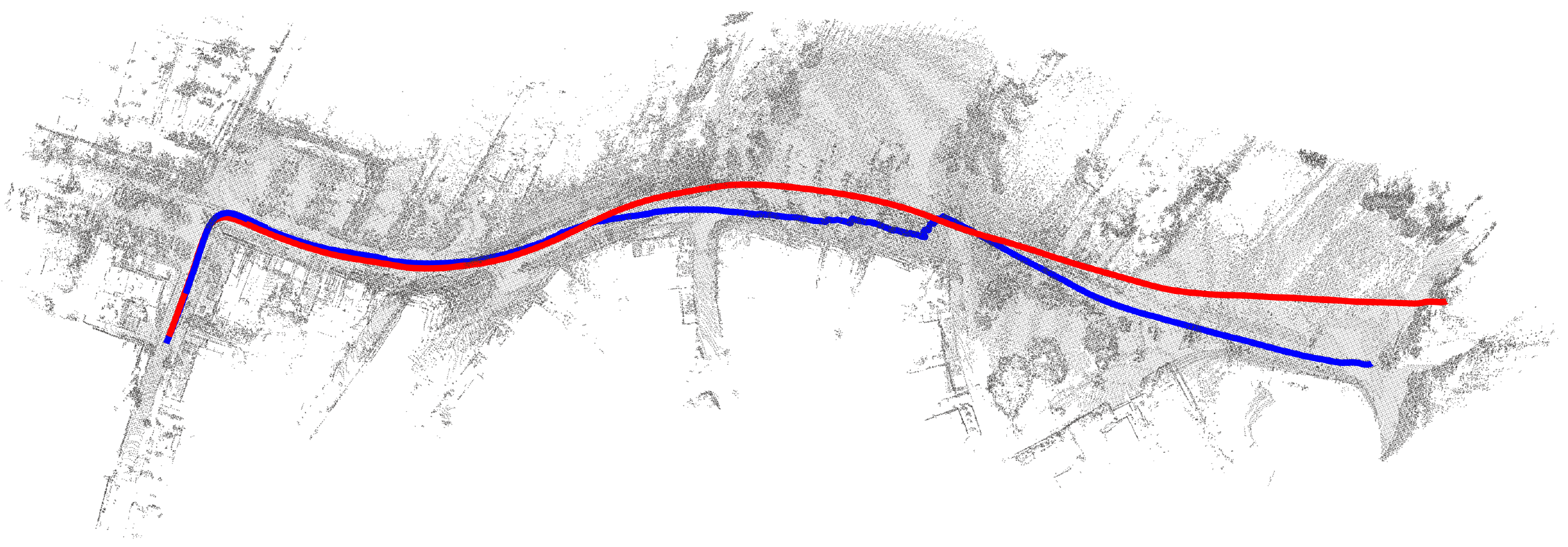}
	}
	\subcaptionbox{\revise{TEASER++}}{
		\includegraphics[width=0.223\textwidth]{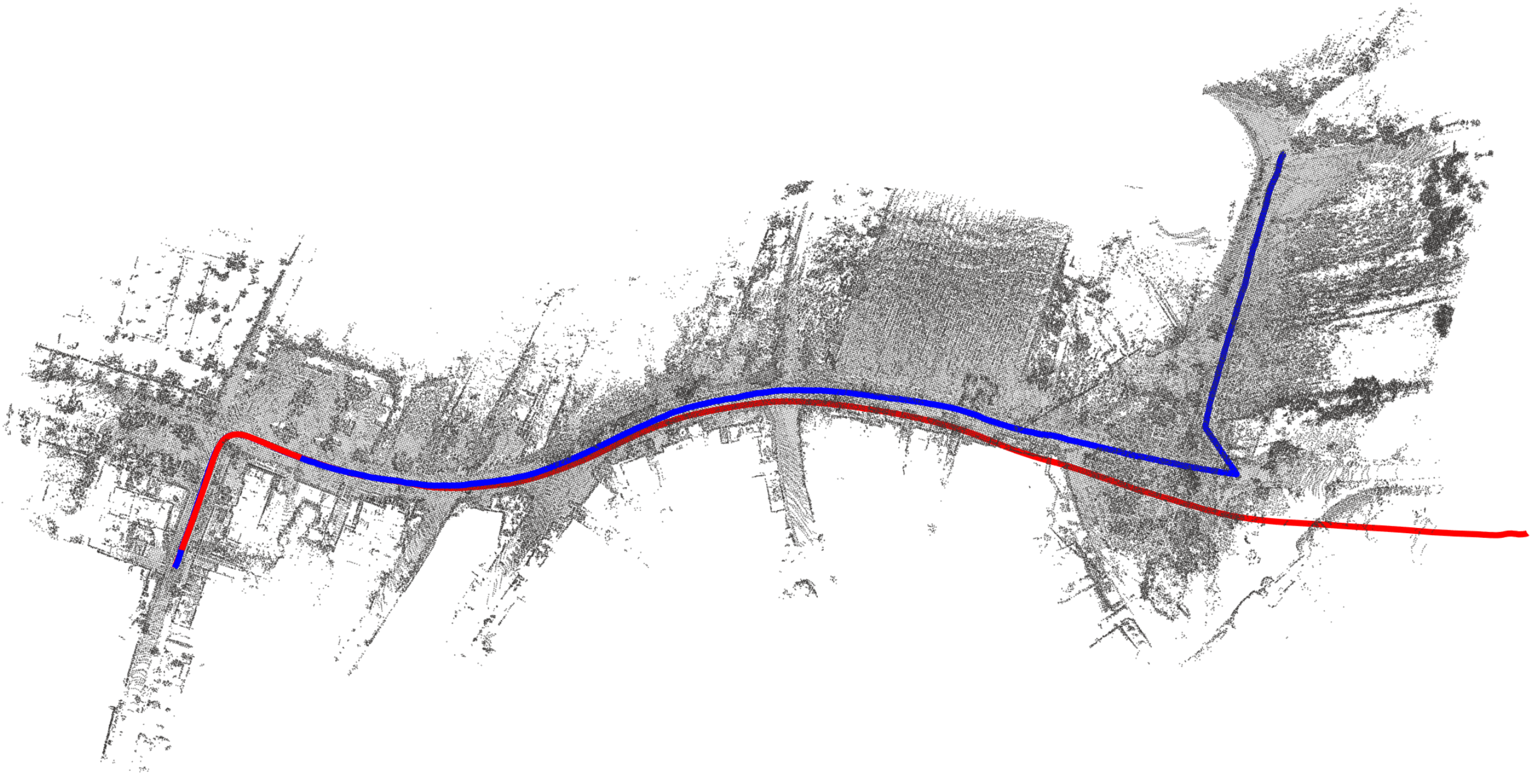}
	}
	\subcaptionbox{\revise{FGR}}{
		\includegraphics[width=0.223\textwidth]{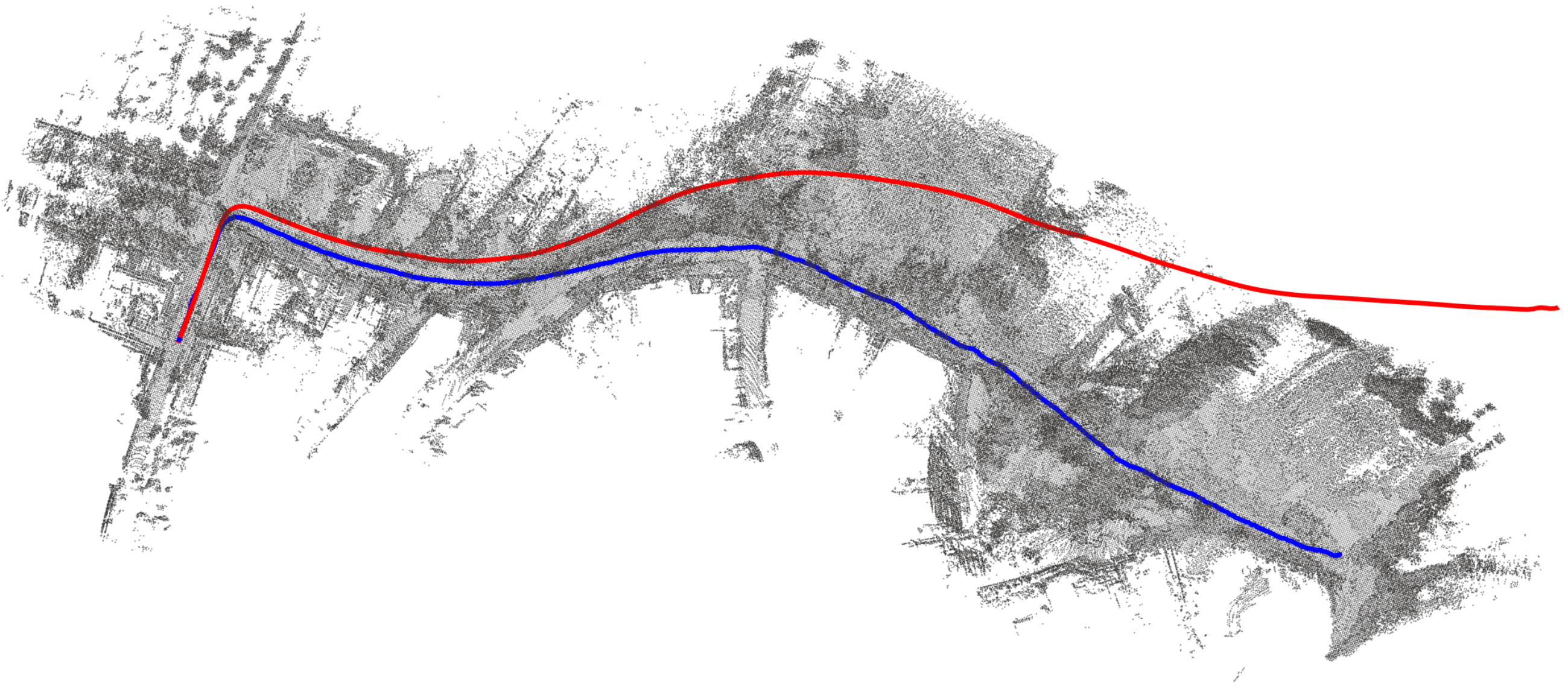}
	}
	\subcaptionbox{\revise{FRICP}}{
	\includegraphics[width=0.223\textwidth]{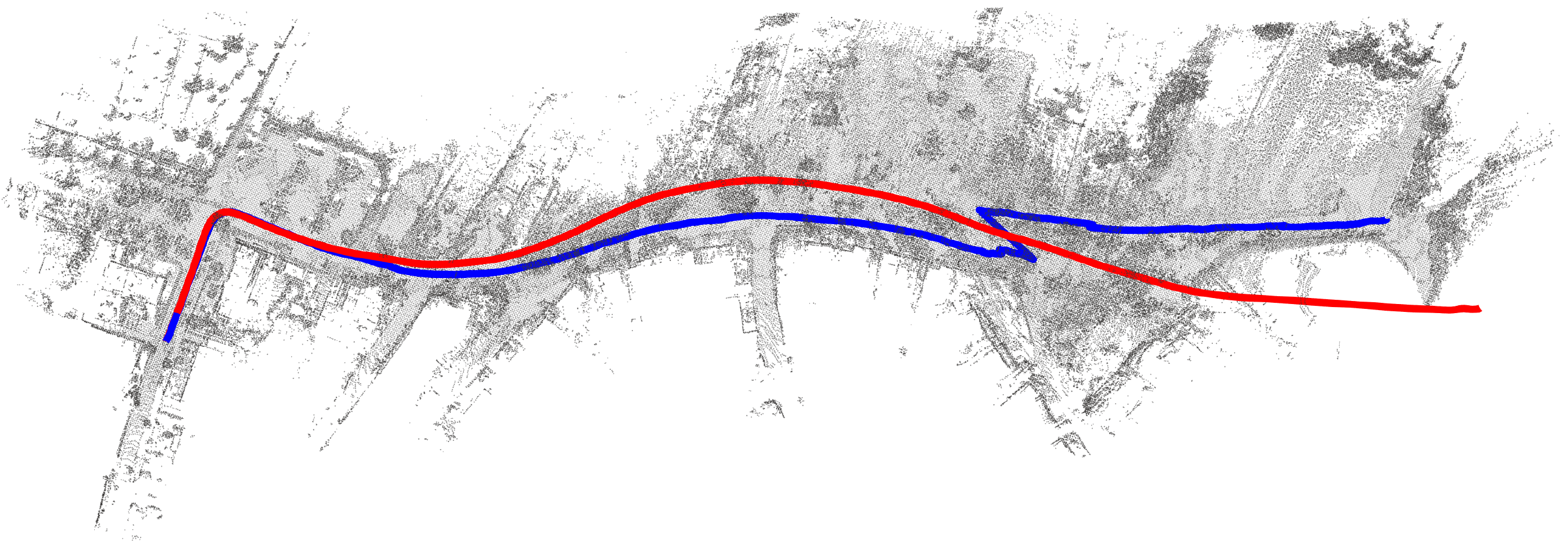}
	}
	\subcaptionbox{PIPL}{	
		\includegraphics[width=0.23\textwidth]{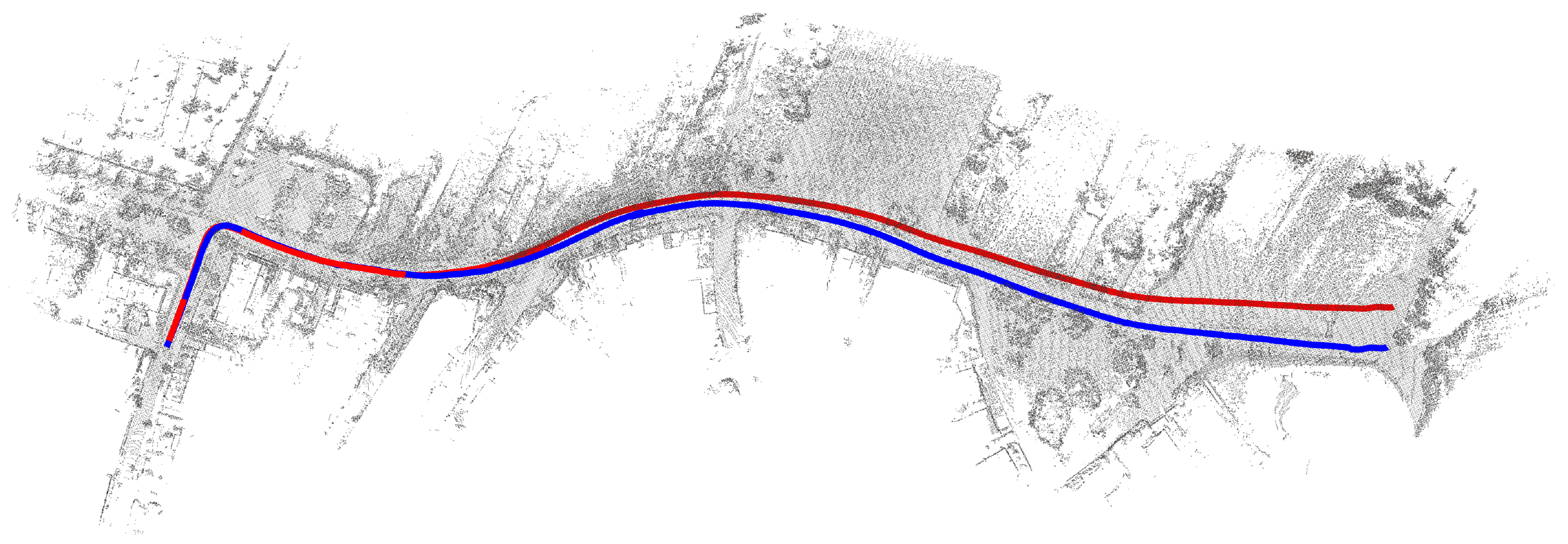}
	}
	\subcaptionbox{Ours}{
		\includegraphics[width=0.23\textwidth]{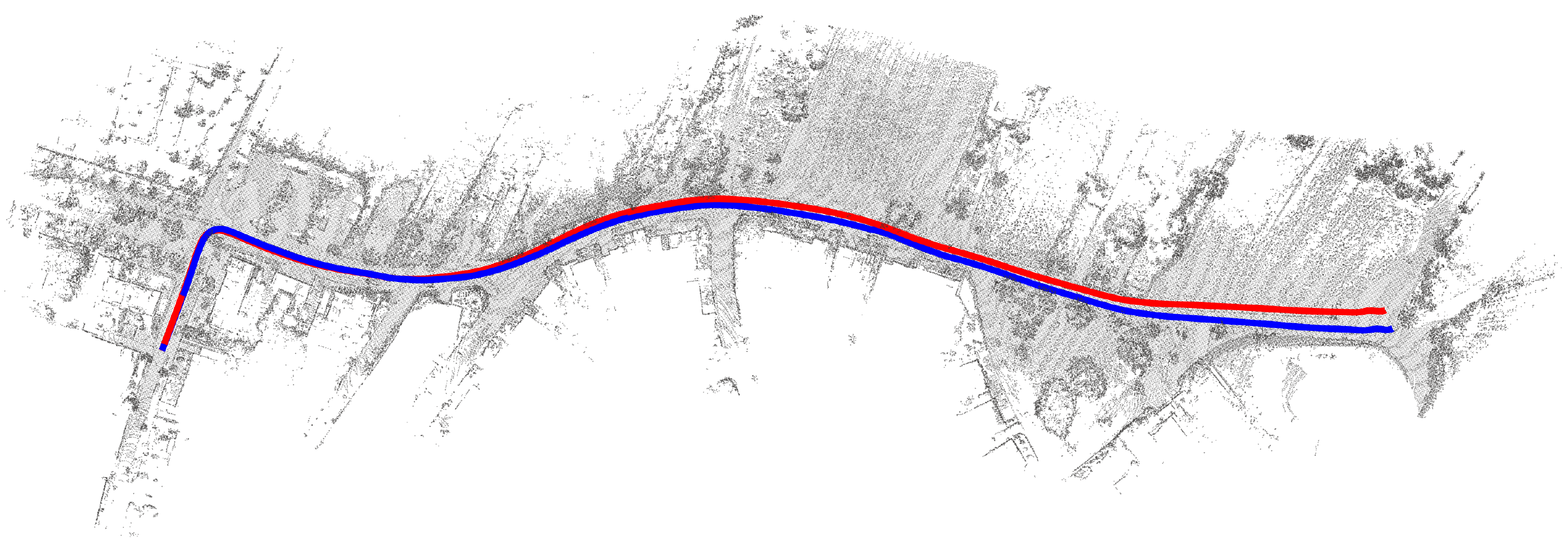}
	}
	
	\caption{Registration results of all compared methods on the KITTI 03 sequence odemetry dataset, where red and blue curves indicate the ground truth and the registration result, respectively. The proposed method shows the highest accuracy.}
	\label{fig:kitti_result}
	\vskip -0.3cm
\end{figure*}

\subsection{Registration on ETH TLS Datasets}
In this experiment, we evaluate our method on a series of partially overlapping and large-scale \emph{ETH terrestrial laser scanner} (TLS) datasets~\cite{DataETHZ}, which have been broadly used in photogrammetry and remote sensing society. We adopt five-pair point clouds from the \textit{Facade} dataset for testing, and downsample them by voxel grid filtering with $\texttt{gridStep=0.1}$ \revise{because each point cloud has} at least $1.5 \times 10^7$ points. Fig.~\ref{fig:ETHZ_example} presents two test examples. We conduct 10 experiments and report the average performance in Table~\ref{tab:ETHZ}, where green and red are the first two highest accuracy. {As observed, our method outperforms competitors with the overall highest precision, followed by FGR and FRICP. The ICP optimized by MATLAB is the fastest, with time consumption around 0.2s. Our algorithm spends about 0.34s on average and is faster than other approaches. TEASER++, FGR, FRICP, and PIPL are also efficient, but their performance is slightly lower than ours \revise{in terms of} accuracy and efficiency. Probabilistic methods again consume the most time, but compared with ICP-based methods, no noticeable improvement of the registration precision \revise{is observed; thus,} they are more suitable for sparse and coarse alignments.} We present examples in Fig.~\ref{fig:ETHZEx}, where values in parentheses are corresponding
 deviations.

\subsection{Registration on KITTI Odometry Datasets}
We also assess the \revise{stability} of the proposed method on the large-scale outdoor \textit{KITTI odometry} benchmark datasets~\cite{geiger2012we} developed by a velodyne laser scanner. The corresponding ground truth poses are determined by \revise{an} IMU measurement and \revise{an} RTK GPS localization system. We adopt the 03 odometry sequence,  including 801 point cloud frames for testing, and downsample them by the voxelized grid with $\texttt{gridStep=0.5}$. {First, we register every other consecutive frame among the whole sequences, \ie, $frame_{801}$ to $frame_{799}$, $frame_{799}$ to $frame_{797}$, $frame_{k}$ to $frame_{k-2}~(k\geq 3)$, $\cdots$, and $frame_{3}$ to $frame_{1}$, to  reconstruct the whole outdoor scene. A total of 401 point cloud pairs are available for registration. Then, we compose these 401 pairwise transformations $\mathcal{T}_1$, $\cdots$, $\mathcal{T}_{401}$ to align frames to the first one. For instance, we use $\mathcal{T}_1\cdot \mathcal{T}_2\cdots\mathcal{T}_{401} (frame_{801})$ to align $frame_{801}$ to $frame_{1}$. Our purpose is to evaluate the performance of the designed algorithm against accumulative deviations}. \revise{Similar to}~\cite{9711460}, we evaluate the average 
errors of the relative and the absolute rotation and translation, as well as {the deviation of the last frame with respect to the first one}. The translation error is the Euclidean distance between the estimated translation vector and the ground truth value.

\begin{figure}[t]
	\centering
	\includegraphics[width=0.235\textwidth]{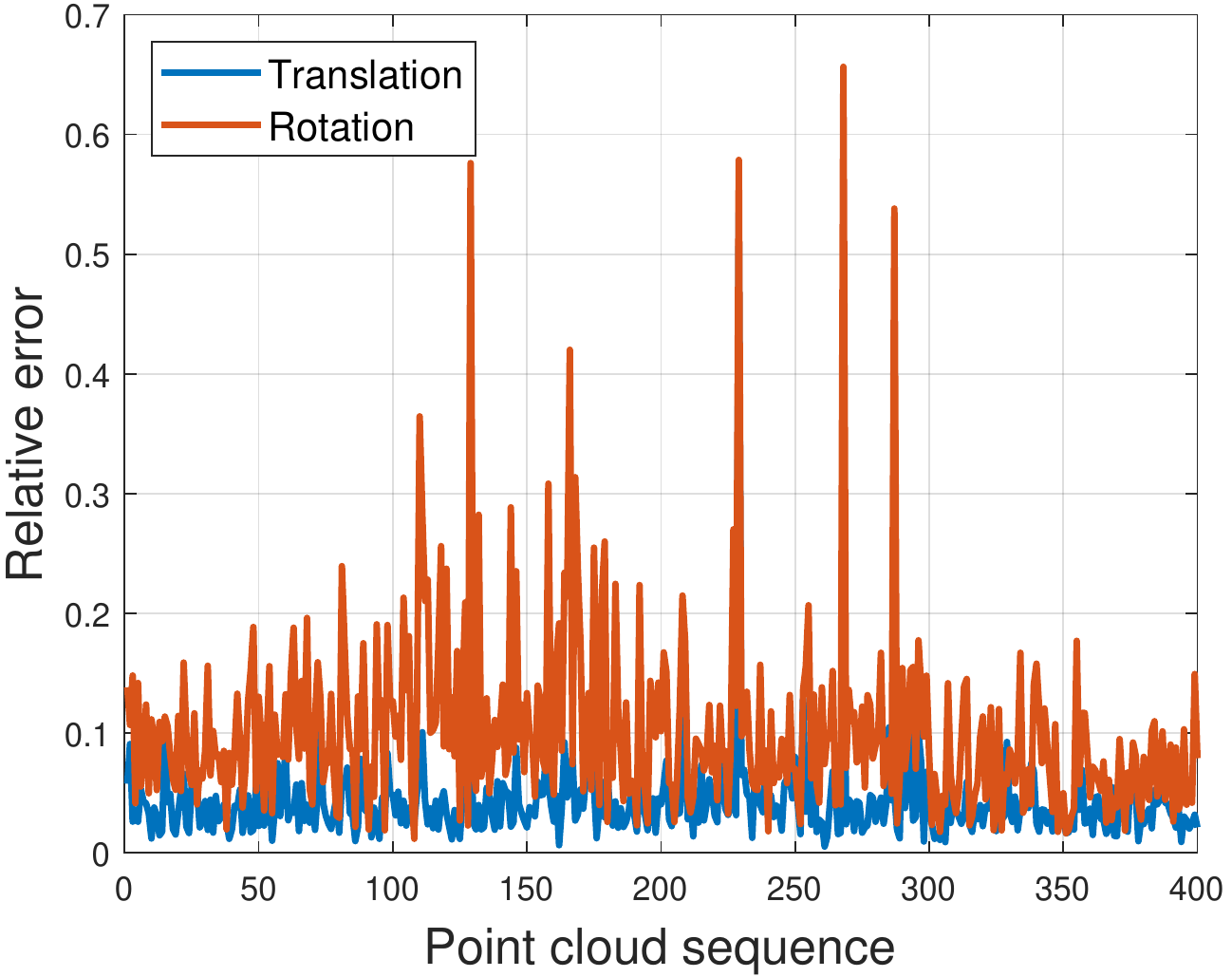}
	\includegraphics[width=0.232\textwidth]{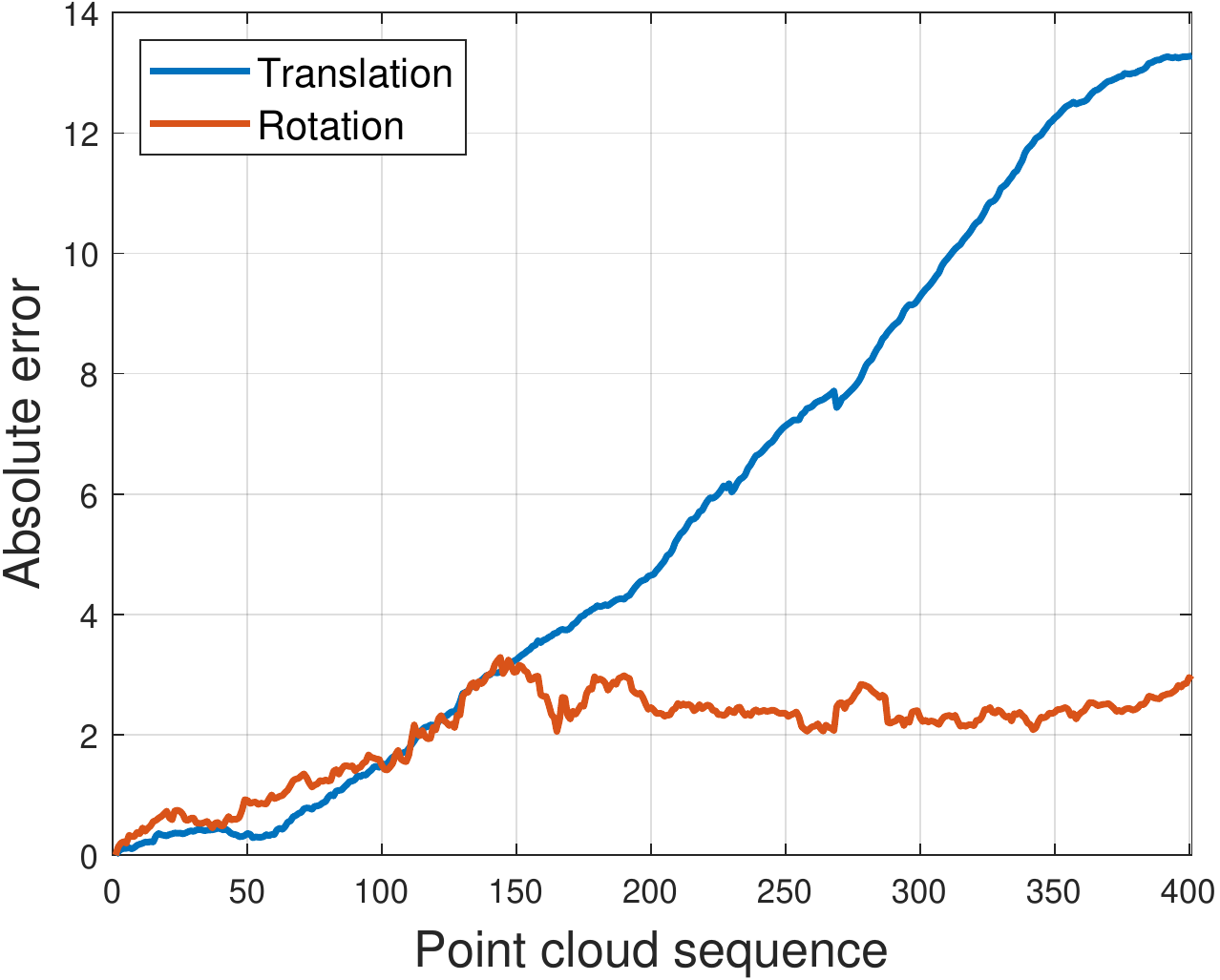}
	\caption{Record of the registration process of the proposed method on 401 point cloud sequences.
	}
	\label{fig:kittiErr}
	\vskip -0.3cm
\end{figure}

\begin{figure*}[!htbp]
	\centering
	\subcaptionbox{Misalignment}{
		\includegraphics[width=0.148\textwidth]{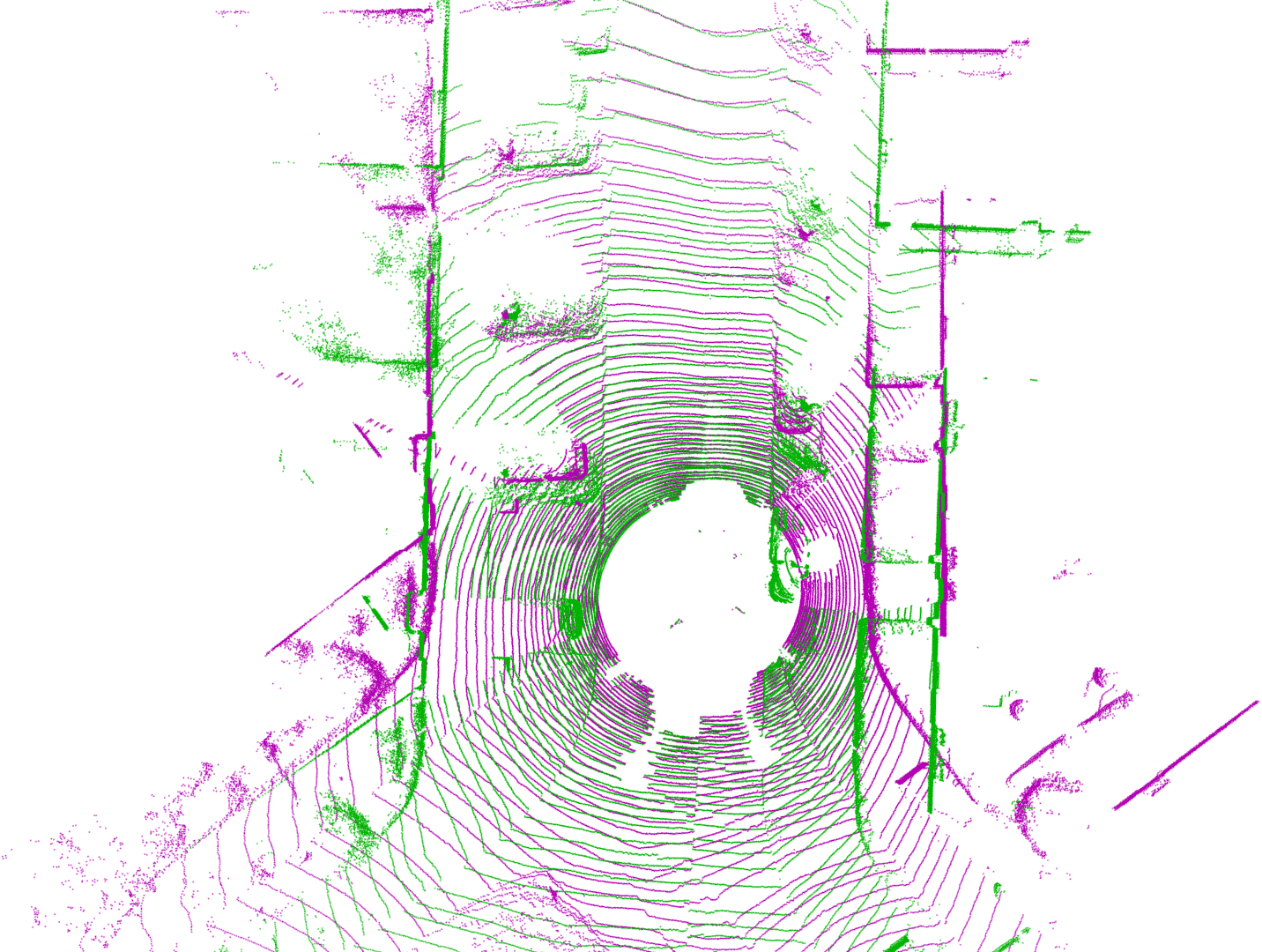}
	}
	\subcaptionbox{ICP}{
		\includegraphics[width=0.148\textwidth]{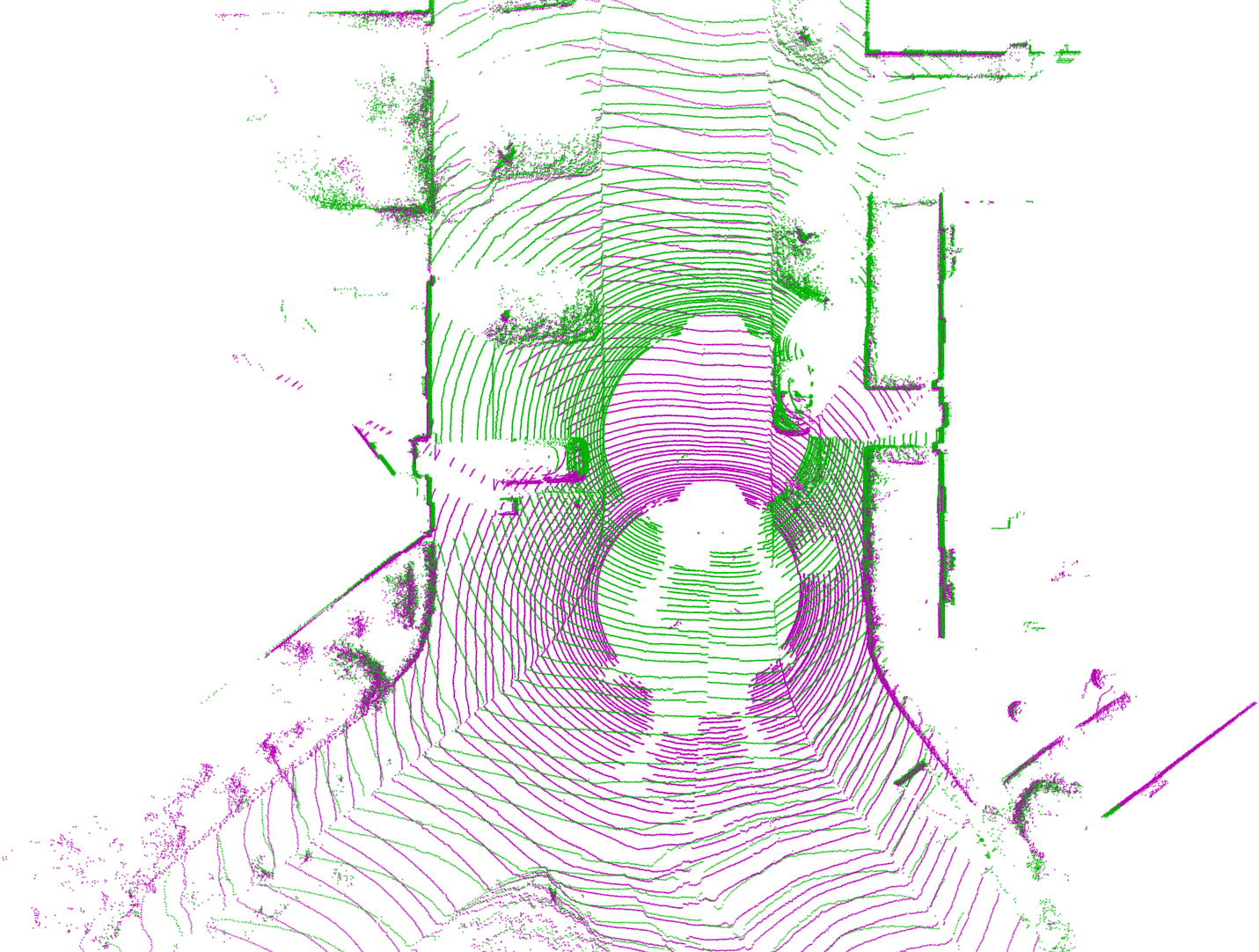}
	}
	\subcaptionbox{ICP-KP}{
		\includegraphics[width=0.148\textwidth]{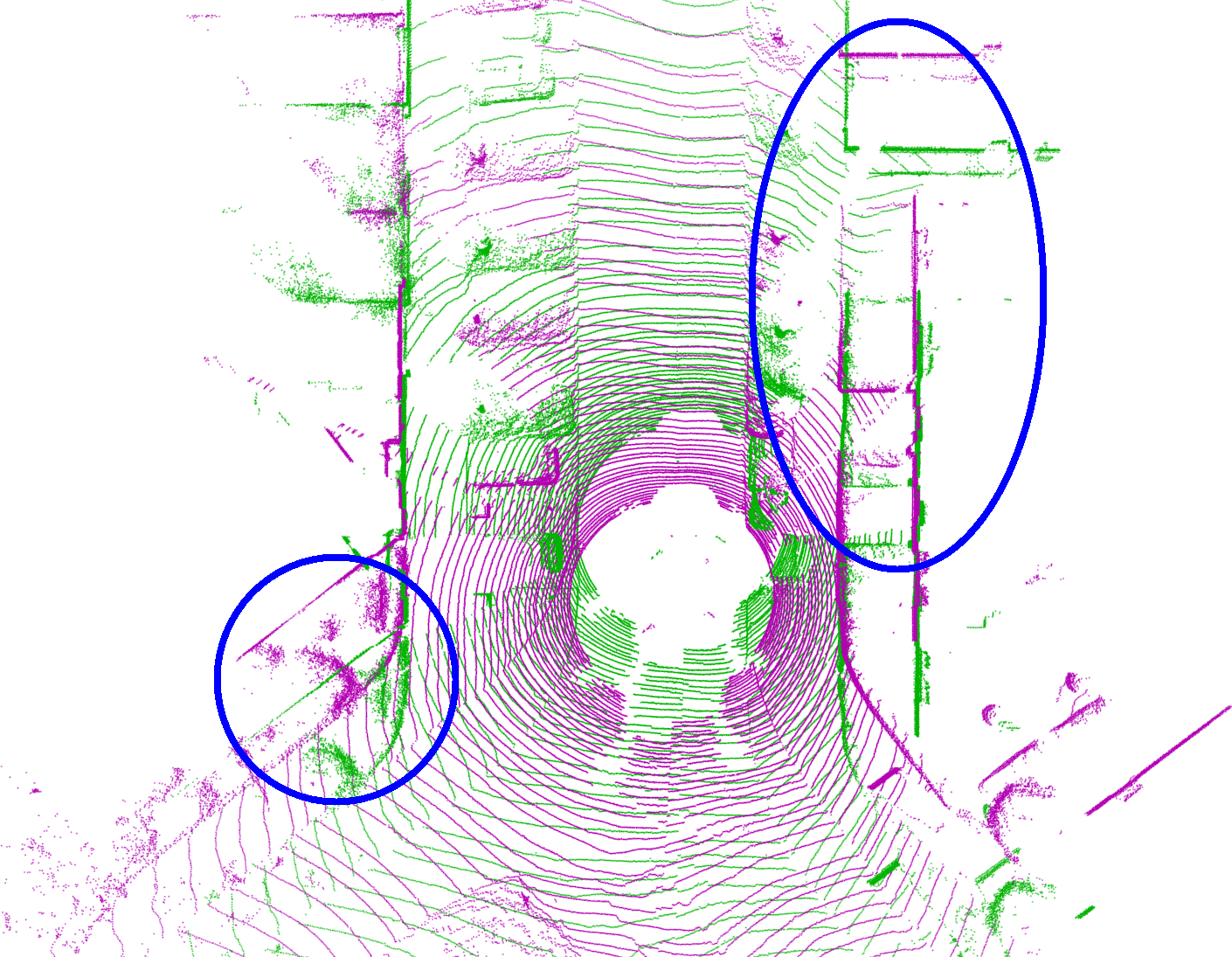}
	}
	\subcaptionbox{ICP-KN}{
		\includegraphics[width=0.148\textwidth]{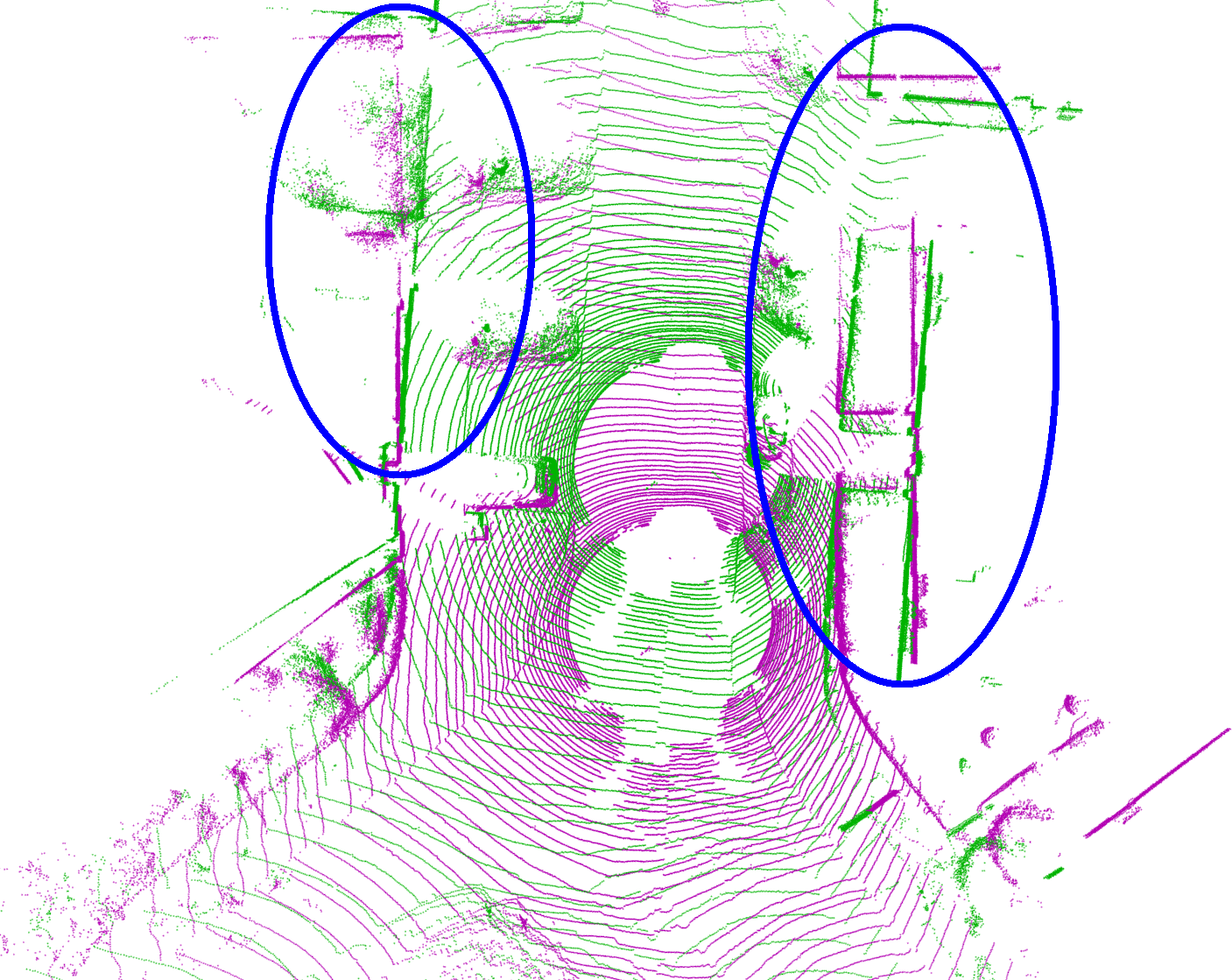}
	}
	\subcaptionbox{GMM}{
		\includegraphics[width=0.148\textwidth]{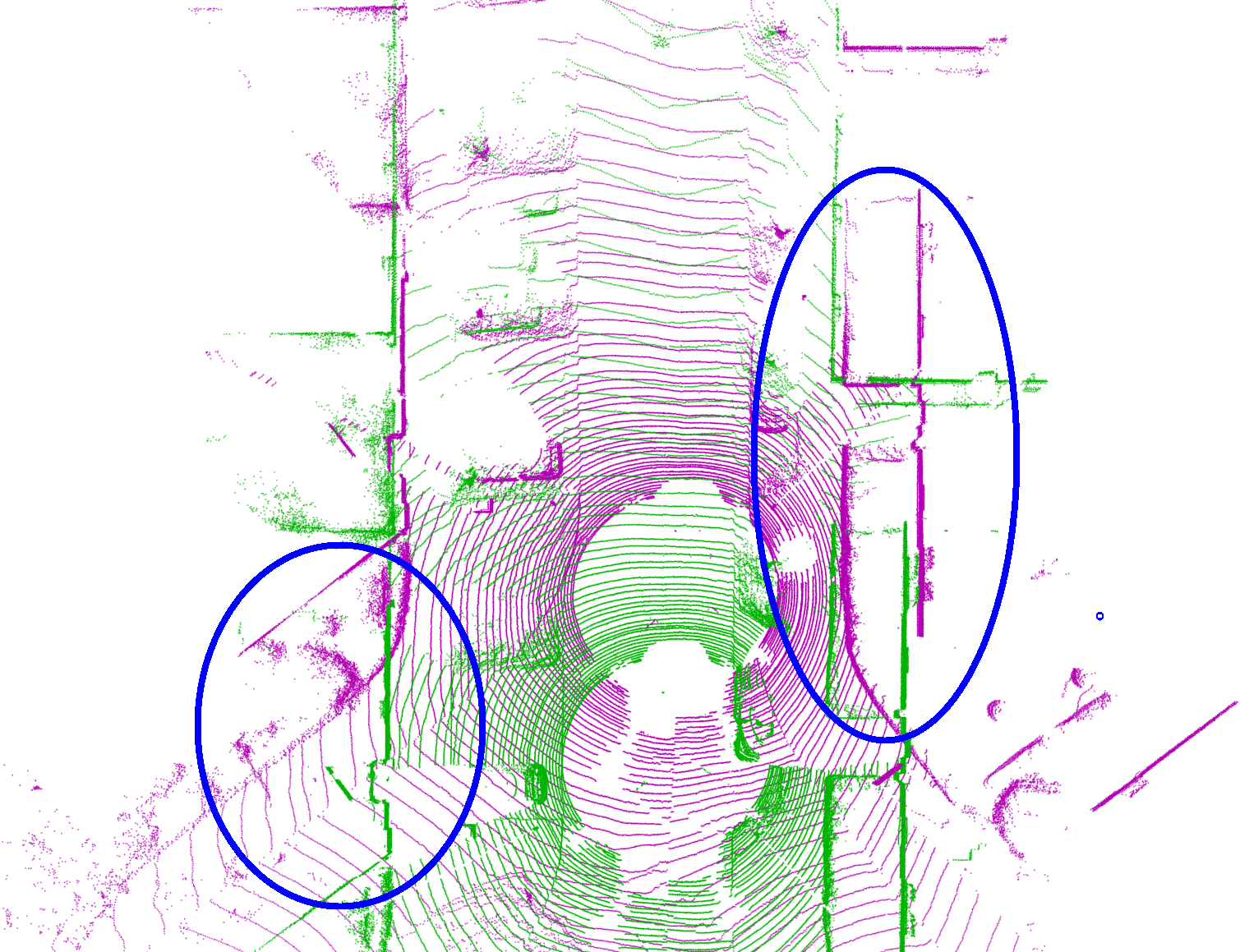}
	}
	\subcaptionbox{CPD}{
		\includegraphics[width=0.148\textwidth]{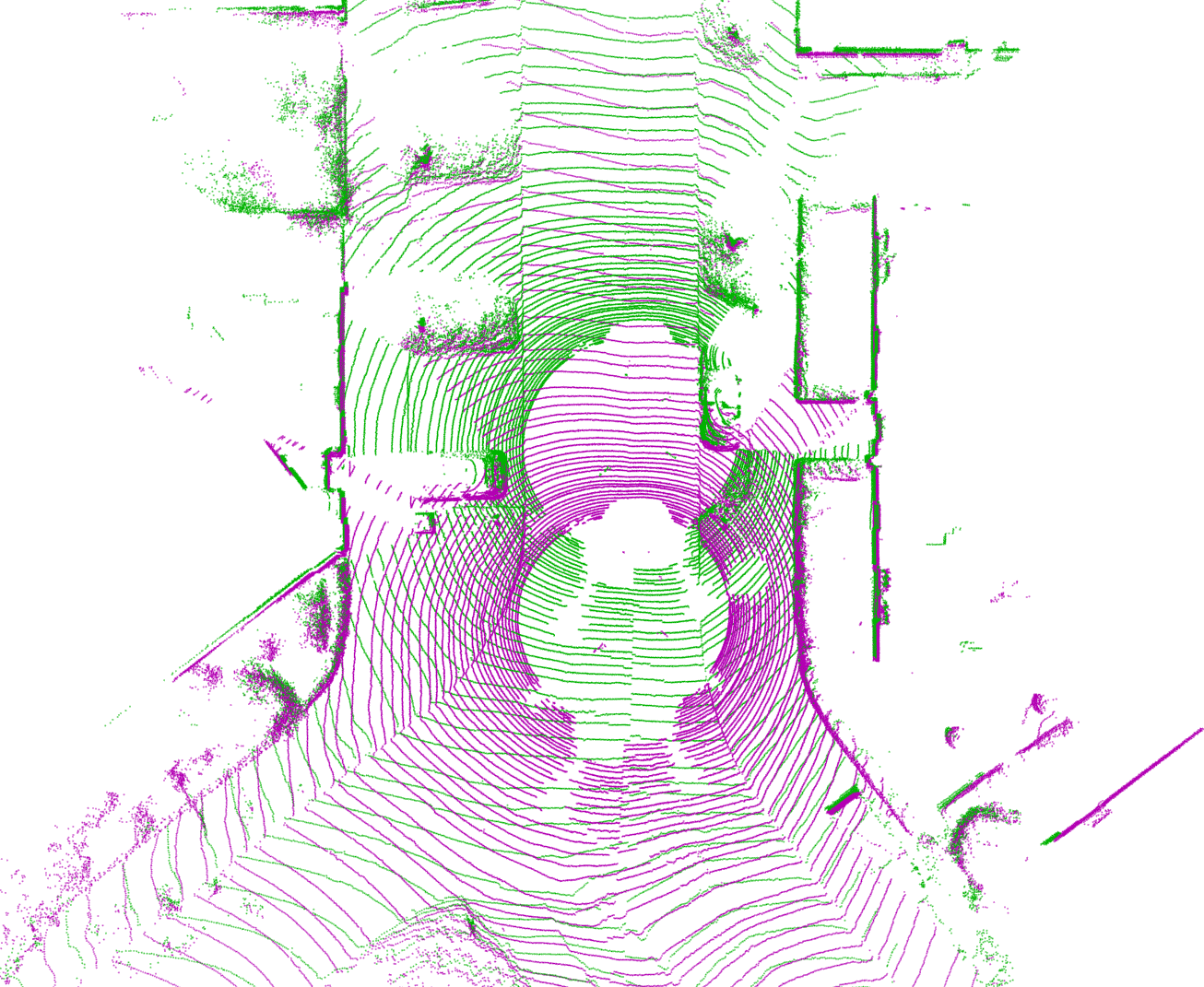}
	}
	
	\subcaptionbox{ECM}{
		\includegraphics[width=0.148\textwidth]{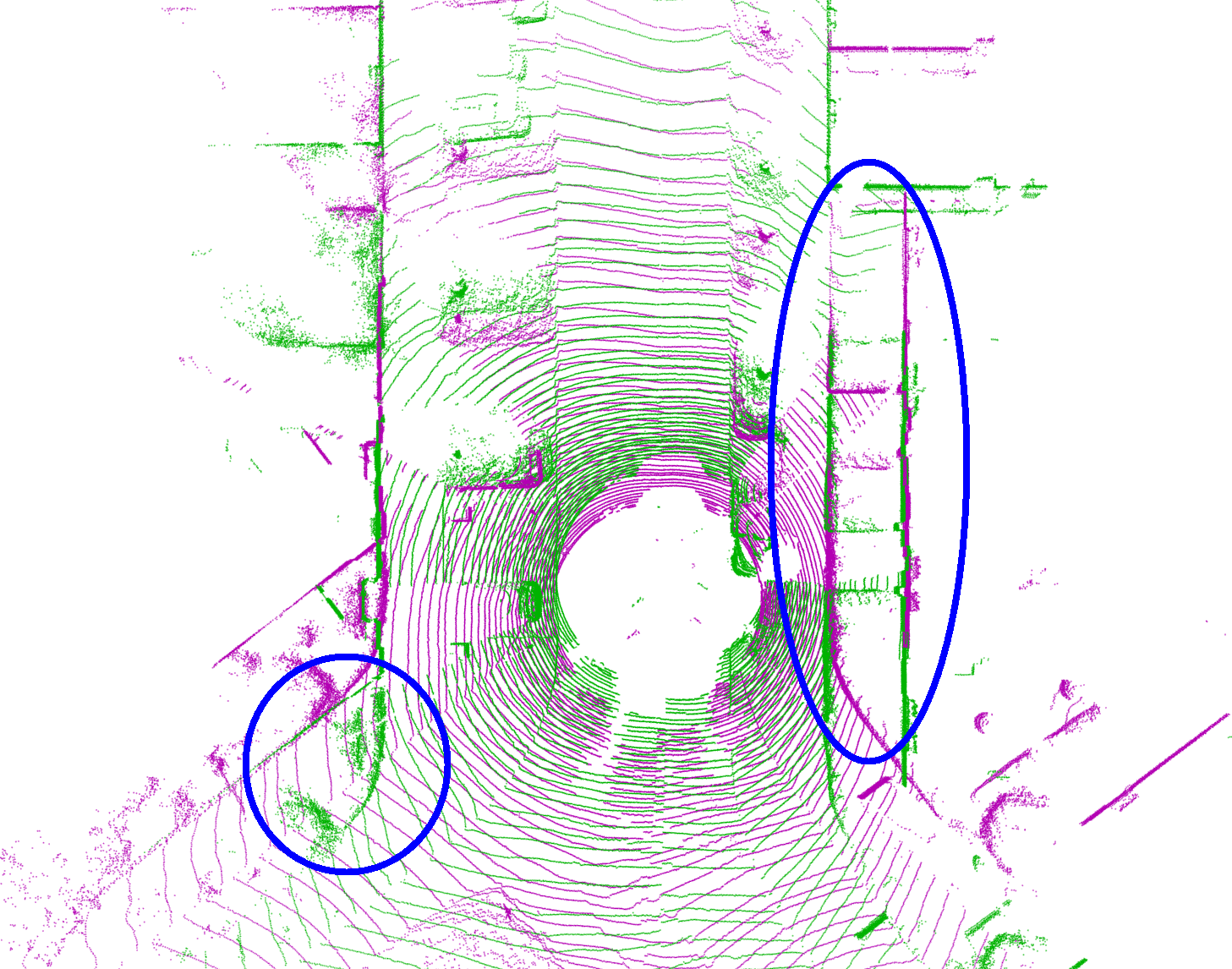}
	}
	\subcaptionbox{\revise{TEASER++}}{
		\includegraphics[width=0.148\textwidth]{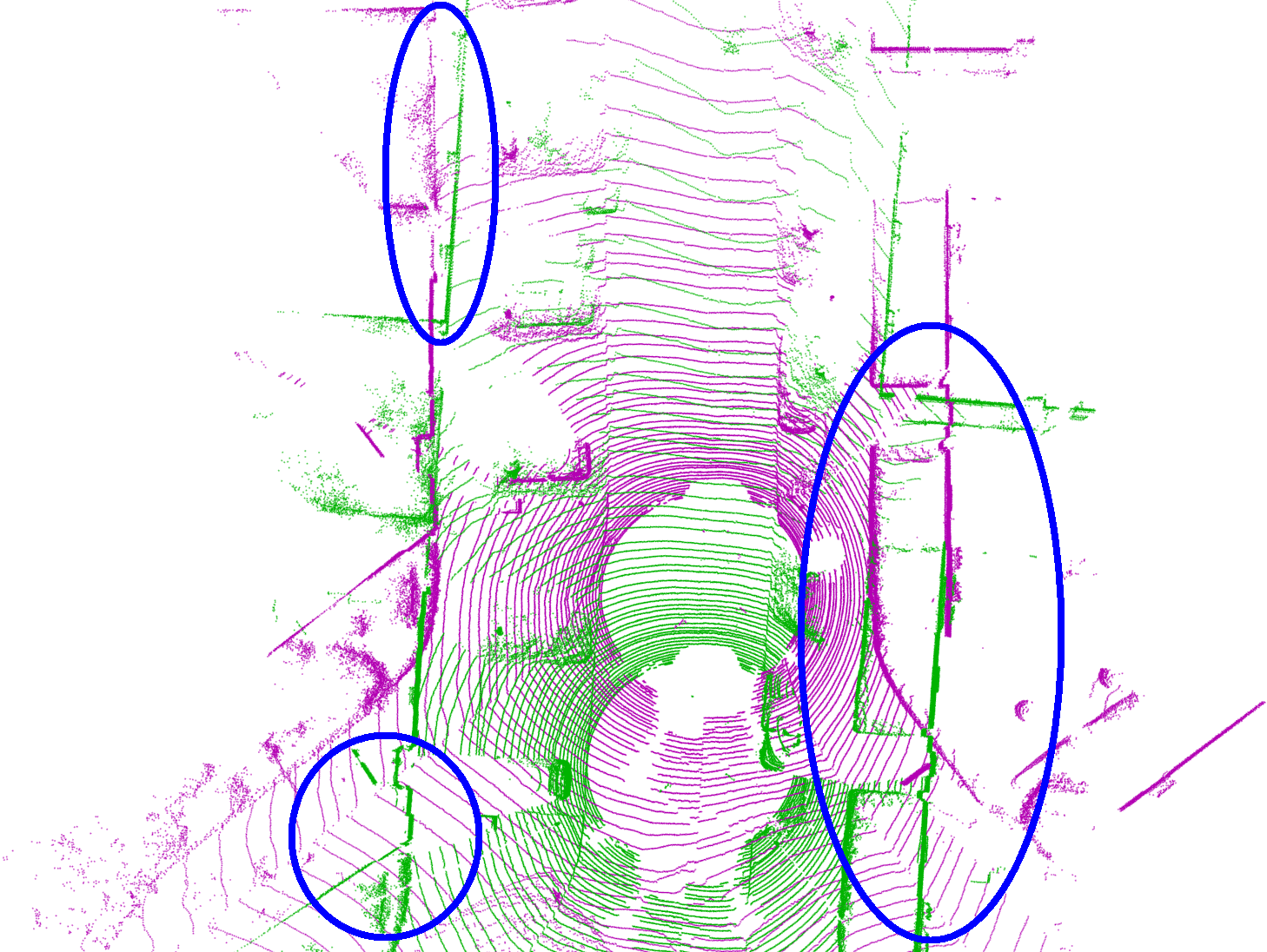}
	}
	\subcaptionbox{\revise{FGR}}{
		\includegraphics[width=0.148\textwidth]{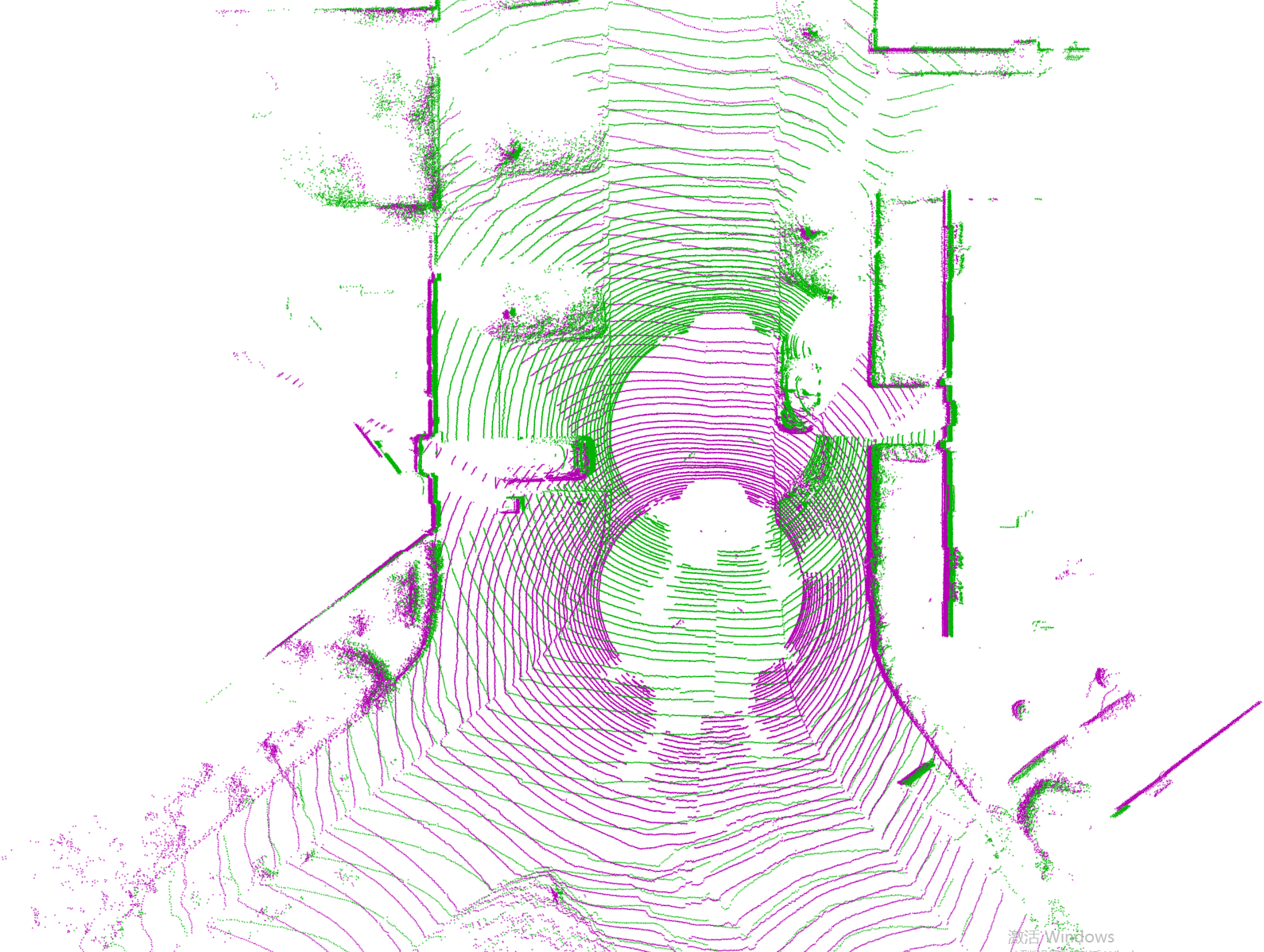}
	}
	\subcaptionbox{\revise{FRICP}}{
		\includegraphics[width=0.148\textwidth]{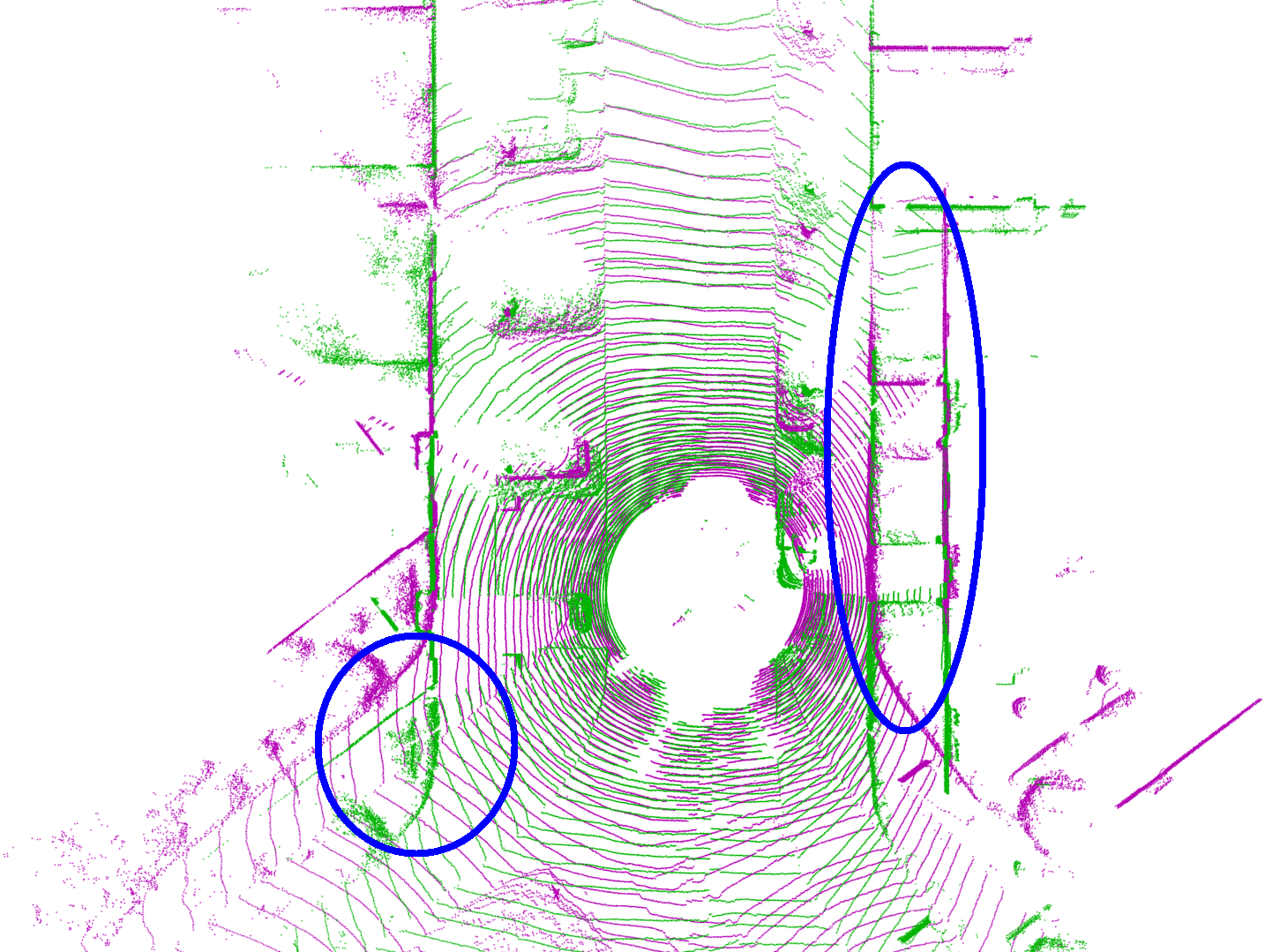}
	}
	\subcaptionbox{PIPL}{
		\includegraphics[width=0.148\textwidth]{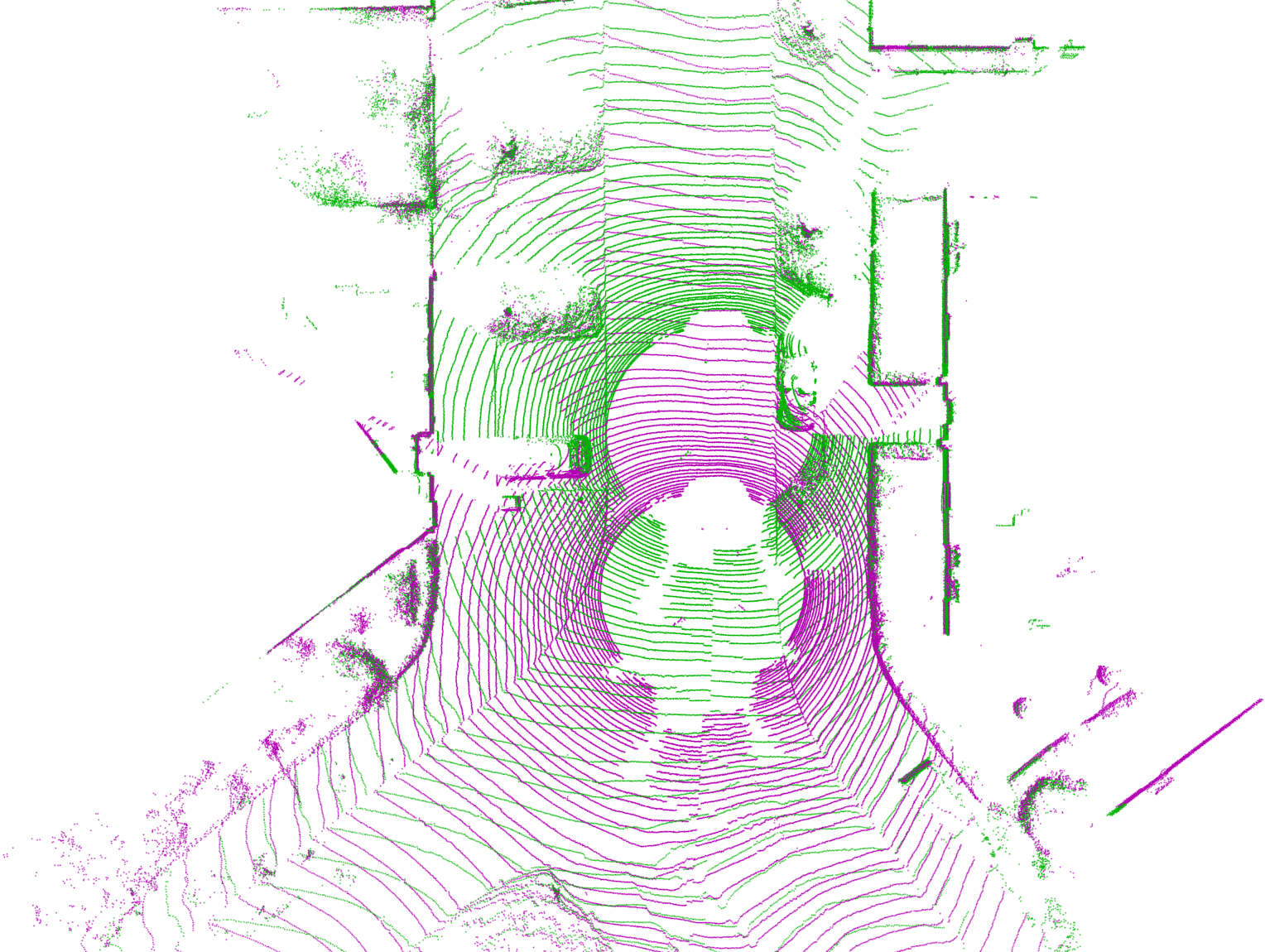}
	}
	\subcaptionbox{Ours}{
		\includegraphics[width=0.148\textwidth]{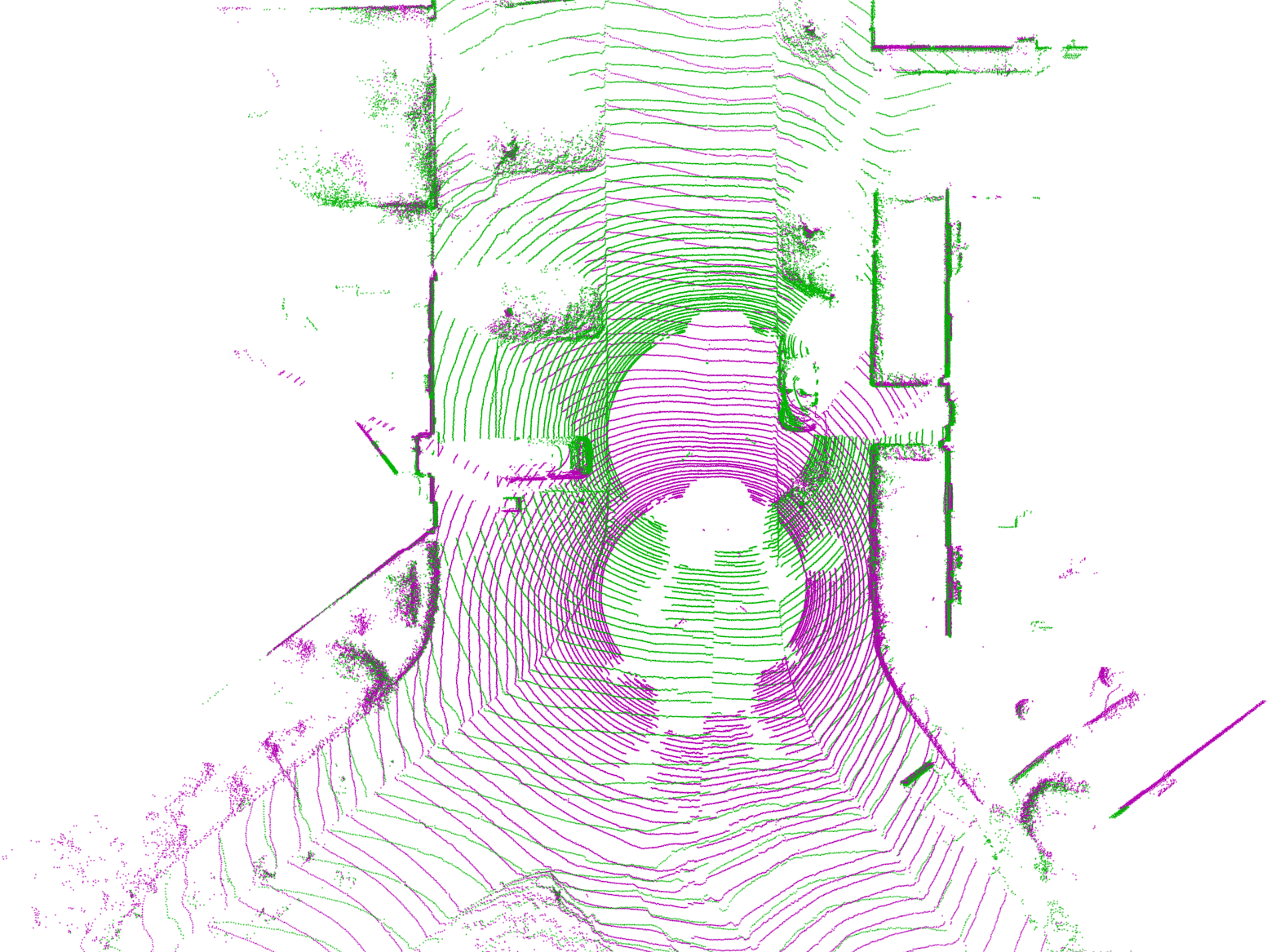}
	}
	\caption{\revise{Qualitative registration results on the KITTI odometry datasets}, where blue ellipses indicate significant errors.} 
	\label{fig:kitti_ex}
	\vskip -0.3cm
\end{figure*}

The quantitative deviations and registration results are reported in Table~\ref{tab:kitti} and Fig.~\ref{fig:kitti_result}, respectively. As observed, our proposed method attains the most accurate results in all situations. The compared methods show minor relative errors but suffer from the absolute case, especially for the translation estimation, indicating that they have relatively large accumulative errors among the registration process. However, {our method performs more stable, has lower errors for rotation and translation estimation of the absolute case, and {achieves more satisfactory registration of the last frame among the whole 401 sequences}}. Fig.~\ref{fig:kittiErr} records the specific registration process of all point cloud frames of our method, \revise{except for} few spikes, all relative errors are small,  with values less than 0.4. Therefore, the whole registration is fairly stable. Absolute translation errors
slightly increase due to the accumulative deviation. However, our method still outperforms competitors with the highest registration precision, and the errors indeed can be further relieved by requesting \revise{additional} advanced techniques, such as stereo images or loop-closure detection. We present some intermediate registration results in Fig.~\ref{fig:kitti_ex}, where blue ellipses indicate large deviations.

\begin{figure*}[t]
	\centering
	
	\subcaptionbox*{}{
		\includegraphics[width=0.48\textwidth]{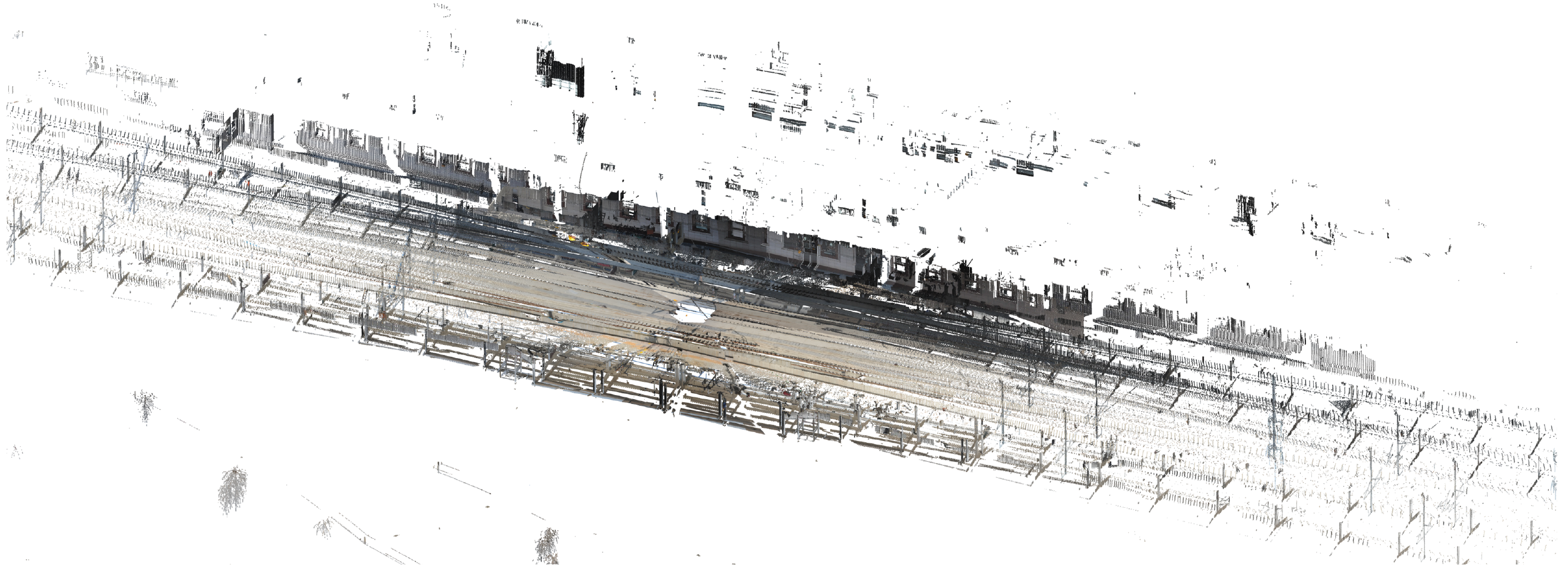}
	}
	\subcaptionbox*{}{
		\includegraphics[width=0.48\textwidth]{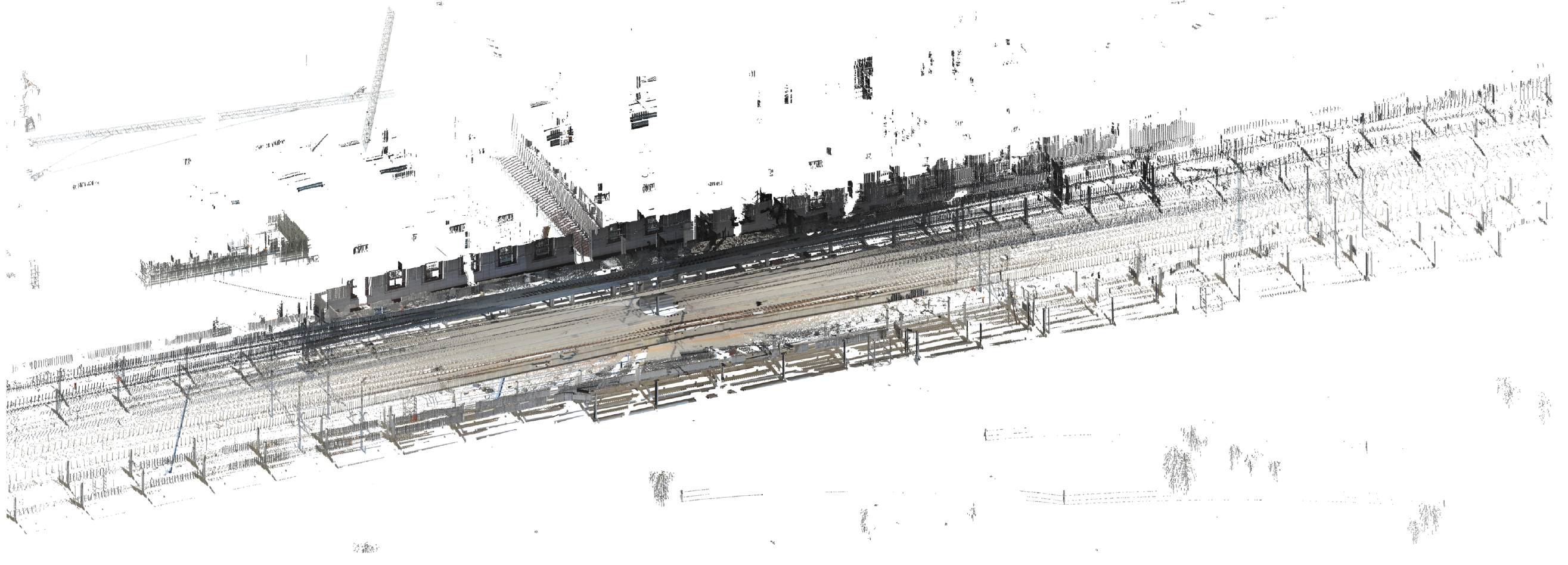}
	}
	
	\caption{\revise{Two pair of scanned real-world railway point clouds endowed with nonuniform point density.}}
	\vskip -0.3cm
	\label{fig:railway}
\end{figure*}
\begin{figure}
	\centering
	\includegraphics[width=0.49\textwidth]{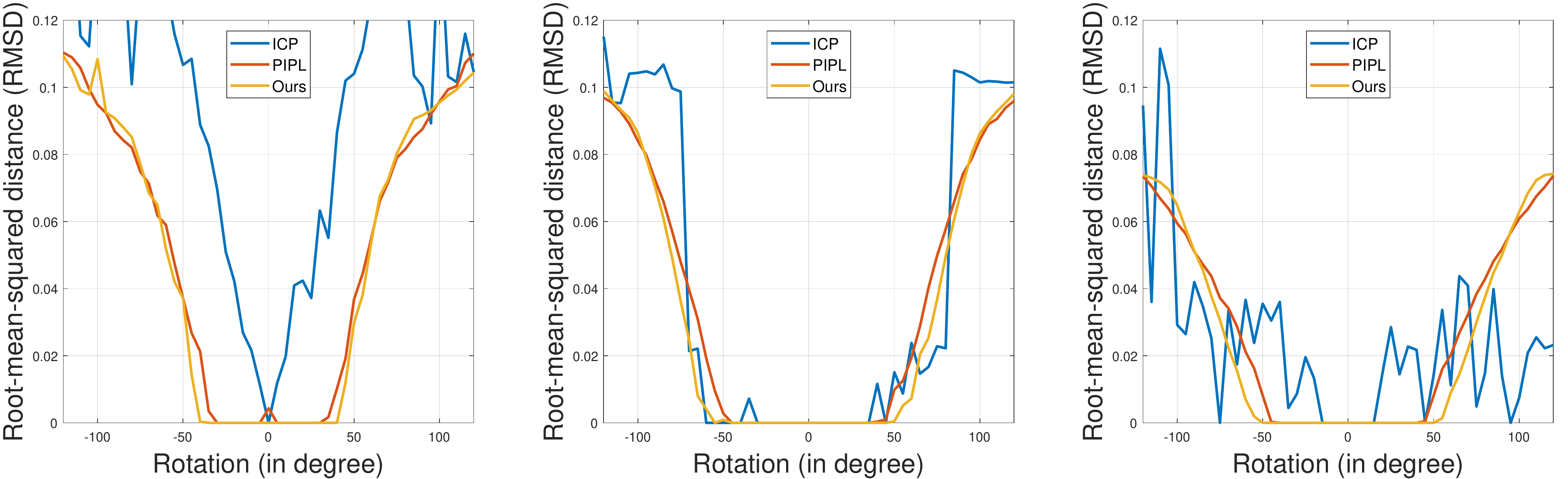}
	\caption{\revise{Robustness test against initial poses. From left to right are the registration results by rotating point clouds along the  $X,$ $Y,$ and $Z$ axes. }}
	\label{fig:rotation}
	\vskip -0.4cm
\end{figure}

\begin{table*}[t]
	
	\centering
	\renewcommand\arraystretch{1.3}
	
	\caption{\revise{Quantitative registration results on the WHU-TLS Railway dataset.}}
	\setlength{\tabcolsep}{0.54mm}{
		\begin{tabular}{|c|l|c|c|c|c|c|c|c|c|c|c|c|}
			\cline{1-13}
			\diagbox{\textbf{Dataset}}{\textbf{Method}}&Metric&ICP\cite{besl1992method} & ICP-KP\cite{kjer2010evaluation}&ICP-KN\cite{kjer2010evaluation}&GMM\cite{jian2010robust}&CPD\cite{myronenko2010point}&ECM\cite{horaud2010rigid}&TEASER++\cite{yang2020teaser}&FGR\cite{zhou2016fast}&FRICP\cite{zhang2021fast}&PIPL\cite{jauer2018efficient}&Ours\\
			\hline
			\multirow{2}{*}{Railway 1}&RMSD&34.4934&40.4979&42.2538&	52.0106&47.8289&	41.8326&179.2847&42.2166&40.4123 &	\cellcolor{second}{37.9666}&	\cellcolor{best}{22.7206}\\ \cline{2-13}
			&AngErr&0.6628&0.6400&3.1715&0.2340&3.7988&0.1641&171.1502&3.7534&\cellcolor{best}{0.0799}&7.5604&\cellcolor{second}{0.1119}\\ \cline{1-13}
			\multirow{2}{*}{Railway 2}&RMSD&54.2542&54.3514&67.3334&	64.7316&54.6875&	50.7399&{213.2731  }&{166.4902}&{40.6263}&	\cellcolor{second}{33.0830}&\cellcolor{best}{30.5180}\\ \cline{2-13}
			&AngErr&2.0299&2.0372&4.7293&0.8820&4.5592&0.1989&73.0722&163.6932&\cellcolor{second}{0.1476}&9.7713&\cellcolor{best}{0.1399}\\ \cline{1-13}
		\end{tabular}
	}
	
	\label{tab:denstiy}
	\vskip -0.3cm
\end{table*}
{\subsection{Robustness against Initial Poses}
We also investigate the robustness of the proposed method against initial poses of point clouds. To this end, we rotate the Dragon model along the $X,$ $Y,$ $Z$ axes separately and change its rotation angle from $[-120^\circ, 120^\circ]$ with the angle interval equal to $5^\circ$. A total of $49\times 3$ point cloud pairs are available for registration. We use the \emph{root-mean-squared distance} (RMSD) defined as
\begin{eqnarray}
	RMSD=\sqrt{\frac{1}{M}\sum_{i=1}^{M}\|\hat{\mathcal{T}}\bm{x}_i-{\mathcal{T}}\bm{x}_i\|_2^2}
\end{eqnarray}
to assess the registration performance, where $\hat{\mathcal{T}}$ and ${\mathcal{T}}$ are the solved and ground truth 
transformation, respectively. Fig.~\ref{fig:rotation} shows the registration results against rotation angles. ICP is involved here as an indicator to the sensitivity of the initial poses. For all cases, the proposed method has wider angle range for correct registration than the baseline PIPL and is much larger than ICP. Even for the first case where ICP requires strict initialization, our method still tolerates nearly $45^\circ$ rotation angle, suggesting its high robustness against initial poses and the conservation of efforts for initialization tuning.}  
{\subsection{Influence of Nonuniform Point Density}
We further perform experiments to investigate the influence of nonuniform point density for point cloud registration. As presented in Fig.~\ref{fig:railway}, we use two pairs of scanned outdoor point clouds named Railway 1 and Railway 2 from the WHU-TLS dataset~\cite{dong2017novel} for testing. As observed, these point clouds have nonuniform point density, and their central parts are much denser than both ends. The registration results regarding RMSD and AngErr are reported in Table~\ref{tab:denstiy}. Except for TEASER++ and FGR which generate significant AngErr, the other approaches attain acceptable AngErr. Moreover, FRICP and our method have the most accurate rotation estimation. Meanwhile, our algorithm also outperforms competitors with the lowest RMSD, indicating that it is more robust and stable against nonuniform point clouds. Considering space limitation, we present the registration results of each method in Supplemental Material.}

\begin{figure}[h]
	\centering
	\subcaptionbox[]{Gaussian}{
		\includegraphics[width=0.11\textwidth]{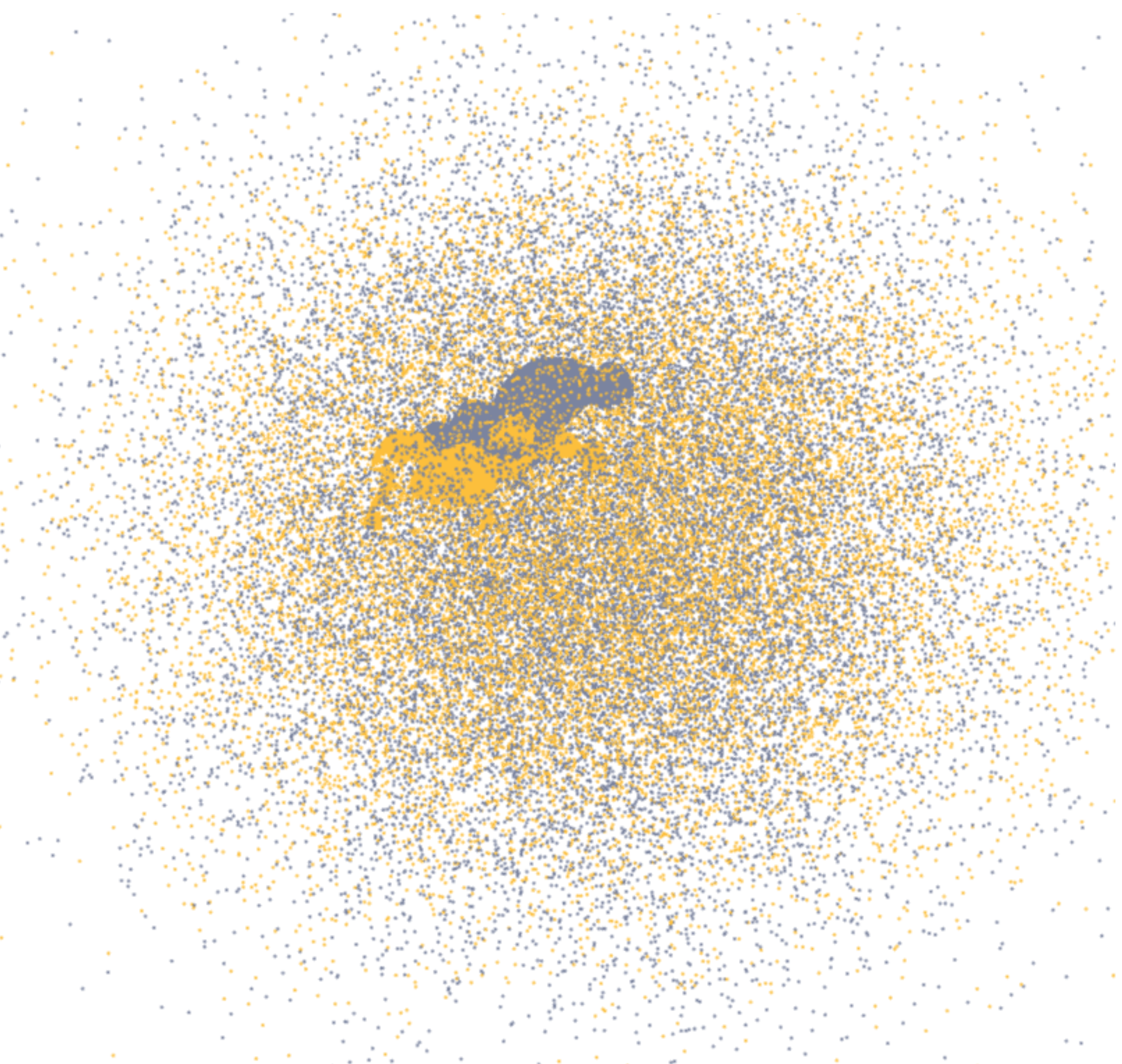}
		}
		\subcaptionbox[]{Uniform}{
		\includegraphics[width=0.11\textwidth]{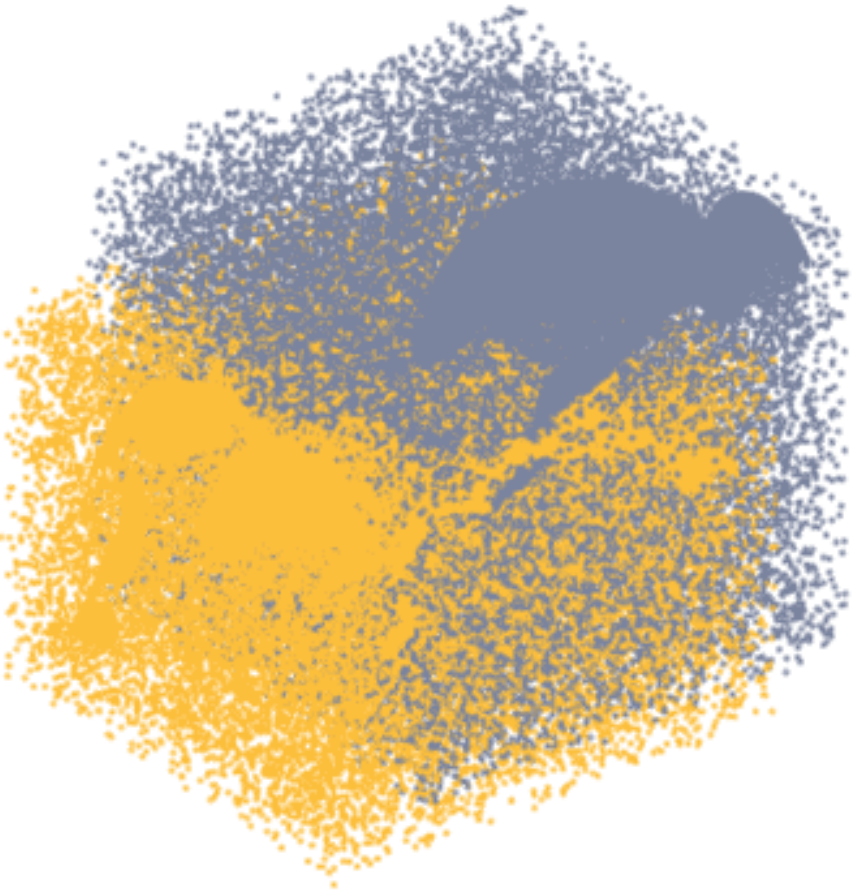}
	}
	\subcaptionbox[]{RMSD}{
		\includegraphics[width=0.2\textwidth]{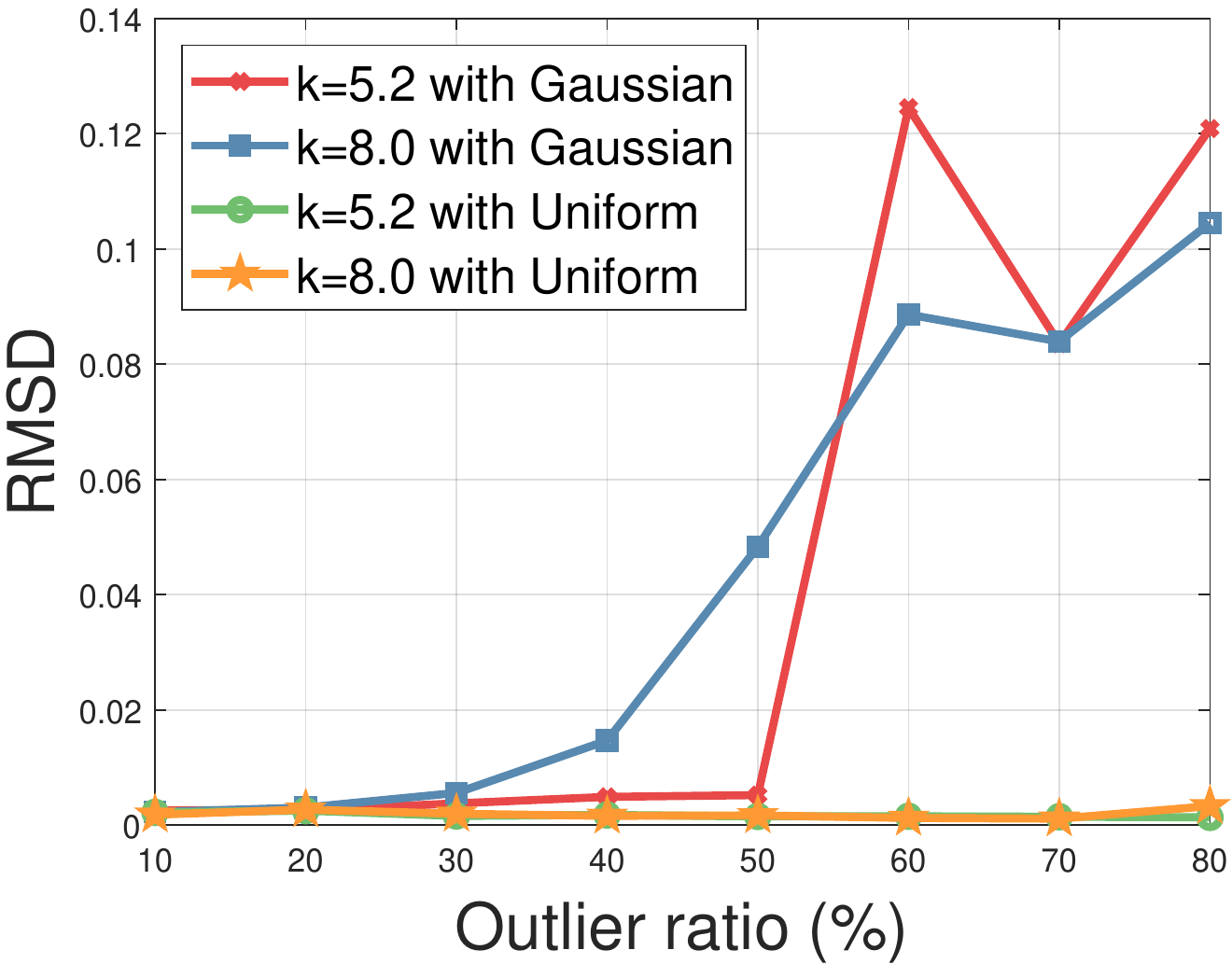}  
	}

	\subcaptionbox[]{RMSD=0.0035 ($k=5.2$)}{
	\includegraphics[width=0.22\textwidth]{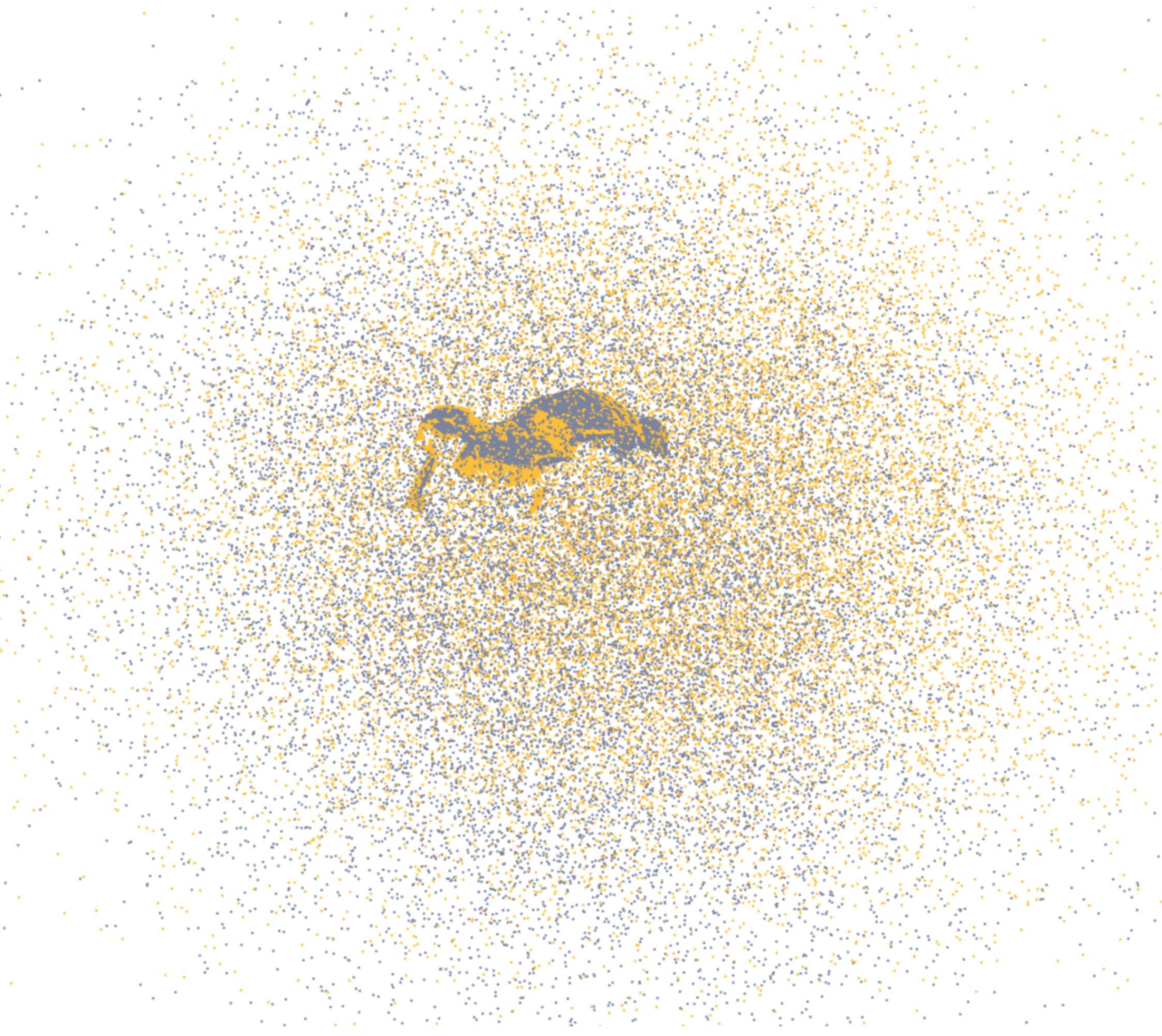}
	}
	\subcaptionbox[]{RMSD=0.0032 ($k=5.2$)}{
	\includegraphics[width=0.22\textwidth]{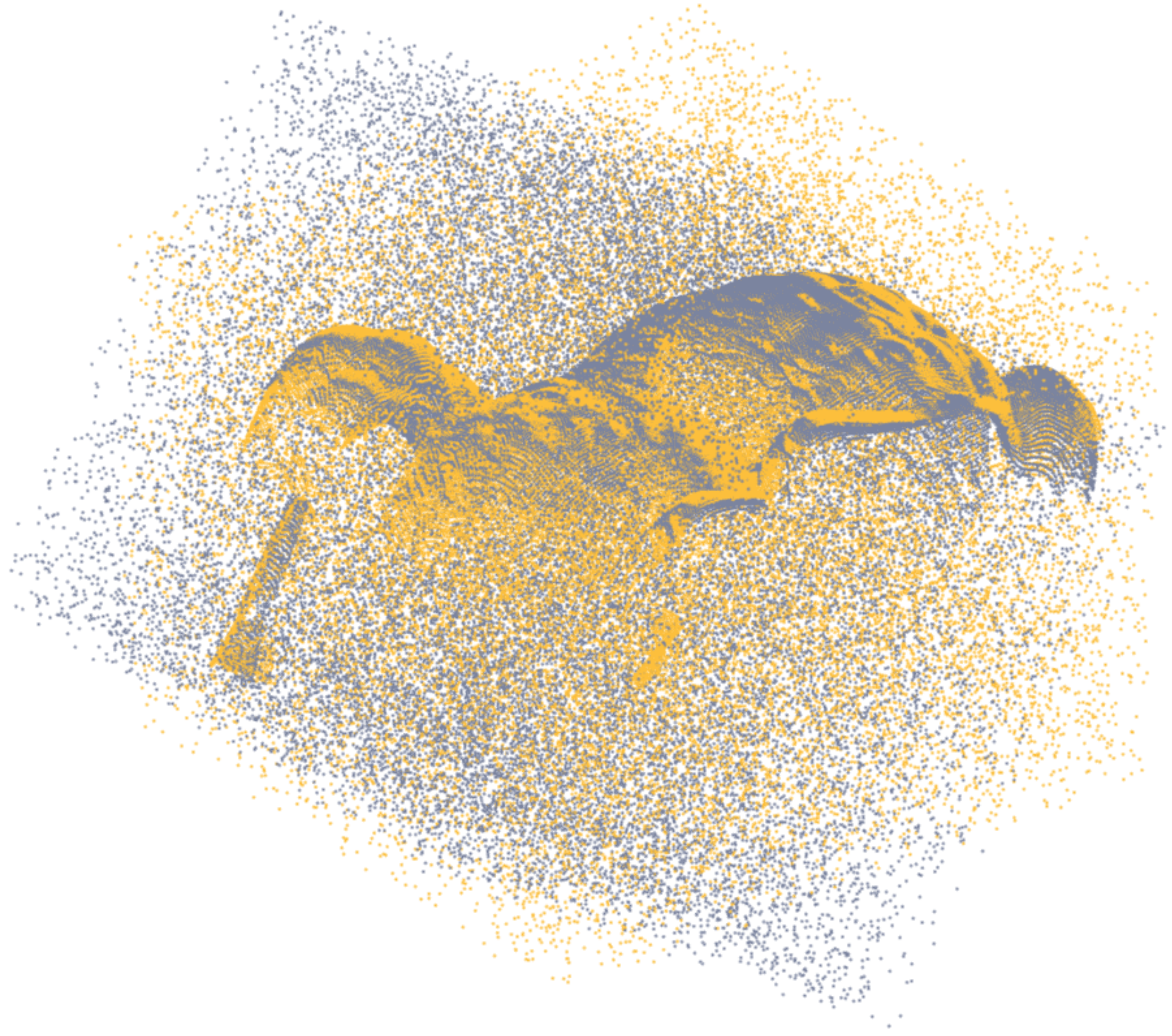}
	}
	\vskip -0.2cm
	\caption{\revise{Investigation of the $X84$ principle for outlier depression. $k=5.2$ gives high robustness and is suggested for practical use especially when the outlier type is unknown.}}
	\label{fig:bunny_outlier}
	\vskip -0.5cm
\end{figure}

\revise{
\subsection{Investigation of the $X84$ Principle}
We also investigate the choice of $\alpha$ in the $X84$ principle, \ie, \revise{$\mathcal{I}(\bm{x}_i)>\alpha \cdot \text{MAD}(\mathcal{I}(\mathbf{X}))$}, for the influence of robustness against outliers. We contaminate the two Bunny {models (000, 045)~\cite{DataStanford} with a set of outliers sampled from either Gaussian or uniform distribution.} The Gaussian distribution is with zero mean, and its standard variance along $X, Y, Z$ is separately equal to the size of the three sides of the point cloud bounding box, whereas the uniform distribution is directly built on the basis of the point cloud bounding box. For instance, Fig.~\ref{fig:bunny_outlier}(a) and (b) give two test samples with 50\% Gaussian outliers and 80\% uniform outliers, respectively. We increase the outlier ratio from 10\% to 80\% and vary $\alpha=5.2$ to  $\alpha=8.0$ for both Gaussian and uniform outliers to assess the $X84$ principle. Results in Fig.~\ref{fig:bunny_outlier}(c) suggest that under Gaussian outlier contamination, $k=5.2$ has the breakdown point up to $50\%$, whereas $k=8.0$ only tolerates $30\%$ of the outliers. As for uniform outliers, $k=5.2$ and $k=8.0$ attain highly robust performance. Therefore, the $X84$ principle is credible, and $k=5.2$ is suggested for practical use, especially without a priori information of the outlier type. Registration results of Fig.~\ref{fig:bunny_outlier}(a) and (b) by our method (with $k=5.2$) are reported in Fig.~\ref{fig:bunny_outlier}(d) and (e), respectively. 
}

\section{Discussion and Conclusions}\label{conclusion}
We presented a precise and efficient pipeline for robust point cloud registration in various 3D scenes from range scanners to LiDAR. Our first contribution is exploiting GSP to describe the local surface variation, from which we define the {point response intensity} and theoretically prove its invariance under rigid transformations. Together, the \revise{point response intensity} \revise{combined} with median absolute deviation in robust statistics successfully removes most noisy outliers, resulting in more stable and accurate registration. Moreover, we take \revise{point response intensity} as a metric to assess the particle mass, which naturally reflects  the position distribution importance. We also embed higher-order features, such as surface normals and curvatures, in GSP to create a novel geometric invariant under rigid transformations, thereby enabling credible correspondence among point pairs, boosting the modeling of forces, and  discerning of different particles. Finally, we introduce the adaptive simulated annealing 
framework for high-efficiency solving of the global optimal transformation. \revise{Our method improves the physics-based point cloud registration framework regarding accuracy, robustness, and efficiency.}

Extensive experiments on a wide variety of datasets \revise{confirm} the accuracy, efficiency, and robustness of our proposed method, indicating its promising performance for processing large-scale scans. Given the convenient feature representation mechanism, GSP-induced invariants can be generalized to ICP-based and probabilistic-based registration, reducing their computational complexity and enhancing the point matching confidence. The currently proposed method is limited to the rigid transformation without scaling and \revise{pays minimal} attention on robust loss functions. Future work will seek for \emph{scale-invariant} graph filters that can further enrich the geometric feature description, which will be conducive to \emph{cross-modal} registration, where scale difference commonly exists, \eg, the point clouds captured by Kinect cameras and \revise{laser scanners}. We also plan to invoke \textit{M/S-estimators}, \emph{adaptive influence function}, or \emph{robust $L_p$ minimization} to current work to improve its robustness further. Finally, \revise{in a future study,} the rigid framework \revise{can be extended} to non-rigid registration by modeling forces via \textit{fluid/elastic mechanics}, and continuous time series through integral \revise{can be used} to analyze the complex deformable process.

\bibliographystyle{IEEEtran}
\bibliography{references_pc}

\vfill

\end{document}